\documentclass[runningheads]{llncs}

\usepackage{accv}

\usepackage{accvabbrv}

\usepackage{graphicx}
\usepackage{booktabs}

\usepackage[accsupp]{axessibility}

\usepackage[breaklinks,colorlinks,citecolor=accvblue]{hyperref}

\usepackage{orcidlink}

\PassOptionsToPackage{table,dvipsnames}{xcolor}
\usepackage{colortbl}

\definecolor{Fraunhofergreen}{RGB}{23,156,125}
\definecolor{Fraunhofersteelblue}{RGB}{0,91,127}
\definecolor{Fraunhofersilvergrey}{RGB}{166,187,200}
\definecolor{Fraunhoferorange}{RGB}{245,130,32}
\definecolor{Fraunhofergraphit}{RGB}{28,63,82}
\definecolor{Fraunhofersand}{RGB}{211,199,174}
\definecolor{Fraunhoferpetrol}{RGB}{0,133,152}
\definecolor{Fraunhoferaqua}{RGB}{57,193,205}
\definecolor{Fraunhoferlime}{RGB}{178,210,53}
\definecolor{Fraunhoferblue}{RGB}{31,130,192}
\definecolor{Fraunhoferred}{RGB}{187,0,86}
\definecolor{Fraunhoferweinrot}{RGB}{124,21,77}

\definecolor{KITred}{RGB}{160,30,40}
\definecolor{KITblue}{RGB}{70,100,170}
\definecolor{KITblue70}{RGB}{125,146,195}
\definecolor{KITblue50}{RGB}{162,177,212}
\definecolor{KITblue30}{RGB}{199,208,229}
\definecolor{KITblue15}{RGB}{227,232,242}

\definecolor{Set2-01}{RGB}{102, 194, 165}
\definecolor{Set2-02}{RGB}{252, 141,  98}
\definecolor{Set2-03}{RGB}{141, 160, 203}
\definecolor{Set2-04}{RGB}{231, 138, 195}

\definecolor{Set3-01}{RGB}{141, 211, 199}
\definecolor{Set3-02}{RGB}{255, 255, 179}
\definecolor{Set3-03}{RGB}{251, 128, 114}
\definecolor{Set3-04}{RGB}{128, 177, 211}
\definecolor{Set3-05}{RGB}{179, 222, 105}
\definecolor{Set3-06}{RGB}{252, 205, 229}
\definecolor{Set3-07}{RGB}{188, 128, 189}
\definecolor{Set3-08}{RGB}{204, 235, 197}

\colorlet{colorFst}{Green!20}
\colorlet{colorSnd}{SpringGreen!45}
\colorlet{colorTrd}{Yellow!30}

\usepackage{tikz}
\usetikzlibrary{
	automata,
	arrows,
	backgrounds,
	calc,
	chains,
	decorations,
	decorations.markings,
	decorations.pathmorphing,
	external,
	fadings,
	fit,
	matrix,
	positioning,
	scopes,
	shadows,
	shapes.geometric,
	shapes.misc,
	shapes.multipart,
	through,
	mindmap,
	backgrounds,
	topaths,
	arrows.meta,
	decorations.pathreplacing
}

\tikzset{
	pics/square/.default={1},
	pics/square/.style = {
		code = {
			\draw[pic actions, ultra thick] (0,0) rectangle (#1,#1);
		}
	},
	icon/.style={
		draw, rounded corners=2pt, minimum width=9mm, minimum height=6mm,
		font=\footnotesize\sffamily, align=center
	},
	src/.style={icon, fill=blue!10},
	eff/.style={icon, fill=green!10},
	grp/.style={icon, fill=orange!20},
}

\usepackage{amsmath}

\usepackage{amsfonts}
\usepackage{amssymb}

\usepackage{indentfirst}
\usepackage{comment}
\usepackage{enumitem}
\usepackage[font=footnotesize,skip=3pt,subrefformat=parens]{subcaption}

\definecolor{darkspringgreen}{rgb}{0.09, 0.45, 0.27}
\definecolor{bistre}{rgb}{0.24, 0.17, 0.12}
\definecolor{auburn}{rgb}{0.43, 0.21, 0.1}
\definecolor{britishracinggreen}{rgb}{0.0, 0.26, 0.15}

\usepackage{pdflscape}
\usepackage{threeparttable}
\usepackage{multirow}
\usepackage{csquotes}
\usepackage{textpos}
\usepackage{adjustbox}
\usepackage[]{todonotes}
\usepackage{arydshln}
\usepackage{pgfplots}

\usepackage{utfsym}
\newcommand{\cmark}{\checkmark}
\newcommand{\xmark}{\scalebox{0.75}{\usym{2613}}}

\usepackage[linesnumbered,ruled,vlined]{algorithm2e}

\newcommand{\smallsec}[1]{\vspace{0.75em}\noindent\textbf{#1}\;}

\newcommand{\subw}{0.15\textwidth}

\newcommand{\smalltable}{
	\scriptsize
	\setlength{\tabcolsep}{3pt}
	\renewcommand{\arraystretch}{0.9}
}
\newcommand{\aurocTableSetup}{
	\smalltable
	\centering
}

\newif\ifshowextra
\showextrafalse

\newcommand{\degfunc}[1]{{'#1'}}

\usepackage{listings}
\usepackage{multicol}
\lstdefinelanguage{yaml}{
	morekeywords={true, false, null},
	sensitive=true,
	morecomment=[l]{\#},
	morestring=[b]",
}
\usepackage{placeins}
\begin{document}
	
	\title{Self-Aware Object Detection via \\ Degradation Manifolds}
	
	\titlerunning{Self-Aware Object Detection via Degradation Manifolds}
	
	\author{
		Stefan Becker\inst{1}\orcidlink{0000-0001-7367-2519} \and
		Simon Weiss\inst{1}\orcidlink{0009-0008-8171-9397} \and
		Wolfgang H{\"u}bner\inst{1}\orcidlink{0000-0001-5634-6324} \and
		Michael Arens\inst{1}\orcidlink{0000-0002-7857-0332}
	}
	
	\authorrunning{S.~Becker et al.}
	
	\institute{Fraunhofer IOSB, Ettlingen, Germany\\
		Fraunhofer Center for Machine Learning \\
		\email{\{firstname.lastname\}@iosb.fraunhofer.de}
	}
	
	\maketitle

	\begin{abstract} 
Object detectors achieve strong performance under nominal imaging conditions but can fail silently when exposed to blur, noise, compression, adverse weather, or resolution changes. 
In safety-critical settings, it is therefore insufficient to produce predictions without assessing whether the input remains within the detector’s nominal operating regime, which we refer to as \emph{self-aware object detection}.

We introduce a degradation-aware framework based on \emph{degradation manifolds}, which structure a detector’s feature space according to image degradation rather than semantic content. 
Our method augments a standard detection backbone with a lightweight embedding head trained via multi-layer contrastive learning, where images sharing the same degradation composition are pulled together while differing configurations are separated. Unlike prior contrastive IQA approaches, our method performs detector-aligned input-fidelity monitoring: it uses task-backbone features and a nominal prototype to detect degradation-induced shifts, rather than estimating perceptual image quality in a standalone representation.

To anchor the representation, we estimate a pristine prototype from clean training embeddings, defining a nominal operating point in feature space. Self-awareness is then derived as geometric deviation from this reference, providing an intrinsic image-level signal of degradation-induced shift that is independent of detection confidence.

Experiments on synthetic corruptions, cross-dataset zero-shot transfer,
and natural weather-induced shifts show strong pristine--degraded separability, consistent behavior across detector architectures, and robust generalization across datasets with substantial content variation. These results suggest that degradation-aware detector representations provide a reliable signal for monitoring input-fidelity shifts.

Source code is available at \href{https://github.com/beckerio/degmani}{code repository}. 
\end{abstract}

	\section{Introduction}
\label{sec: intro}
Modern object detectors achieve strong performance when imaging conditions match those seen during training. However, in real-world deployments, image quality can vary substantially due to noise, blur, compression, adverse weather, or resolution changes. 
Under such degradations, detectors may fail silently: predictions can remain confident even when visual evidence has deteriorated~\cite{Quionero_book_2009,Michaelis_NeurIPSW_2019}. 
In safety-critical and unconstrained environments, it is therefore insufficient to produce predictions without assessing whether inputs remain within the detector’s nominal operating regime. We refer to this capability as \emph{self-aware object detection}.

A common strategy to estimate reliability is to rely on confidence scores or predictive uncertainty. 
While effective under mild perturbations, these signals are tied to prediction outcomes and may break down under strong degradation. 
Severely degraded inputs may yield sparse or no detections, often accompanied by high confidence in the absence of objects. Crucially, the absence of detections does not imply reliable perception.

To reason about reliability independently of prediction content, we consider a degradation-aware formulation:
\begin{equation}
	P_{\mathrm{deg}}(\mathbf{y} \mid \mathbf{x}, \mathcal{D})
	=
	P(\mathbf{y} \mid \mathbf{x}, \mathcal{D})
	\cdot
	P_{\mathrm{deg}}(\mathbf{x}),
	\label{eq:deg}
\end{equation}
where $\mathbf{y}$ denotes the detector output, $\mathbf{x}$ the input image, and $\mathcal{D}$ the training data. 
The term $P_{\mathrm{deg}}(\mathbf{x})$ captures whether the input lies within an acceptable visual regime. 
We assume no failure-labeled or unsafe samples during training, reflecting realistic deployment constraints.

Most existing approaches estimate $P_{\mathrm{deg}}(\mathbf{x})$ via out-of-distribution (OoD) detection. 
However, OoD methods transfer imperfectly to object detection and often respond to semantic novelty or dataset identity rather than image fidelity~\cite{Du_ICLR_2022,Ovadia_NeurIPS_2019,Zhang_PMLR_2021,Zhuang_ICLR_2025}. 
Consequently, clean but novel scenes may be flagged while degraded inputs remain undetected. 
In practice, many detector failures arise from progressive loss of visual fidelity rather than semantic novelty, making degradation a largely content-independent source of distribution shift.
We therefore focus on degradation-driven shift and use \emph{image degradation} as an umbrella term for corruptions, distortions, and artifacts that deviate from ideal image formation.

Recent advances in no-reference IQA show that degradation-aware representations can be learned without explicit supervision. 
Contrastive approaches such as ARNIQA~\cite{Agnolucci_WACV_2024} demonstrate that degraded views of the same image can be embedded into structured spaces encoding degradation type and severity.

Motivated by this, we equip object detectors with degradation-aware representations. 
We augment a detector backbone with a lightweight embedding head trained via multi-layer contrastive learning, pulling together images with similar degradation while separating differing degradation configurations. 
This induces a \emph{degradation manifold} within the detector’s feature space, organizing representations according to image fidelity rather than semantic content.

To anchor the manifold, we compute a pristine prototype from clean embeddings, defining a nominal operating point in representation space. 
Self-awareness is then obtained as geometric deviation from this reference, providing an intrinsic image-level signal of degradation-induced input shift that is independent of detection confidence.

Unlike prior IQA-based approaches, which target perceptual quality estimation, our method integrates degradation-aware representation learning directly into detector feature spaces and aligns it with deployment-time reliability monitoring. In this sense, the proposed degradation manifold provides detector-aligned input-fidelity monitoring using task-backbone features and a nominal prototype, rather than a standalone perceptual quality score.
We demonstrate consistent separability under synthetic corruptions, cross-dataset transfer, and natural weather shifts, suggesting that degradation-aware self-awareness is best understood as a representation learning problem for reliable perception rather than an IQA problem.

	\section{Related Work}
\label{sec:relatedwork}
\smallsec{OoD Detection and Reliability in Object Detection.}
Most approaches to estimating input validity are framed as out-of-distribution (OoD) detection.
However, most OoD methods were developed for classification, where each input yields a single global prediction.
These methods often operate on penultimate-layer features, logits, or softmax scores, including maximum softmax probability (MSP)~\cite{Hendrycks_ICLR_2017}, ODIN~\cite{Liang_ICLR_2018}, GradNorm~\cite{Huang_NeurIPS_2021}, ReAct~\cite{Sun_NeurIPS_2021}, and energy-based approaches~\cite{Liu_NeurIPS_2020}.

In object detection, predictions are defined at the detection level across many spatial locations or queries.
Consequently, OoD signals derived from logits or class scores primarily capture semantic novelty (e.g., open-set detection~\cite{Du_ICLR_2022}) rather than providing a reliable image-level assessment of imaging conditions.
Runtime monitors \cite{Kang_MLSYS_2020,Cheng_DATE_2021} verify geometric or logical consistency and address complementary system-level properties. They are orthogonal to our focus on shifts caused by degradation. Recent work also adapts OoD scoring developed for classification to detector features \cite{Zimmermann_CVPRW_2026,Becker_ICPR_2026,Kirchheim_2022_CVPR}. These methods, however, capture generic distributional deviation rather than explicitly structuring detector representations according to degradation characteristics.

\smallsec{Semantic OoD versus Degradation-Driven Shift.}
Likelihood-based models may assign low likelihood to clean images from novel scenes while assigning comparatively high likelihood to degraded inputs whose low-level statistics resemble the training distribution~\cite{Nalisnick_ICLR_2018,Kirichenko_NEURIPS_2020,Ovadia_NeurIPS_2019}. This illustrates that semantic novelty and degradation-induced fidelity shift are distinct monitoring problems. The latter is typically continuous, may preserve scene semantics, and progressively alters task-relevant visual information.

\smallsec{Generative Models for OoD Detection.}
Generative approaches estimate likelihood under a learned data distribution.
Variational autoencoders~\cite{Kingma_ICLR_2014} approximate likelihood via the ELBO, while normalizing flows (NFs)~\cite{Kobyzev_TPAMI_2021} provide exact and tractable densities.
However, image-level likelihood is known to be an unreliable OoD indicator~\cite{Nalisnick_ICLR_2018,Nalisnick_ICLR_2020,Serra_ICLR_2020}.
Recent work mitigates this by modeling multi-layer feature activations rather than raw pixels~\cite{Viviers_ECCVW_2024,Lotfi_arxi_2025}.
We adopt feature-level NFs as a baseline, but note that density modeling captures distributional similarity and does not explicitly disentangle degradation from semantic variation

\smallsec{Uncertainty Estimation in Object Detection.}
Detection-level uncertainty aims to quantify predictive uncertainty through probabilistic modeling. Exact Bayesian inference is intractable~\cite{MacKay_NC_1992}, and practical approximations include MC dropout~\cite{Gal_PMLR_2016} and variance networks~\cite{Kendall_NIPS_2017}. Variance networks parameterize predictive distributions and are trained with scoring rules such as negative log-likelihood, energy score~\cite{Gneiting_ASA_2007}, or direct moment matching~\cite{Feng_Can_We_2020_IROS}. 
While effective for modeling predictive uncertainty, these methods operate at the detection level. Image-level aggregation (e.g., top-$k$ uncertainty~\cite{Oksuz_CVPR_2023}) provides a coarse proxy but remains tied to detection outcomes rather than directly assessing input fidelity.

\smallsec{Robust Object Detection under Image Degradation.}
Robustness of detectors to common corruptions has been extensively studied, revealing significant performance drops under noise, blur, weather effects, and compression~\cite{Michaelis_NeurIPSW_2019,Yu_CVPR_2020}.
These works primarily aim to improve robustness via augmentation or architectural modifications.
In contrast, we focus on enabling detectors to assess their operating regime at test time without requiring failure supervision.

\smallsec{Image Quality Assessment.}
Image quality assessment (IQA) models degradation from the perspective of human perceptual quality.
Classical full-reference metrics such as SSIM~\cite{Wang_TIP_2004} and FSIM~\cite{Zhang_2011}, and no-reference methods such as BRISQUE~\cite{Mittal_TIP_2012} and NIQE~\cite{Mittal_SPL_2013}, estimate perceptual fidelity.
Recent learning-based approaches, including contrastive methods such as ARNIQA~\cite{Agnolucci_WACV_2024}, show that degradations induce structured geometry in feature space.

While conceptually related, IQA targets perceptual quality, whereas our setting monitors nominal imaging conditions for the detector.

	\section{Method}
\label{sec:method}
We aim to learn a \emph{degradation-aware representation} within a detector architecture that organizes images according to image degradation rather than semantic content.
To this end, we attach a lightweight embedding head to selected backbone layers and shape the feature space to encode structured degradation geometry.
Depending on the training configuration, backbone layers may optionally be fine-tuned. In our main experiments, however, we employ an auxiliary monitoring branch operating alongside a standard detector.

The resulting degradation manifold is designed to: (i) separate different degradation types and severities, (ii) map images with identical degradation characteristics consistently despite nuisance factors such as scale or resolution, and (iii) be geometrically oriented with respect to clean operating conditions to enable distance-based reliability estimation.

We adopt the degradation-aware formulation introduced earlier,
\begin{equation}
	P_{\mathrm{deg}}(\mathbf{y} \mid \mathbf{x}, \mathcal{D})
	=
	P(\mathbf{y} \mid \mathbf{x}, \mathcal{D})
	\cdot
	P_{\mathrm{deg}}(\mathbf{x}),
\end{equation}
and approximate $P_{\mathrm{deg}}(\mathbf{x})$ by a geometry-derived score $S_{\mathrm{deg}}(\mathbf{x})$ computed as a distance in a learned embedding space. The score serves as a geometry-derived indicator of deviation from nominal imaging conditions without explicit density estimation or likelihood modeling.

At deployment, a monitoring decision is obtained via
\begin{equation}
	\hat{r}(\mathbf{x})
	=
	\mathbb{I}\big[S_{\mathrm{deg}}(\mathbf{x}) \le \tau\big],
\end{equation}
where $\tau$ defines the acceptable degradation regime. Operationally, $\tau$ is calibrated from pristine validation scores for a target nominal false-positive rate and then fixed within a deployment domain. Changes in camera characteristics may require recalibration. An overview of the framework is shown in Fig.~\ref{fig:or_cfd}. Implementation and training details are provided in the Supplementary Material.
\begin{figure*}[t]
	\centering
	\includegraphics[width=.8\textwidth]{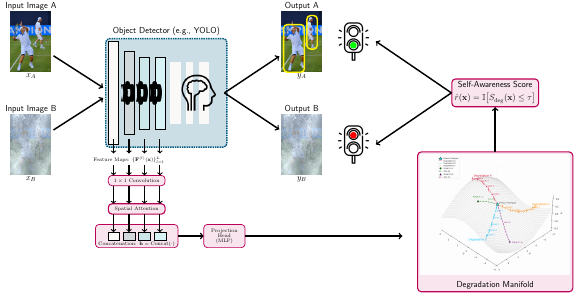}
	\caption{Proposed degradation-manifold framework for self-awareness.
		Multi-layer backbone features are fused via $1{\times}1$ projections and
		attention pooling and mapped into a normalized embedding space trained with
		contrastive degradation compositions.
		A pristine prototype, computed from clean training images, anchors the
		manifold to nominal operating conditions.
		At inference, cosine distance to this prototype yields the image-level
		degradation score $S_{\mathrm{deg}}(\mathbf{x})$.}
	\label{fig:or_cfd}
\end{figure*}

\subsection{Multi-Layer Degradation Representation}
We extract feature maps from multiple backbone stages,
\[
\{\mathbf{F}^{(l)}(\mathbf{x})\}_{l=1}^{L},
\]
capturing complementary degradation cues across depths, as low-level corruptions affect early features while severe degradations propagate into higher-level representations~\cite{Wang_MTA_2020,Bianco_PRL_2021}.

Each feature map is reduced via a $1\times1$ convolution and pooled using a learnable attention-based operator, producing vectors $\mathbf{a}^{(l)} \in \mathbb{R}^{d}$.
The per-layer vectors are concatenated to form a multi-scale descriptor
\begin{equation}
	\mathbf{h} = \mathrm{Concat}(\mathbf{a}^{(1)}, \dots, \mathbf{a}^{(L)}),
\end{equation}
capturing complementary degradation cues across depths.

The fused representation is projected into a low-dimensional embedding space via a lightweight MLP,
\begin{equation}
	\mathbf{z} = g(\mathbf{h}) \in \mathbb{R}^{D},
\end{equation}
followed by $\ell_2$ normalization.

\subsection{Contrastive Degradation Manifold Learning}
To organize embeddings according to degradation composition and severity, we adopt a SimCLR-style contrastive objective \cite{chen_ICML_2020}, inspired by the degradation chaining strategy of ARNIQA \cite{Agnolucci_WACV_2024}.

For each clean image $\mathbf{x}$, we generate degraded views by sampling a composition
\[
C_i = \{d_1, \dots, d_k\},
\]
where each $d_j$ denotes a degradation operator (e.g., blur, noise, compression) applied with randomly sampled parameters. Sequential compositions produce structured and progressively stronger perturbations without requiring degradation labels.

Two different pristine images degraded with the same composition $C_i$ 
form a positive pair $(z_{i,A}, z_{i,B})$. To enforce finer separation, we construct hard negatives via resolution perturbation: degraded images are downscaled and resized, preserving semantics while introducing additional fidelity loss.

For a batch of size $N_B$, we optimize the NT-Xent objective
\begin{equation}
	\mathcal{L}_{\text{deg}}
	=
	\frac{1}{4N_B}
	\sum_{i=1}^{N_B}
	\Bigl(
	\ell(z_{i,A}, z_{i,B})
	+
	\ell(z_{i,B}, z_{i,A})
	+
	\ell(z_{i,\tilde{A}}, z_{i,\tilde{B}})
	+
	\ell(z_{i,\tilde{B}}, z_{i,\tilde{A}})
	\Bigr),
\end{equation}
where $\ell(\cdot,\cdot)$ denotes the standard NT-Xent loss with temperature $\tau_c$, and negatives are drawn from all remaining embeddings in the batch. 

Unlike contrastive IQA methods such as ARNIQA, which learn quality-sensitive representations for perceptual estimation, our method learns degradation geometry directly in detector representations for deployment-time reliability monitoring.

\subsection{Pristine Prototype and Degradation Score}
To orient the manifold with respect to clean operating conditions, we maintain a pristine prototype $\boldsymbol{\mu}_{\text{pristine}}$ computed from embeddings of pristine images.

The prototype is initialized after a warm-up phase (half of the total training epochs) and updated via an exponential moving average:
\begin{equation}
	\boldsymbol{\mu}_{\text{pristine}}
	\leftarrow
	\alpha \boldsymbol{\mu}_{\text{pristine}}
	+
	(1-\alpha)\,\mathbb{E}[\mathbf{z}_{\text{pristine}}],
\end{equation}
with momentum $\alpha \in [0,1)$ and without gradient propagation.

At inference time, we define the degradation score as the cosine distance to the pristine prototype,
$S_{\mathrm{deg}}(\mathbf{x}) = 1 - \mathbf{z}(\mathbf{x})^\top \boldsymbol{\mu}_{\text{pristine}}$,
where the equality follows from $\ell_2$ normalization and larger values indicate increasing deviation from nominal imaging conditions.

\subsection{Auxiliary Monitoring Branch}
The degradation manifold can in principle be optimized jointly with the detection objective.
However, contrastive degradation learning encourages sensitivity to fidelity changes, whereas detection training promotes invariance to nuisance variation.

In our main experiments, we adopt an auxiliary two-path configuration, where the degradation head operates alongside a standard detector without modifying its primary objective. This preserves detection performance while enabling degradation-aware monitoring. Joint training results are provided in the Supplementary Material.

	\section{Evaluation}
\label{sec:eval}
We evaluate self-aware detection by analyzing whether degradation-induced covariate shift can be detected reliably at the image level and how this relates to task performance under shift. Our goal is not to predict per-instance correctness, but to identify deviations from nominal operating conditions. Following standard OoD and robustness evaluation, we measure \emph{pristine--degraded separability} using AUROC and report degradation-induced drops in detection accuracy (mAP) for context.

\subsection{Dataset and Degradation Benchmark}
Our primary benchmark is COCO~\cite{Lin_ECCV_2014}. The training split is used to train detectors and baseline monitors and to compute pristine reference prototypes; the unaltered validation split serves as the in-distribution (ID) reference.

To simulate degradation-induced covariate shift in $p(\mathbf{x})$~\cite{Quionero_book_2009}, we apply synthetic degradations from the robustness suite of~\cite{Michaelis_NeurIPSW_2019}, based on~\cite{Hendrycks_ICLR_2019}. Each corruption is evaluated at five severity levels, yielding progressively degraded splits with unchanged labels.

Evaluation degradations are \emph{not identical} to those used during contrastive training. Training uses degradation compositions inspired by~\cite{Agnolucci_WACV_2024}, whereas evaluation follows robustness-style corruptions from~\cite{Hendrycks_ICLR_2019,Michaelis_NeurIPSW_2019}. Results therefore reflect generalization beyond the exact transformations observed during training.

For mixed-corruption evaluation, we use a round-robin assignment scheme such that each image is consistently associated with a specific corruption across severity levels.

\subsection{Models and Compared Monitors}
\smallsec{Detector Backbones and Degradation Manifold.}
Our degradation manifold is implemented as a lightweight embedding head attached to the feature hierarchy of a standard object detector. To assess detector dependence, we instantiate the method on multiple COCO-pretrained backbones: YOLOv9~\cite{Wang_ECCV_2024}, YOLOv10~\cite{Wang_NeurIPS_2024}, YOLOv11~\cite{yolo11_ultralytics}, and RT-DETR~\cite{Zhao_2024_CVPR}. Details are provided in the Supplementary Material. At inference, self-awareness is computed as cosine distance between the image embedding and a pristine prototype estimated from clean COCO training images.

\smallsec{Probabilistic Detectors.}
We compare against image-level uncertainty derived from probabilistic detectors using top-$k$ aggregation~\cite{Oksuz_CVPR_2023}. These detectors are realized as variance networks that output multivariate Gaussian predictive distributions~\cite{Harakeh_NeurIPS_2021} and are trained with negative log-likelihood (NLL), energy score (ES)~\cite{Gneiting_ASA_2007}, and direct moment matching (DMM)~\cite{Feng_Can_We_2020_IROS}. Monitoring scores are derived from confidence, classification entropy, and determinant, trace, or entropy of Gaussian box distributions~\cite{Murphy_book1_2022}.

\smallsec{Generative Modeling.}
We include a likelihood-based baseline using normalizing flows (NFs) trained on detector features (e.g., RealNVP~\cite{Dinh_ICLR_2017}), following feature-space OoD modeling~\cite{lotfi2025medoodflow}. We use a single NF on concatenated multi-layer features and a multi-scale NF (M-NF) with one NF per layer. Following~\cite{lotfi2025medoodflow}, features are extracted from a ResNet backbone~\cite{He_CVPR_2016} and globally average pooled. The monitoring score is $-\log p_{\mathrm{deg/ID}}(\mathbf{x})$.

\smallsec{Feature-space OoD Baselines.}
We additionally evaluate adapted OoD baselines on detector feature
representations. Specifically, we consider GRAM~\cite{Sastry_ICML_2020},
Mahalanobis~\cite{Lee_NEURIPS_2018}, KNN~\cite{Sun_ICML_2022}, VIM~\cite{Wang_2022_CVPR},
and activation histograms~\cite{Becker_ICPR_2026}. These baselines test whether generic
feature-space OoD scoring suffices for degradation monitoring~\cite{Becker_ICPR_2026,Zimmermann_CVPRW_2026}.

\smallsec{Image Quality Assessment Baselines.}
We include no-reference IQA methods as baselines. We evaluate them both via
predicted quality scores and via cosine distance between their learned embeddings and a pristine prototype computed from clean COCO training images~\cite{Snell_NIPS_2017,Mensink_TPAMI_2013}. All IQA models are evaluated
zero-shot on degraded COCO images.

\subsection{Metrics}
\smallsec{Self-Awareness under Degradation.}
We quantify monitor quality using the \emph{area under the receiver operating characteristic} (AUROC) between pristine validation images and degraded variants. Degraded samples are treated as the positive class, and all scores are oriented such that larger values indicate stronger deviation from pristine conditions. Scores include negative log-likelihood $-\log p_{\mathrm{deg/ID}}(x)$ for NFs, detector-derived signals, IQA scores, and cosine distance in embedding space. 
Final AUROC values are computed after z-score normalization~\cite{Viviers_ECCVW_2024}. Operating-point metrics are reported in the Supplementary Material.

\smallsec{Detection Performance under Degradation.}
To contextualize monitor behavior, we report \emph{mean average precision} (mAP@.5:.95 and mAP@0.5) under increasing severity.

\subsection{Results}
\subsubsection{Detector Performance.}
Fig.~\ref{fig:map_vs_severity_hendrycks} summarizes detection performance under synthetic degradation from two complementary perspectives. The left panel shows mAP degradation across multiple detector architectures on a mixed-corruption benchmark. For YOLOv10-m~\cite{Wang_NeurIPS_2024}, the right panel shows a per-corruption
analysis using robustness-style degradations from~\cite{Michaelis_NeurIPSW_2019,Hendrycks_ICLR_2019}.

Across both views, detection performance degrades systematically with increasing severity. The mixed benchmark shows a consistent trend across detectors, while the per-corruption analysis reveals heterogeneous drops across degradation types, consistent with prior robustness studies~\cite{Oksuz_CVPR_2023,Michaelis_NeurIPSW_2019,Liu_IJCV_2024}. Additional results are provided in the Supplementary Material.

\renewcommand{\subw}{0.45\textwidth}
\begin{figure*}[h!]
	\captionsetup[subfigure]{justification=centering}
	\centering
	\begin{subfigure}{\subw}
		\centering
		\includegraphics[width=\textwidth]{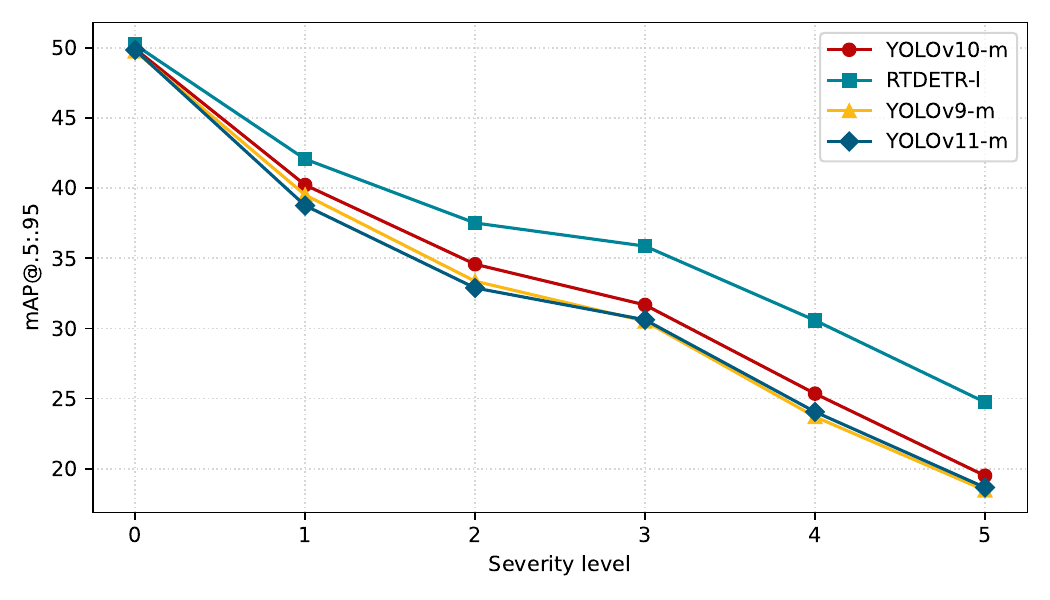}
		\caption*{Mixed-corruption benchmark}
	\end{subfigure}
	\hfill
	\begin{subfigure}{\subw}
		\centering
		\includegraphics[width=\textwidth]{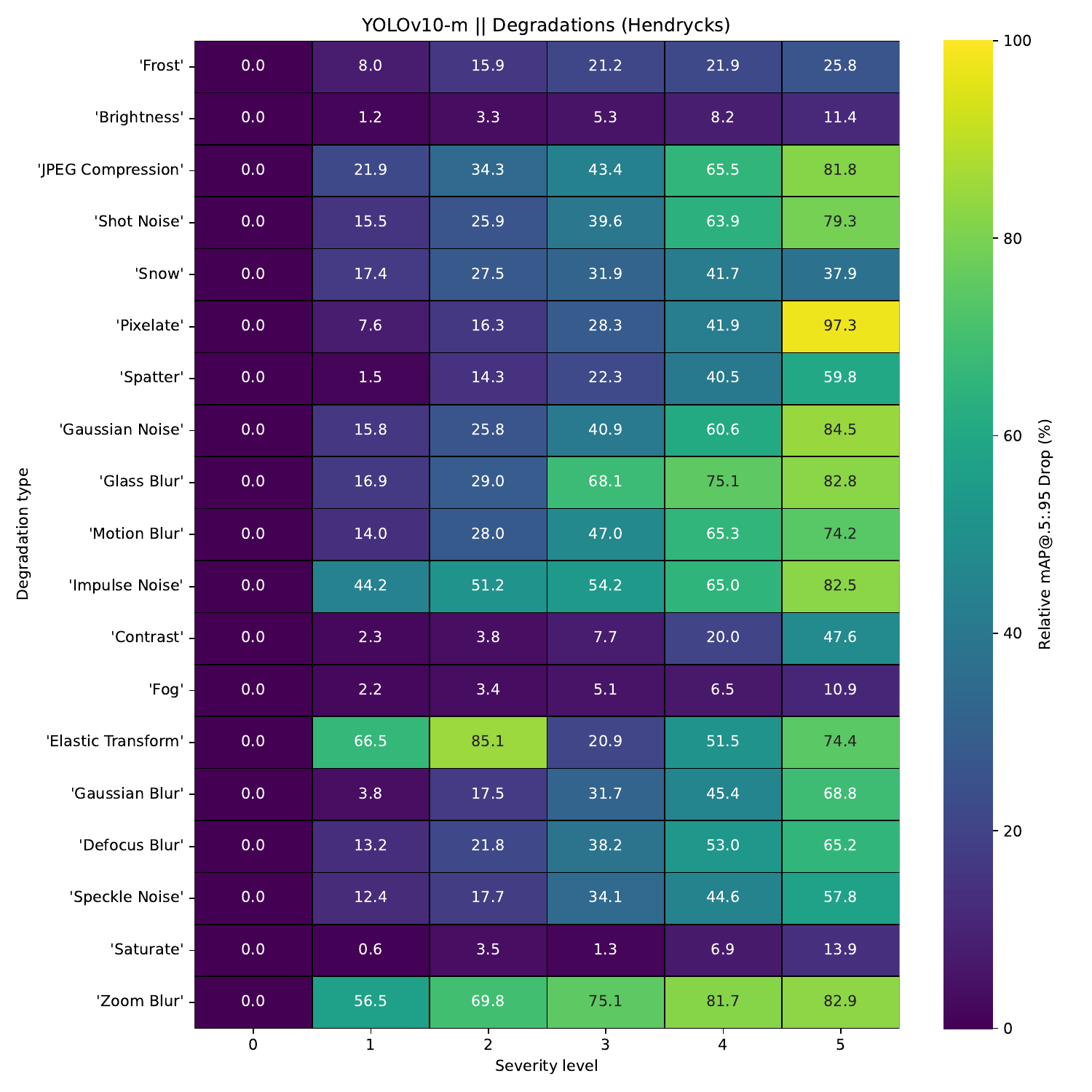}
		\caption*{Per-corruption analysis from~\cite{Hendrycks_ICLR_2019}}
	\end{subfigure}
	\caption{
		Detection performance under synthetic degradation from two complementary perspectives. Left: mAP@.5:.95 of multiple detectors on the mixed-corruption COCO \texttt{val2017} benchmark under increasing severity. Right: relative mAP@0.5:0.95 drop of YOLOv10-m~\cite{Wang_NeurIPS_2024} on COCO \texttt{val2017} for individual robustness-style degradations from~\cite{Michaelis_NeurIPSW_2019,Hendrycks_ICLR_2019}. Each cell shows the percentage drop relative to the clean baseline; brighter cells indicate larger drops.
	}
	\label{fig:map_vs_severity_hendrycks}
	\vspace{-0.5cm}
\end{figure*}

\subsubsection{Degradation Manifold Analysis.}
\begin{figure*}
	\centering
	\begin{subfigure}{0.8\linewidth}
		\includegraphics[width=\linewidth]{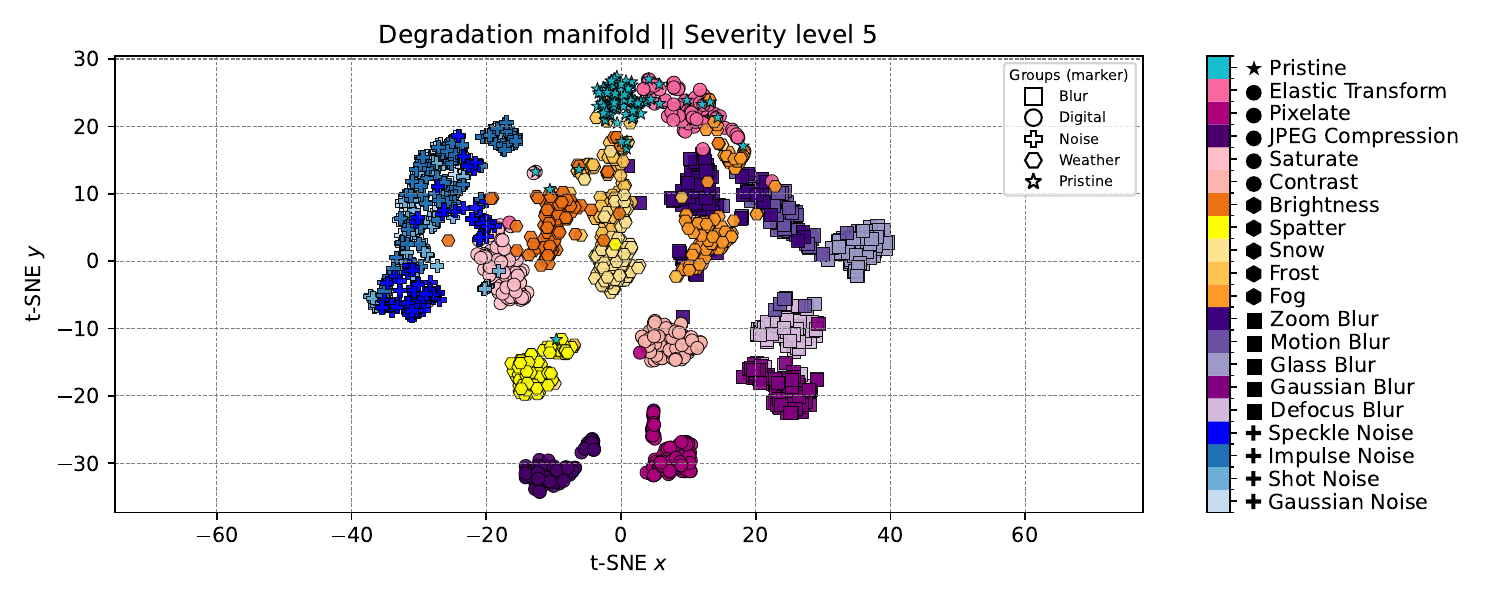}
		\caption*{}
	\end{subfigure}
	\vspace{-0.5cm}
	\caption{
		t-SNE visualization~\cite{Hinton_Roweis_2003} of the learned degradation manifold for robustness-style corruptions from~\cite{Michaelis_NeurIPSW_2019}, based on~\cite{Hendrycks_ICLR_2019}. Colors denote degradation types and marker styles denote degradation groups. For visualization, 100 COCO~\cite{Lin_ECCV_2014} samples are corrupted at severity level~5. Training uses degradation compositions from~\cite{Agnolucci_WACV_2024}, indicating zero-shot generalization of the learned geometry across corruption families.
	}
	\label{fig:embeddding_hendrycks_main}
	\vspace{-0.5cm}
\end{figure*}

Fig.~\ref{fig:embeddding_hendrycks_main} visualizes the learned embedding space using t-SNE~\cite{Hinton_Roweis_2003} for severity-5 robustness-style corruptions from~\cite{Hendrycks_ICLR_2019}. Degraded samples form clearly separated clusters by degradation type despite the absence of degradation labels during training, indicating that the representation is organized primarily by degradation characteristics rather than semantic content. Additional t-SNE visualizations for other severity levels, IQA-style degradations, and cross-dataset results under substantial content variation are provided in the Supplementary Material.

\subsubsection{Self-Awareness under Degradation.}
Table~\ref{tab:auroc_degradation_manifold_vs_all} reports AUROC for pristine--degraded separability on COCO under mixed corruptions across five severity levels. Corresponding operating-point metrics (FPR@95TPR and FNR@95TNR) are provided in the Supplementary Material.

The proposed degradation manifold achieves the strongest separability across all detector backbones, reaching $97.14$ AUROC at severity~5 while remaining above $88$ at severity~1.

Compared to probabilistic detector uncertainties, feature-density modeling, adapted feature-space OoD baselines, and IQA-based monitors, our method yields a clear margin across all severity levels. Among the adapted OoD baselines, GRAM and activation histograms perform strongest, but remain well below the proposed manifold, indicating that generic feature-space OoD scoring captures part of the degradation-induced shift without recovering the degradation-aware geometry induced by our training objective.

Among IQA baselines, ARNIQA embeddings transfer best, whereas CLIP-based IQA embeddings perform substantially worse, suggesting representations dominated more by semantic alignment than degradation geometry. Among detector-derived uncertainty measures, raw confidence performs best (up to $77.65$ at severity~5) but depends on stable object hypotheses. Normalizing flows improve mainly under stronger corruption and remain below $70$ AUROC at severity~5.

Overall, explicitly structuring degradation geometry within task-aligned representations provides a stronger signal for degradation-induced covariate shift than generic OoD scoring, output uncertainty, density modeling, or perceptual quality estimation.

\begin{table*}[h!]
	\aurocTableSetup
		\begin{tabular}{l ccccc}
			\toprule
			& \multicolumn{5}{c}{\underline{Severity levels AUROC $\uparrow$}} \\
			Model
			& 1 & 2 & 3 & 4 & 5 \\
			\midrule
			{\textbf{Image Quality Assessment (IQA)}}\\			
			\midrule    
			MANIQA \cite{yang2022maniqa} (score-based)                  & 52.88 & 54.49 & 55.27 & 56.67 & 61.56 \\
			MANIQA \cite{yang2022maniqa} (embedding-based; prototype anchored)         & 65.87 & 71.55 & 73.42 & 76.97 & 77.30 \\
			QualiCLIP \cite{Agnolucci_arxiv_2025} (score-based)       & 53.17 & 56.77 & 66.28 & 70.60 & 72.53 \\            
			QualiCLIP \cite{Agnolucci_arxiv_2025} (embedding-based; prototype anchored)  & 45.36 & 41.53 & 46.77 & 49.79 & 56.16 \\
			QualiCLIP+ \cite{Agnolucci_arxiv_2025} (score-based)      & 52.56 & 57.90 & 65.49 & 69.18 & 70.87 \\
			QualiCLIP+ \cite{Agnolucci_arxiv_2025} (embedding-based; prototype anchored)  & 47.04 & 42.45 & 45.31 & 50.23 & 56.24 \\
			CLIPIQA \cite{Wang_AAAI_2023} (score-based)              & 57.78 & 60.43 & 62.29 & 69.28 & 69.71 \\        
			CLIPIQA \cite{Wang_AAAI_2023} (embedding-based; prototype anchored)         & 49.36 & 47.71 & 47.58 & 46.65 & 45.76 \\
			CLIPIQA+ \cite{Wang_AAAI_2023} (score-based)             & 56.42 & 58.11 & 61.03 & 63.45 & 67.81 \\
			CLIPIQA+ \cite{Wang_AAAI_2023} (embedding-based; prototype anchored)        & 49.85 & 48.65 & 47.57 & 46.61 & 45.33 \\
			ARNIQA \cite{Agnolucci_WACV_2024} (score-based)           & 52.09 & 52.83 & 58.26 & 58.43 & 60.23 \\            
			ARNIQA \cite{Agnolucci_WACV_2024} (embedding-based; prototype anchored)     & 73.60 & 80.07 & 80.58 & 83.30 & 85.74 \\			
			\midrule
			{\textbf{Top-$3$ Detection Uncertainties} (\cite{Oksuz_CVPR_2023,Harakeh_NeurIPS_2021})}\\
			\midrule
			Trace (RetinaNet ES)                 & 52.61 & 55.31 & 57.14 & 60.29 & 65.44 \\
			Trace (DETR NLL)                     & 51.14 & 54.83 & 58.96 & 59.68 & 63.61 \\
			Determinant (RetinaNet NLL)          & 49.48 & 51.09 & 50.94 & 54.13 & 58.23 \\
			Determinant (RetinaNet ES)           & 49.65 & 50.57 & 51.29 & 52.66 & 56.39 \\
			Entropy (Regression) (RetinaNet ES)  & 52.73 & 55.15 & 57.67 & 60.82 & 66.51 \\
			Entropy (Regression) (FasterRCNN NLL) & 51.65 & 51.66 & 55.72 & 56.90 & 64.87 \\
			Confidence Score (RetinaNet NLL)     & 52.91 & 56.93 & 61.51 & 69.69 & 77.65 \\
			Confidence Score (RetinaNet ES)      & 52.55 & 56.77 & 60.72 & 68.10 & 75.28 \\
			Entropy (Classification) (FasterRCNN ES)  & 54.70 & 59.97 & 62.84 & 68.47 & 77.60 \\
			Entropy (Classification) (FasterRCNN NLL) & 53.31 & 59.65 & 64.35 & 70.43 & 77.33 \\			
			\midrule
			{\textbf{Feature-space OoD methods}} (\cite{Zimmermann_CVPRW_2026,Becker_ICPR_2026}) \\			
			\midrule    
			GRAM \cite{Sastry_ICML_2020} (YOLOv10-m feature maps) & 63.92 & 71.86 & 77.81 & 83.29 & 85.88 \\
			Mahalanobis \cite{Lee_NEURIPS_2018} (YOLOv10-m feature maps) & 62.92 & 68.08 & 70.34 & 73.81 & 74.59 \\
			KNN \cite{Sun_ICML_2022} (YOLOv10-m feature maps) & 65.88 & 70.50 & 72.06 & 76.31 & 78.91 \\
			VIM \cite{Wang_2022_CVPR} (YOLOv10-m feature maps)  & 63.32 & 68.09 & 70.00 & 73.63 & 75.19 \\
			Activation histograms \cite{Becker_ICPR_2026} (YOLOv10-m feature maps) & 64.42 & 72.62 & 77.75 & 83.82 & 86.24 \\			
			\midrule       
			\textbf{Normalizing Flows} (adapted from \cite{lotfi2025medoodflow})\\
			\midrule                
			NF pooled feature maps (C4, C5)            
			& 49.60 & 49.90 & 51.12 & 56.20 & 61.53\\
			NF pooled feature maps (C1, C2, C3)        
			& 49.70 & 50.10 & 55.12 & 64.40 & 73.56\\
			NF pooled feature maps (all; C1--C5)       
			& 49.80 & 50.20 & 56.12 & 64.90 & 73.45\\            
			M-NF pooled feature maps (C4, C5)          
			& 49.70 & 50.10 & 55.76 & 61.60 & 67.54\\
			M-NF pooled feature maps (C1, C2, C3)      
			& 49.80 & 50.20 & 58.61 & 63.80 & 68.94\\
			M-NF pooled feature maps (all; C1--C5)     
			& 49.90 & 50.25 & 58.76 & 63.90 & 69.12\\
			\midrule
			{ \textcolor{Fraunhoferred}{\textbf{Degradation Manifold} DM (Ours)} }\\  
			\midrule			
			DM (YOLOv10-m)  & \colorbox{colorFst}{88.64} & \colorbox{colorSnd}{89.70}  & \colorbox{colorTrd}{89.75}  & \colorbox{colorSnd}{95.28} & \colorbox{colorFst}{97.14}  \\  
			DM (RT-DETR-l)  & 83.88 & 88.29 & \colorbox{colorSnd}{90.37} & \colorbox{colorFst}{95.61} & \colorbox{colorTrd}{97.11} \\
			DM (YOLOv9-m)   & \colorbox{colorSnd}{87.64} & 89.06 & \colorbox{colorFst}{90.07} & \colorbox{colorTrd}{95.66} & 96.62 \\
			DM (YOLOv11-m)  & \colorbox{colorTrd}{85.84} & \colorbox{colorFst}{89.12} & 89.95 & 94.58 & \colorbox{colorSnd}{96.04} \\            
			\bottomrule
		\end{tabular}
	\caption{AUROC scores of detecting degradation-induced covariate shift for synthetically corrupted variants of the COCO dataset. Best results are highlighted as \colorbox{colorFst}{first}, \colorbox{colorSnd}{second}, and \colorbox{colorTrd}{third}.}
	\label{tab:auroc_degradation_manifold_vs_all}
	\vspace{-.5cm}
\end{table*}
Beyond the synthetic COCO benchmark, we evaluate the degradation manifold in cross-dataset zero-shot transfer and natural weather-induced distribution shift.

\smallsec{Cross-Dataset Evaluation.}
The degradation manifold is trained on COCO \cite{Lin_ECCV_2014}. We evaluate it without adaptation on KITTI \cite{Geiger_CVPR_2012}, VISDRONE \cite{zhu2021detection}, DETRAC \cite{Wen_CVIU_2020}, UAVDT \cite{Du_ECCV_2018}, and FLIR (VIS) \cite{flirDataset}. Table~\ref{tab:auroc_per_level_datasets} summarizes the cross-dataset results. The method maintains strong pristine--degraded separability despite substantial variation in scene layout, camera setup, and object scale. This indicates that the learned representation captures degradation structure rather than dataset-specific semantics. Mixed-dataset results are provided in the Supplementary Material.

\begin{table*}[h!]
	\centering
	\aurocTableSetup
	\begin{tabular}{l ccccc} 
		\toprule			
		\multicolumn{6}{l}{\textcolor{Fraunhoferred}{\textbf{Degradation Manifold (Ours)}}} \\
		\midrule
		& \multicolumn{5}{c}{\underline{Severity levels AUROC $\uparrow$}} \\
		Dataset & 1 & 2 & 3 & 4 & 5\\	
		\midrule
		VISDRONE \cite{zhu2021detection}  & 83.66 & 89.15 & 92.08 & 95.85 & 96.79 \\		
		KITTI \cite{Geiger_CVPR_2012}     & 94.19 & 96.37 & 97.16 & 97.25 & 98.52 \\		
		DETRAC~\cite{Wen_CVIU_2020}       & 94.65 & 97.05 & 98.87 & 99.69 & 99.81 \\	
		FLIR (VIS) \cite{flirDataset}     & 82.83 & 91.39 & 92.72 & 95.11 & 96.66 \\	
		UAVDT~\cite{Du_ECCV_2018}         & 92.12 & 94.71 & 95.56 & 98.78 & 98.83\\ 
		\bottomrule 
	\end{tabular}
	\caption{AUROC scores for detecting degradation-induced covariate shift using our degradation manifold across different datasets.}
	\label{tab:auroc_per_level_datasets}
	\vspace{-0.5cm}
\end{table*}

\smallsec{Natural Weather Shift.}
To assess real-world distribution shift, we evaluate on weather-affected subsets of Seeing Through Fog (STF)~\cite{Bijelic_2020_STF}, BDD~\cite{Yu_CVPR_2020}, and DSERT-RoLL~\cite{Cho_CVPR_2026}. Weather metadata is used to construct an ID (nominal) vs. OoD (adverse-weather) split. We retain only conditions with strong visible weather effects and focus on heavy-weather subsets for DSERT-RoLL, where light conditions induce a more ambiguous ID/OoD boundary. Compared to synthetic corruptions, separability under natural weather shift is weaker but remains clearly above chance. Training with synthetic weather corruptions~\cite{Hendrycks_ICLR_2019} consistently improves transfer to \texttt{heavy snow}, \texttt{dense fog}, and \texttt{heavy rain}.

\begin{table}[h!]
	\centering
	\aurocTableSetup
	\begin{tabular}{l l c c}
		\toprule
		\multicolumn{4}{l}{\textcolor{Fraunhoferred}{\textbf{Degradation Manifold (Ours)}}} \\
		\midrule
		Dataset & Subset (OoD) & Weather aug. (train) & AUROC $\uparrow$ \\
		\midrule
		STF~\cite{Bijelic_2020_STF}     & \texttt{heavy snow} & \xmark & 68.85 \\
		STF~\cite{Bijelic_2020_STF}     & \texttt{heavy snow} & \cmark & 71.94 \\
		STF~\cite{Bijelic_2020_STF}     & \texttt{dense fog}  & \xmark & 77.37 \\
		STF~\cite{Bijelic_2020_STF}     & \texttt{dense fog}  & \cmark & 81.50 \\
		STF~\cite{Bijelic_2020_STF}     & \texttt{heavy rain} & \xmark & 78.84 \\
		STF~\cite{Bijelic_2020_STF}     & \texttt{heavy rain} & \cmark & 86.66 \\
		BDD~\cite{Yu_CVPR_2020}         & \texttt{heavy rain} & \xmark & 64.96 \\
		BDD~\cite{Yu_CVPR_2020}         & \texttt{heavy rain} & \cmark & 88.52 \\
		DSERT-RoLL~\cite{Cho_CVPR_2026} & \texttt{heavy snow} & \xmark & 85.32 \\
		DSERT-RoLL~\cite{Cho_CVPR_2026} & \texttt{heavy snow} & \cmark & 95.31 \\
		DSERT-RoLL~\cite{Cho_CVPR_2026} & \texttt{heavy rain} & \xmark & 64.20 \\
		DSERT-RoLL~\cite{Cho_CVPR_2026} & \texttt{heavy rain} & \cmark & 81.33 \\
		DSERT-RoLL~\cite{Cho_CVPR_2026} & \texttt{dense fog}  & \xmark & 91.04 \\
		DSERT-RoLL~\cite{Cho_CVPR_2026} & \texttt{dense fog}  & \cmark & 97.30 \\
		\bottomrule
	\end{tabular}
	\caption{Natural weather-induced distribution shifts on STF~\cite{Bijelic_2020_STF}, BDD~\cite{Yu_CVPR_2020}, and DSERT-RoLL~\cite{Cho_CVPR_2026}. AUROC measures separability between nominal-weather (ID) and adverse-weather (OoD) subsets. Training with synthetic weather corruptions improves transfer to real-world heavy weather conditions.}
	\label{tab:auroc_naturalshifts_dataset_weather}
	\vspace{-0.5cm}
\end{table}
\smallsec{Single-Degradation Analysis.}
Per-corruption AUROC across severity levels is reported in the Supplementary Material. Although the manifold is trained using IQA-style degradations~\cite{Agnolucci_WACV_2024}, separability increases with corruption strength for most robustness corruptions~\cite{Hendrycks_ICLR_2019}, indicating structured degradation geometry beyond the exact training transformations. 

\smallsec{Ablation Study.}
We ablate the main architectural and training choices of the proposed degradation
manifold: multi-layer feature aggregation, attention-based pooling, and hard-negative
mining. The full configuration yields the best pristine--degraded separability across
all severity levels, indicating that degradation monitoring benefits from combining
hierarchical detector features with a pooling mechanism that emphasizes localized
fidelity cues. Removing hard negatives leads to the most pronounced degradation at high
severity, suggesting that fine-grained contrast between related fidelity regimes is
important for shaping a useful monitoring geometry. Full quantitative results are
reported in the Supplementary Material. 

	\section{Discussion and Limitations}
\label{sec:limitations}

\smallsec{Data-side Self-Awareness.}
We formulate self-awareness through an explicit \emph{data-side} degradation term (Eq.~\ref{eq:deg}) that separates prediction confidence from input fidelity. The learned score $S_{\mathrm{deg}}(\mathbf{x})$ is not intended as a calibrated predictor of task failure, but as an operational warning signal for deviation from nominal imaging conditions. This reflects realistic deployment settings, where exhaustive supervision of failure cases under shift is unavailable. Accordingly, we evaluate monitors via pristine--degraded separability and treat the monitor as a gating mechanism rather than a replacement for task-level evaluation. The learned degradation manifold exhibits clear structure across severity levels and degradation types. This enables flexible operating points and can support targeted downstream responses such as input re-acquisition, adaptive preprocessing, or model switching. Thus, even without directly predicting task performance, reliable separation provides actionable system-level information.

\smallsec{Alternative Monitors.}
Signals derived from detector outputs reflect prediction uncertainty rather than input fidelity.
They depend on the presence and stability of object hypotheses and implicitly assume detections in the scene.
Under strong degradations, detections may disappear or become unstable, limiting the reliability of image-level aggregation.
Performance further depends on the training objective (ES, NLL, DMM), with \emph{energy score} variants yielding more balanced behavior.
Although raw confidence achieves the strongest AUROC among detector-based signals, it remains below the proposed manifold and inherits the assumption of stable object hypotheses.

Normalizing-flow baselines aim to model feature likelihood under nominal conditions.
However, density estimation in high-dimensional detector feature spaces at realistic resolutions typically requires strong compression (e.g., global pooling), which limits expressiveness.
Moreover, likelihood models are known to exhibit counterintuitive behavior under distribution shift~\cite{Nalisnick_ICLR_2018,Kirichenko_NEURIPS_2020}.
In contrast, our approach structures representation geometry directly through contrastive learning and avoids explicit density modeling, leading to more stable separability and improved transfer across datasets.

Modern no-reference IQA models provide a relevant comparison because they learn degradation-sensitive representations. Empirically, ARNIQA embeddings transfer comparatively well, whereas CLIP-based IQA embeddings remain weaker, suggesting that semantic alignment does not necessarily yield degradation-oriented geometry.

A similar limitation holds for adapted feature-space OoD baselines: although they respond to distributional deviation in detector features, they do not explicitly organize representations by image fidelity and therefore remain less effective under degradation-driven shift.

\smallsec{Limitations and Outlook.}
The proposed manifold measures deviation from nominal imaging conditions but
does not directly estimate task correctness. As with other shift monitors,
degradation distance does not guarantee performance degradation on a
per-instance basis: some degraded images may remain detectable, while failures
may also occur under nominal conditions due to semantic or contextual factors.

In addition, our degradation sampling focuses on image-formation effects
(noise, blur, compression, resolution, and related distortions) and does not
explicitly address semantic shifts, domain changes, or label noise.

Changing the detector architecture or selected feature representations may
require offline retraining of the lightweight monitoring branch, although the
detector itself does not need to be retrained in our auxiliary configuration.
Furthermore, both the detector and the proposed monitor operate frame-wise;
exploiting temporal consistency in video is complementary and requires a
dedicated sequence-based evaluation.

Finally, design choices such as multi-layer readout, pooling, and hard-negative
construction influence the resulting geometry. While our ablations validate key
components, more principled fusion strategies may further improve robustness
under complex or mixed degradations. A detailed analysis of the geometric
behavior of the degradation manifold under increasing severity is provided in the Supplementary Material.
	\section{Conclusion}
\label{sec: conclusion}
We presented a degradation-aware framework for self-aware object detection that structures detector representations according to image fidelity rather than semantic content. By augmenting standard detector backbones with a lightweight contrastive embedding head, we learn a degradation manifold that enables intrinsic image-level reliability estimation via geometric distance to a pristine reference.

Across synthetic corruptions, cross-dataset transfer, and natural weather shifts, the proposed approach achieves strong pristine--degraded separability and consistently outperforms detector-derived uncertainty, feature-density modeling, and IQA baselines.

These results suggest that degradation-aware representation geometry provides a practical foundation for self-aware perception across multiple detector architectures operating under real-world visual variability. Future work may further explore tighter integration with task-level reliability estimation and broader forms of distribution shift.

	\clearpage
	\appendix
	\section*{Supplementary Material}
	\begin{abstract} 
This supplementary material provides additional implementation details, extended experimental results, and supporting analyses for the main paper. In particular, it includes detector configurations, dataset construction details for natural weather shift, additional evaluation results, operating-point metrics, a joint detection--degradation training study, and further analysis of the learned degradation geometry.
\end{abstract}

\section{Detector Architectures}
\label{app:Detectors}
To assess behavior across detector architectures, we instantiate the proposed degradation manifold on multiple detector architectures and evaluate both monitoring performance and degradation-induced detection loss under identical corruption settings. We consider YOLOv9~\cite{Wang_ECCV_2024}, YOLOv10~\cite{Wang_NeurIPS_2024}, YOLOv11~\cite{yolo11_ultralytics}, and RT-DETR~\cite{Zhao_2024_CVPR} as representative one-stage and transformer-based detectors.

\paragraph{YOLOv9.}
YOLOv9~\cite{Wang_ECCV_2024} is a one-stage detector that revisits information preservation in deep architectures. It introduces \emph{Programmable Gradient Information} (PGI), an auxiliary supervision scheme with additional training branches designed to mitigate error accumulation as network depth increases. YOLOv9 further proposes the \emph{Generalized Efficient Layer Aggregation Network} (GELAN), a flexible backbone that supports different computational blocks while maintaining efficiency.

\paragraph{YOLOv10.}
YOLOv10~\cite{Wang_NeurIPS_2024} is a one-stage detector that removes the need for non-maximum suppression (NMS) at inference. This is achieved via \emph{consistent dual assignments} during training, implemented through paired one-to-many and one-to-one matching along with an additional one-to-one detection head. Both heads are optimized jointly to provide richer supervision while enabling NMS-free inference.

\paragraph{YOLOv11.}
YOLOv11~\cite{yolo11_ultralytics} is a recent YOLO-family variant that introduces incremental architectural changes aimed at improving the speed--accuracy trade-off. Compared to prior Ultralytics models, it uses updated building blocks (e.g., C3k2) and integrates attention-style modules (e.g., C2PSA), while retaining standard multi-scale aggregation components such as SPPF.

\paragraph{RT-DETR.}
RT-DETR~\cite{Zhao_2024_CVPR} is an end-to-end transformer-based detector that, like other DETR variants, performs set prediction and is therefore NMS-free by design. To reduce the computational cost typical of DETR-style models, RT-DETR employs a hybrid encoder for multi-scale feature processing and uses a query-selection strategy to initialize decoder queries more effectively.

We use the Ultralytics~\cite{yolo11_ultralytics} implementations\footnote{\url{https://github.com/ultralytics/ultralytics} (accessed May 2026)} with weights pretrained on COCO \texttt{train2017} and default settings, including an input size of $640$. For detection-performance evaluation, we use a confidence threshold of $0.001$ for all detectors to standardize score filtering across architectures.

The full \emph{backbone} and \emph{head} layers of RT-DETR-l are listed in Listing~\ref{lst:RTDETRlayers}. For feature extraction, we use layers $[0, 2, 4, 8, 9]$.

\begin{figure}[h]
	\centering
	\begin{lstlisting}[caption=RT-DETR-l \emph{backbone} and \emph{head} layers (\href{https://github.com/ultralytics/ultralytics/blob/main/ultralytics/cfg/models/v10/yolov10m.yaml}{configuration file}; last accessed May 2026)., label={lst:RTDETRlayers}]
		backbone:
		# [from, repeats, module, args]
		- [-1, 1, HGStem, [32, 48]] # 0-P2/4
		- [-1, 6, HGBlock, [48, 128, 3]] # stage 1		
		- [-1, 1, DWConv, [128, 3, 2, 1, False]] # 2-P3/8
		- [-1, 6, HGBlock, [96, 512, 3]] # stage 2		
		- [-1, 1, DWConv, [512, 3, 2, 1, False]] # 4-P3/16
		- [-1, 6, HGBlock, [192, 1024, 5, True, False]] # 5
		- [-1, 6, HGBlock, [192, 1024, 5, True, True]] # 6
		- [-1, 6, HGBlock, [192, 1024, 5, True, True]] # 7 		
		- [-1, 1, DWConv, [1024, 3, 2, 1, False]] # 8-P4/32
		- [-1, 6, HGBlock, [384, 2048, 5, True, False]] # 9		
		head:
		- [-1, 1, Conv, [256, 1, 1, None, 1, 1, False]] # 10 
		- [-1, 1, AIFI, [1024, 8]] #11
		- [-1, 1, Conv, [256, 1, 1]] # 12, 
		- [-1, 1, nn.Upsample, [None, 2, "nearest"]] #13
		- [7, 1, Conv, [256, 1, 1, None, 1, 1, False]] # 14 
		- [[-2, -1], 1, Concat, [1]] # 15
		- [-1, 3, RepC3, [256]] # 16 
		- [-1, 1, Conv, [256, 1, 1]] # 17 	
		- [-1, 1, nn.Upsample, [None, 2, "nearest"]] # 18
		- [3, 1, Conv, [256, 1, 1, None, 1, 1, False]] # 19 
		- [[-2, -1], 1, Concat, [1]] # 20 
		- [-1, 3, RepC3, [256]] # 21 		
		- [-1, 1, Conv, [256, 3, 2]] # 22, 
		- [[-1, 17], 1, Concat, [1]] # 23
		- [-1, 3, RepC3, [256]] # 24 	
		- [-1, 1, Conv, [256, 3, 2]] # 25
		- [[-1, 12], 1, Concat, [1]] # 26
		- [-1, 3, RepC3, [256]] # 27
	\end{lstlisting}	
\end{figure}
For YOLOv9-m, the main \emph{backbone} and \emph{head} layers are shown in Listing~\ref{lst:YOLOv9layers}. For feature extraction, we use layers $[0, 1, 3, 5, 7, 9]$.

\begin{figure}[h]
	\centering
	\begin{lstlisting}[caption=YOLOv9-m \emph{backbone} and \emph{head} layers (\href{https://github.com/ultralytics/ultralytics/blob/main/ultralytics/cfg/models/v9/yolov9m.yaml}{configuration file}; last accessed May 2026)., label={lst:YOLOv9layers}]
		backbone:
		- [-1, 1, Conv, [32, 3, 2]] # 0-P1/2
		- [-1, 1, Conv, [64, 3, 2]] # 1-P2/4
		- [-1, 1, RepNCSPELAN4, [128, 128, 64, 1]] # 2
		- [-1, 1, AConv, [240]] # 3-P3/8
		- [-1, 1, RepNCSPELAN4, [240, 240, 120, 1]] # 4
		- [-1, 1, AConv, [360]] # 5-P4/16
		- [-1, 1, RepNCSPELAN4, [360, 360, 180, 1]] # 6
		- [-1, 1, AConv, [480]] # 7-P5/32
		- [-1, 1, RepNCSPELAN4, [480, 480, 240, 1]] # 8
		- [-1, 1, SPPELAN, [480, 240]] # 9		
		head:
		- [-1, 1, nn.Upsample, [None, 2, "nearest"]] # 10
		- [[-1, 6], 1, Concat, [1]] # 11
		- [-1, 1, RepNCSPELAN4, [360, 360, 180, 1]] # 12		
		- [-1, 1, nn.Upsample, [None, 2, "nearest"]] #13
		- [[-1, 4], 1, Concat, [1]] # 14
		- [-1, 1, RepNCSPELAN4, [240, 240, 120, 1]] # 15		
		- [-1, 1, AConv, [180]] #16
		- [[-1, 12], 1, Concat, [1]] # 17
		- [-1, 1, RepNCSPELAN4, [360, 360, 180, 1]] # 18 (P4/16-medium)		
		- [-1, 1, AConv, [240]] # 19
		- [[-1, 9], 1, Concat, [1]] # 20
		- [-1, 1, RepNCSPELAN4, [480, 480, 240, 1]] # 21 (P5/32-large)
	\end{lstlisting}	
\end{figure}
The primary \emph{backbone} and \emph{head} layers for YOLOv10-m are shown in Listing~\ref{lst:YOLOv10layers}. For feature extraction, we use layers $[0, 1, 3, 5, 7, 10]$.

\begin{figure}[h]
	\begin{lstlisting}[caption=YOLOv10-m \emph{backbone} and \emph{head} layers (\href{https://github.com/ultralytics/ultralytics/blob/main/ultralytics/cfg/models/v10/yolov10m.yaml}{configuration file}; last accessed May 2026), label={lst:YOLOv10layers}]
		backbone:
		# [from, repeats, module, args]
		- [-1, 1, Conv, [64, 3, 2]] # 0-P1/2
		- [-1, 1, Conv, [128, 3, 2]] # 1-P2/4
		- [-1, 3, C2f, [128, True]] # 4
		- [-1, 1, Conv, [256, 3, 2]] # 3-P3/8
		- [-1, 6, C2f, [256, True]] # 4
		- [-1, 1, SCDown, [512, 3, 2]] # 5-P4/16
		- [-1, 6, C2f, [512, True]] # 6
		- [-1, 1, SCDown, [1024, 3, 2]] # 7-P5/32
		- [-1, 3, C2fCIB, [1024, True]]
		- [-1, 1, SPPF, [1024, 5]] # 9
		- [-1, 1, PSA, [1024]] # 10	
		head:
		- [-1, 1, nn.Upsample, [None, 2, "nearest"]] # 11
		- [[-1, 6], 1, Concat, [1]] # 12 P4
		- [-1, 3, C2f, [512]] # 13
		- [-1, 1, nn.Upsample, [None, 2, "nearest"]] # 14
		- [[-1, 4], 1, Concat, [1]] # 15 P3
		- [-1, 3, C2f, [256]] # 16 (P3/8-small)
		- [-1, 1, Conv, [256, 3, 2]] # 17
		- [[-1, 13], 1, Concat, [1]] # 18 
		- [-1, 3, C2fCIB, [512, True]] # 19 (P4/16-medium)
		- [-1, 1, SCDown, [512, 3, 2]] # 20
		- [[-1, 10], 1, Concat, [1]] # 21 
		- [-1, 3, C2fCIB, [1024, True]] # 22 (P5/32-large)	
	\end{lstlisting}	
\end{figure}
For YOLOv11, the selected layers used for activation extraction are listed in Listing~\ref{lst:YOLOv11layers}. For feature extraction, we use layers $[0, 1, 3, 5, 7, 10]$.

\begin{figure}[h]
	\centering
	\begin{lstlisting}[caption=YOLOv11 \emph{backbone} and \emph{head} layers (\href{https://github.com/ultralytics/ultralytics/blob/main/ultralytics/cfg/models/11/yolo11.yaml}{configuration file}; last accessed May 2026)., label={lst:YOLOv11layers}]
		backbone:
		# [from, repeats, module, args]
		- [-1, 1, Conv, [64, 3, 2]] # 0-P1/2
		- [-1, 1, Conv, [128, 3, 2]] # 1-P2/4
		- [-1, 2, C3k2, [256, False, 0.25]]
		- [-1, 1, Conv, [256, 3, 2]] # 3-P3/8
		- [-1, 2, C3k2, [512, False, 0.25]]
		- [-1, 1, Conv, [512, 3, 2]] # 5-P4/16
		- [-1, 2, C3k2, [512, True]] # 6
		- [-1, 1, Conv, [1024, 3, 2]] # 7-P5/32
		- [-1, 2, C3k2, [1024, True]]
		- [-1, 1, SPPF, [1024, 5]] # 9
		- [-1, 2, C2PSA, [1024]] # 10	
		head:
		- [-1, 1, nn.Upsample, [None, 2, "nearest"]] #11
		- [[-1, 6], 1, Concat, [1]] # 12
		- [-1, 2, C3k2, [512, False]] # 13		
		- [-1, 1, nn.Upsample, [None, 2, "nearest"]] # 14
		- [[-1, 4], 1, Concat, [1]] # 15
		- [-1, 2, C3k2, [256, False]] # 16 (P3/8-small)		
		- [-1, 1, Conv, [256, 3, 2]] # 17
		- [[-1, 13], 1, Concat, [1]] # 18
		- [-1, 2, C3k2, [512, False]] # 19 (P4/16-medium)
		- [-1, 1, Conv, [512, 3, 2]] # 20
		- [[-1, 10], 1, Concat, [1]] # 21
		- [-1, 2, C3k2, [1024, True]] # 22 (P5/32-large)		
	\end{lstlisting}	
\end{figure}

\clearpage
\section{Datasets and Natural Weather Splits}
\label{app:Datasets}
To evaluate self-awareness under \emph{natural} distribution shift, we construct weather-based ID/OoD splits on Seeing Through Fog (STF)~\cite{Bijelic_2020_STF} and BDD100K \cite{Yu_CVPR_2020}. In both datasets, the in-distribution (ID) partition corresponds to nominal clear-weather conditions, while the out-of-distribution (OoD) partition contains images with strong, visually apparent adverse weather.

STF provides weather annotations including \texttt{dense fog} and \texttt{heavy snow}. We define ID as clear or nominal conditions and construct corresponding OoD subsets using these labels. For the \texttt{rain} condition, we further refine the subset by visual inspection to exclude ambiguous cases, such as scenes dominated by wet surfaces without clearly visible precipitation. This yields a more conservative OoD split that better reflects perceptually strong weather shift.

BDD100K includes attribute metadata such as \texttt{clear} and \texttt{rainy}. We define ID using the \texttt{clear} subset and OoD using the \texttt{rainy} subset. However, the \texttt{rainy} label includes scenes with weak or ambiguous cues. We therefore manually exclude such cases and retain only images with strong visible weather effects, such as rain streaks, reduced visibility, or windshield artifacts.

DSERT-RoLL provides weather conditions with varying severity. For consistency with the main paper, we use the \texttt{heavy} weather subsets for evaluation, namely \texttt{dense fog}, \texttt{heavy snow}, and \texttt{heavy rain}. We do not use the corresponding \texttt{light} conditions in the main benchmark, since they define a substantially more ambiguous boundary between nominal and adverse weather and therefore are less suitable for a clean ID/OoD evaluation. As with STF and BDD100K, the goal is to isolate conditions with clear scene-level weather impact rather than subtle appearance changes.

Natural weather shift is inherently heterogeneous and entangled with scene content, acquisition conditions, and weather severity. Our dataset construction is therefore intentionally conservative: it aims to isolate clearly adverse conditions and provide a cleaner evaluation of whether the monitor separates nominal from strongly shifted inputs. Example images from the resulting ID/OoD splits are shown in Fig.~\ref{fig:stf_examples}, Fig.~\ref{fig:bdd_examples}, and Fig.~\ref{fig:dsertroll_examples}.

\renewcommand{\subw}{0.3\textwidth}
\begin{figure*}[h!]
	\centering
		\begin{subfigure}{\subw}
			\includegraphics[width=\linewidth]{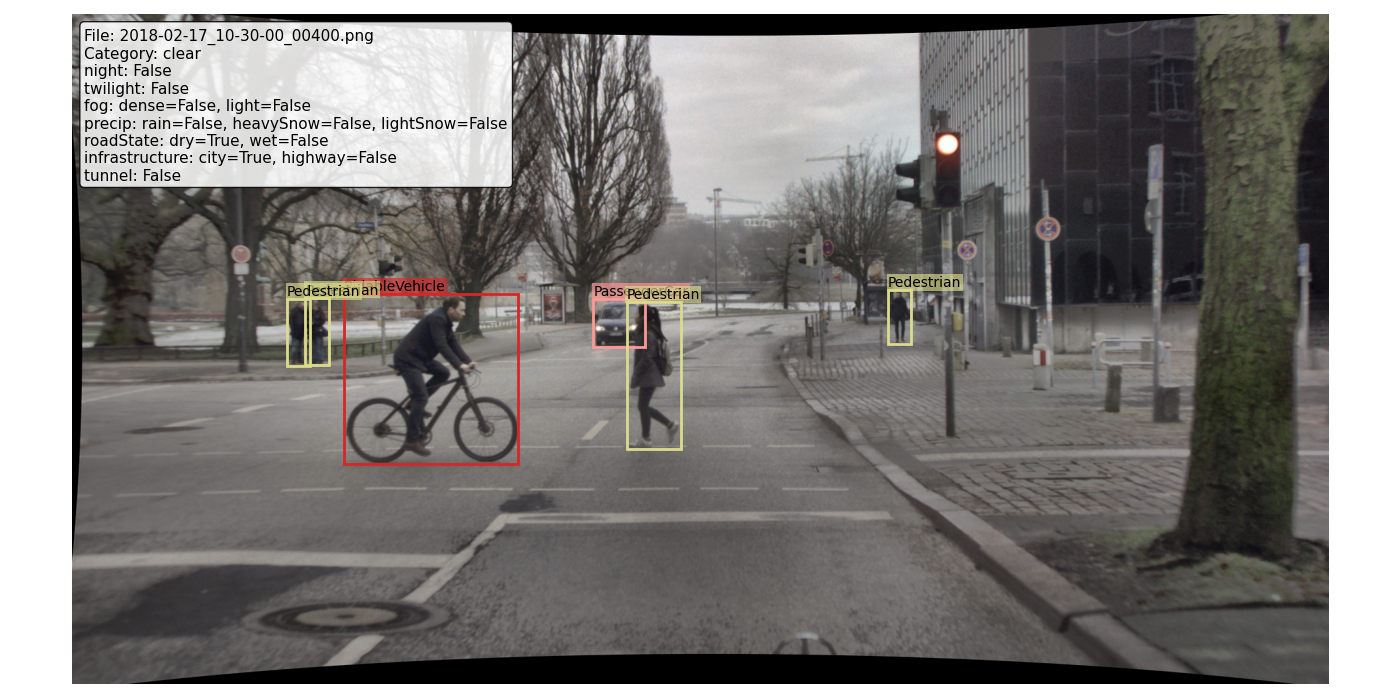}
			\caption{\texttt{clear}}
		\end{subfigure}		
		\hfill					
		\begin{subfigure}{\subw}
			\includegraphics[width=\linewidth]{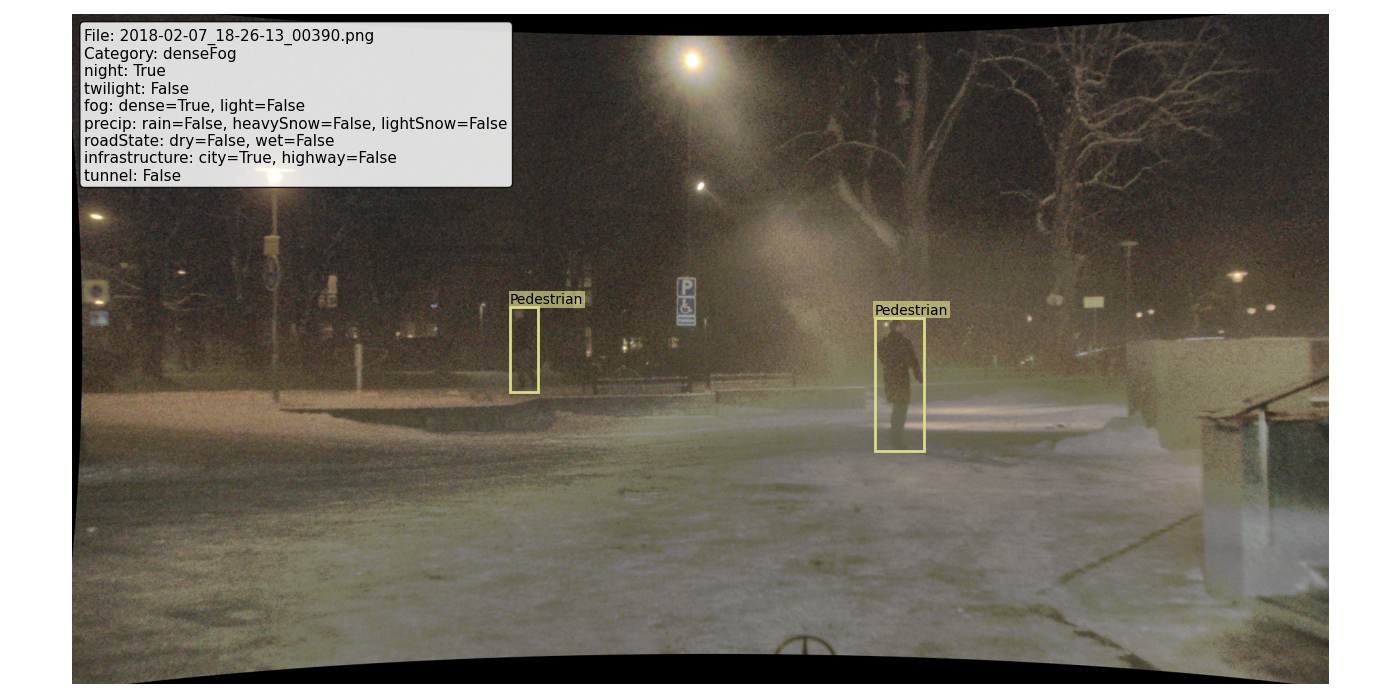}
			\caption{\texttt{dense fog}}
		\end{subfigure}
		\hfill		
		\begin{subfigure}{\subw}
			\includegraphics[width=\linewidth]{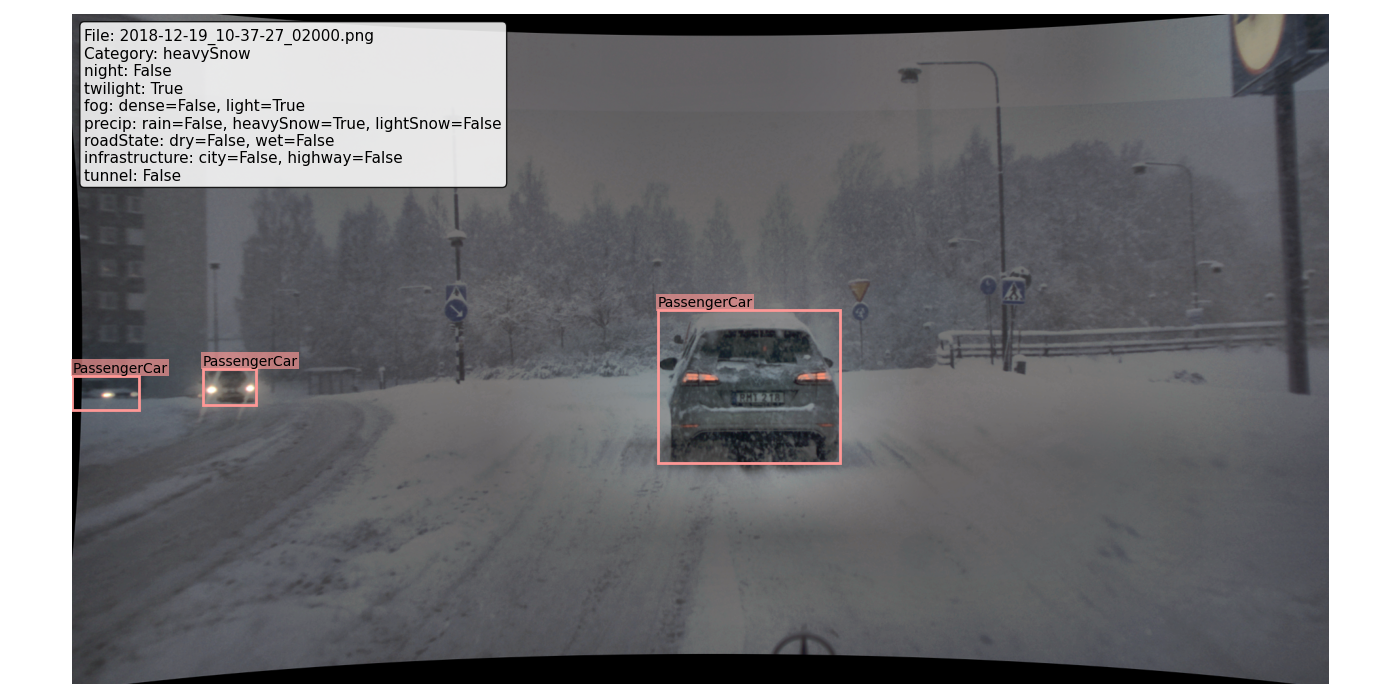}
			\caption{\texttt{heavy snow}}
		\end{subfigure}
		\caption{
			Examples from Seeing Through Fog (STF)~\cite{Bijelic_2020_STF} used for the natural weather-shift evaluation. 
			We construct an ID split from clear/nominal scenes and OoD splits from adverse-weather conditions (\texttt{dense fog}, \texttt{heavy snow}, and \texttt{heavy rain} after filtering).
		}		
		\label{fig:stf_examples}  
		\vspace{-0.5cm} 
\end{figure*}
\renewcommand{\subw}{0.3\textwidth}
\begin{figure*}[h!]
	\centering
		
		\begin{subfigure}{\subw}
			\includegraphics[width=\linewidth]{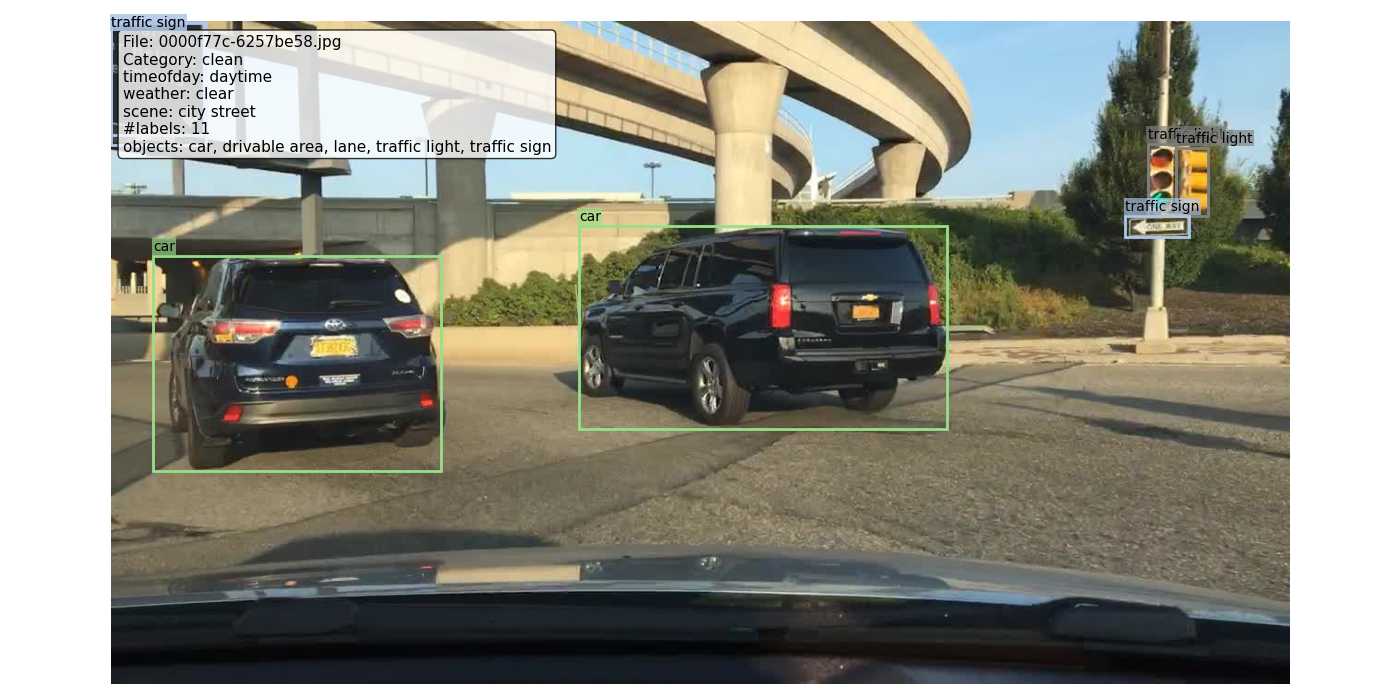}
			\caption{\texttt{clear}}
		\end{subfigure}
		\hfill		
		\begin{subfigure}{\subw}
			\includegraphics[width=\linewidth]{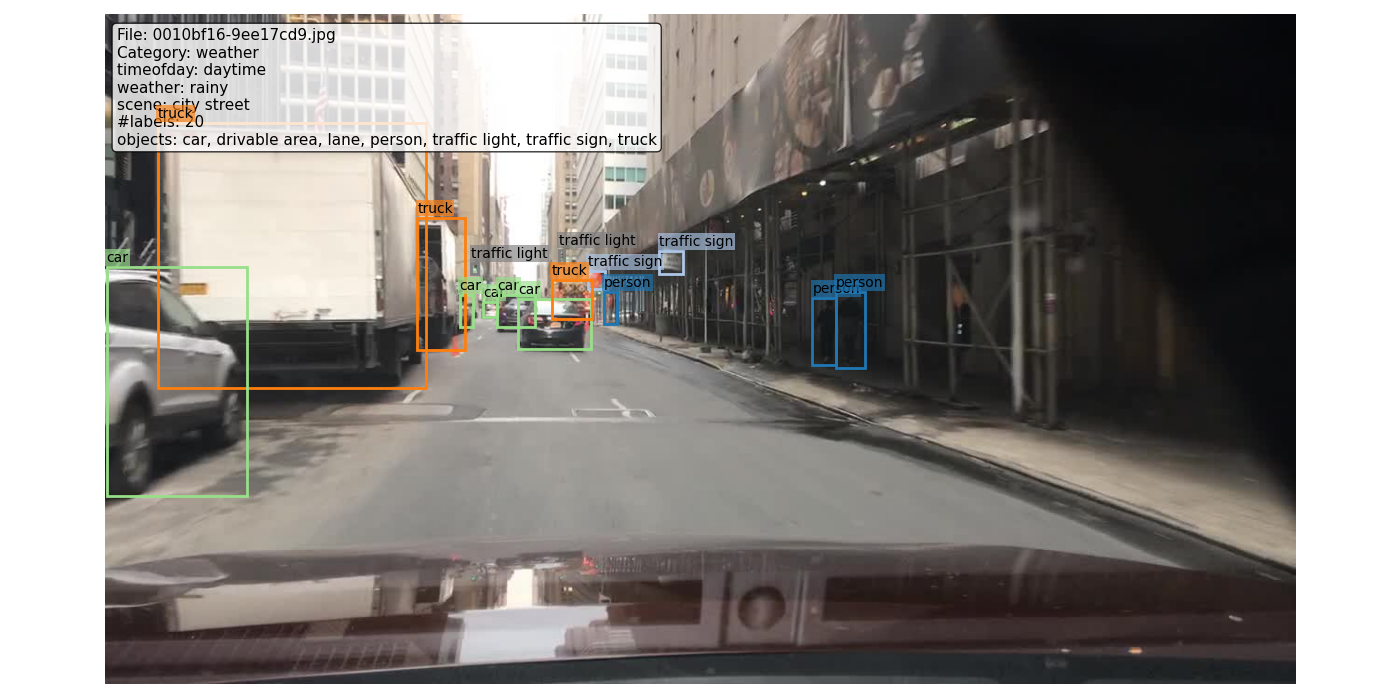}
			\caption{\texttt{rainy} (excluded)}
		\end{subfigure}
		\hfill	
		\begin{subfigure}{\subw}
			\includegraphics[width=\linewidth]{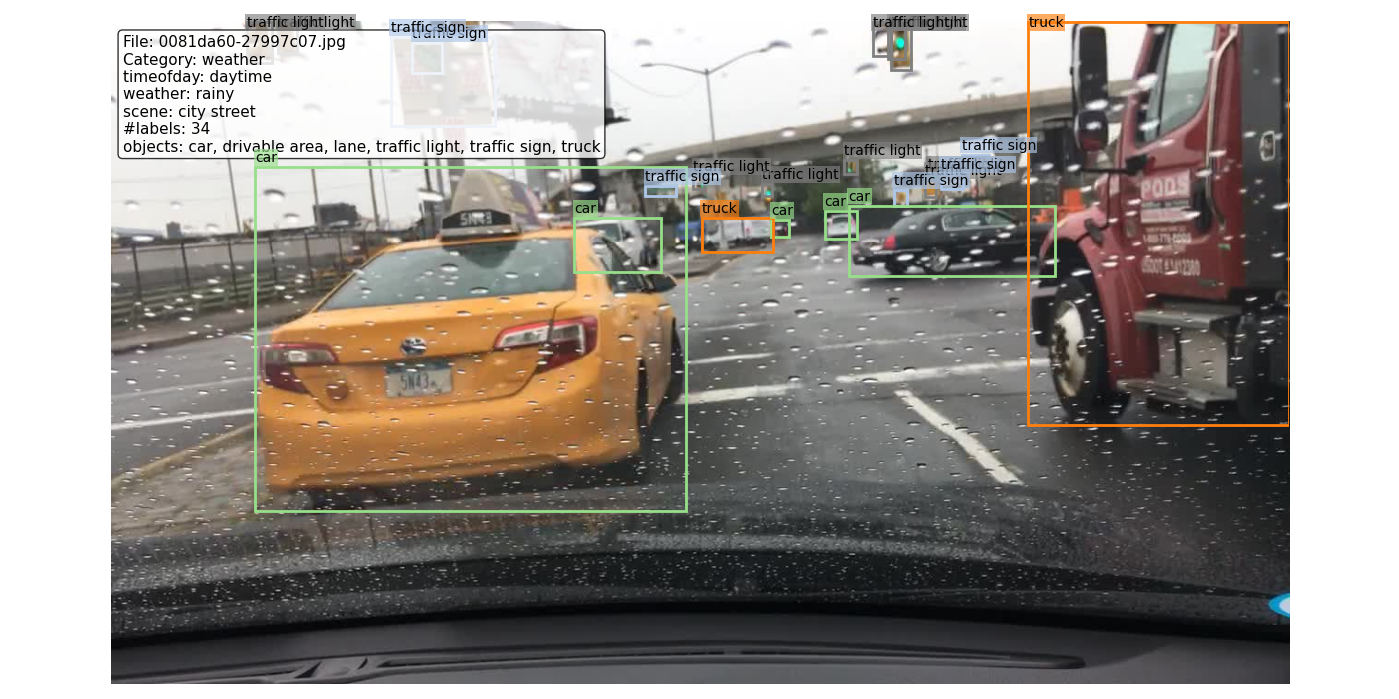}
			\caption{\texttt{heavy rain}}
		\end{subfigure}		
		\caption{
			Examples from BDD100K~\cite{Yu_CVPR_2020} illustrating the weather-based ID/OoD construction.
			Left: ID example from the \texttt{clear} subset.
			Middle: ambiguous \texttt{rainy} example with weak cues (e.g., wet road only), which is excluded by manual refinement.
			Right: retained OoD example with strong visible rain cues (\texttt{heavy rain}).
		}			
		\label{fig:bdd_examples}  
		\vspace{-0.5cm} 
\end{figure*}

\renewcommand{\subw}{0.3\textwidth}
\begin{figure*}[h!]
	\centering
		\begin{subfigure}{\subw}
			\includegraphics[width=\linewidth]{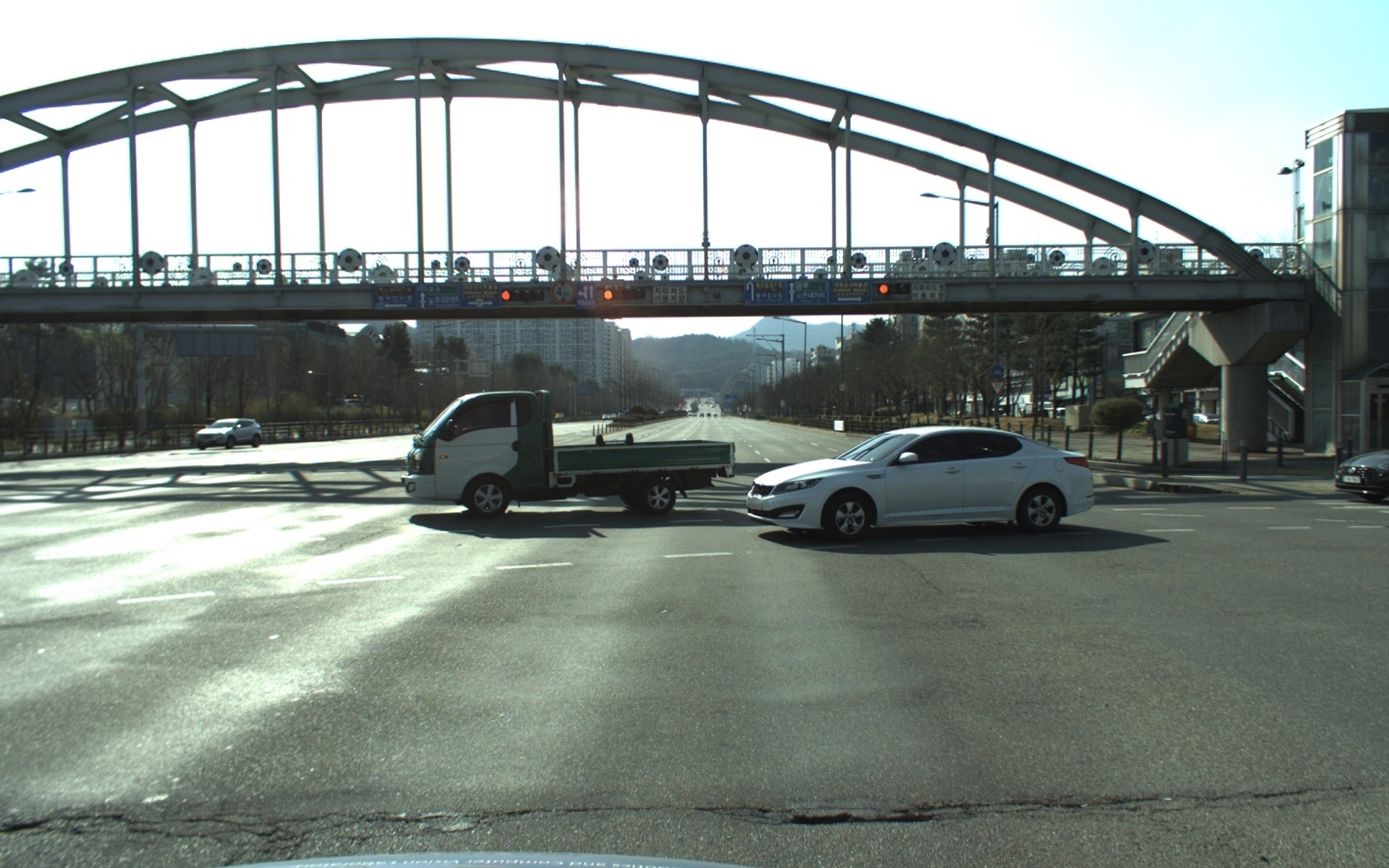}
			\caption{\texttt{clear}}
		\end{subfigure}		
		\hfill					
		\begin{subfigure}{\subw}
			\includegraphics[width=\linewidth]{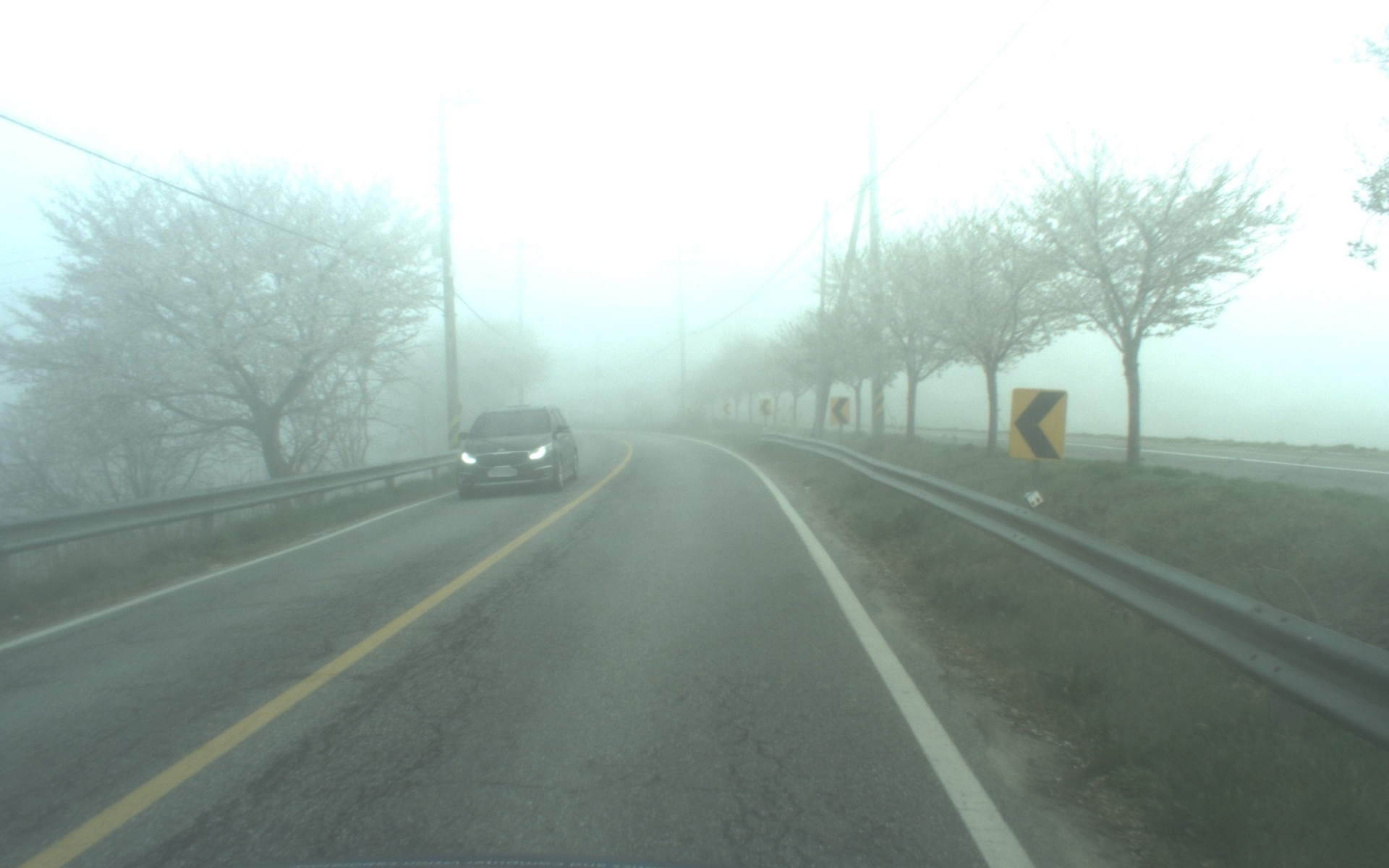}
			\caption{\texttt{dense fog}}
		\end{subfigure}
		\hfill		
		\begin{subfigure}{\subw}
			\includegraphics[width=\linewidth]{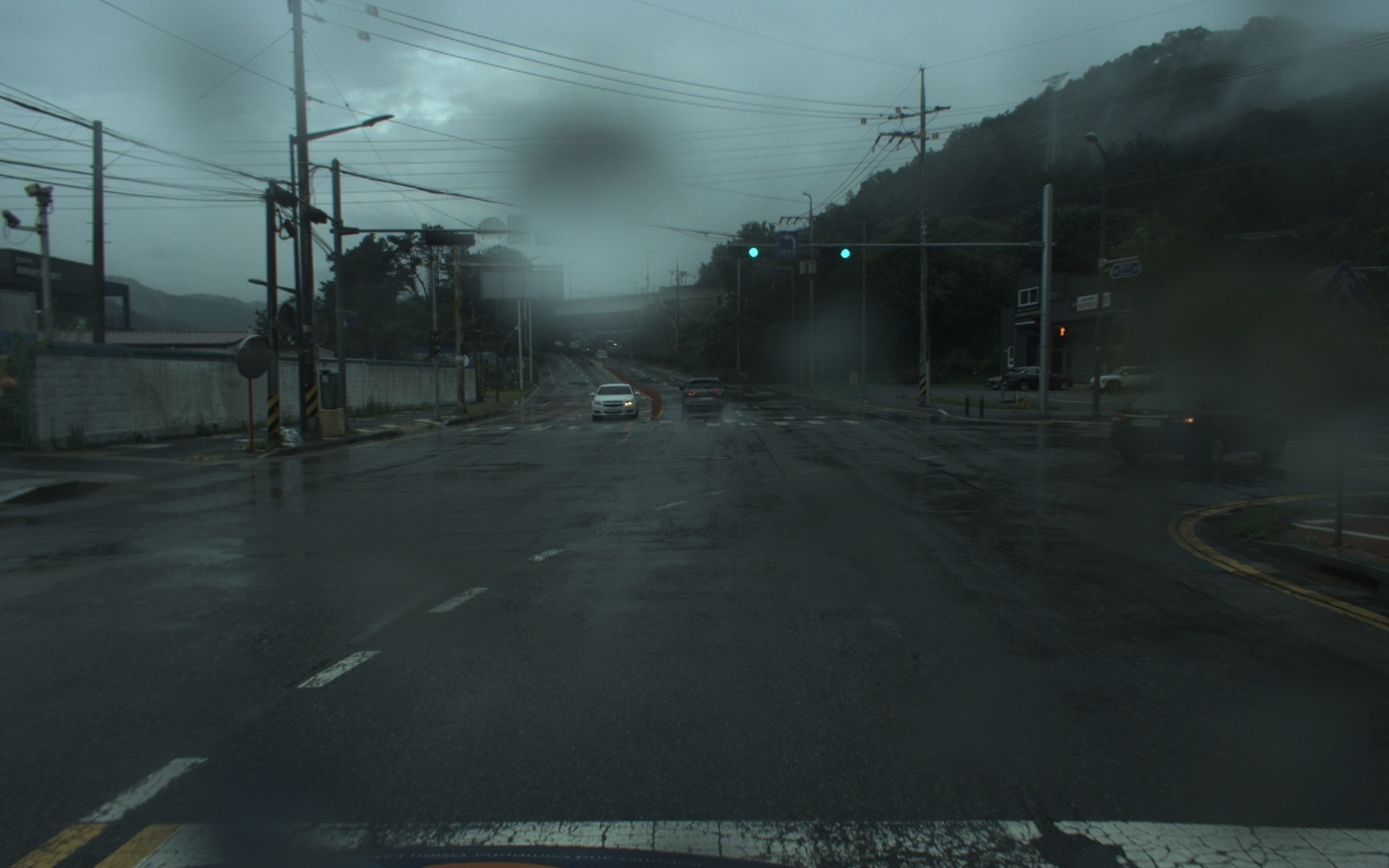}
			\caption{\texttt{heavy rain}}
		\end{subfigure}
		\caption{
			Examples from DSERT-RoLL~\cite{Cho_CVPR_2026} used for the natural weather-shift evaluation.
			We construct an ID split from clear/nominal scenes and OoD splits from adverse-weather conditions (\texttt{dense fog}, \texttt{heavy snow}, and \texttt{heavy rain}).
		}		
		\label{fig:dsertroll_examples}  
		\vspace{-0.5cm} 
\end{figure*}

\clearpage
\section{Extended Evaluation}
\label{app:evaluation}
This section reports additional results complementing the main paper, including extended detector-performance plots, additional embedding visualizations, mixed-dataset evaluation, ablations, operating-point metrics, and per-degradation analyses.

\subsection{Detector Performance}
Fig.~\ref{fig:map_coco_c} shows mAP degradation under increasing corruption severity for multiple detectors on mixed-corruption COCO \texttt{val2017}. All detectors exhibit a similar monotonic degradation trend as severity increases.

Fig.~\ref{fig:map_vs_severity_all} summarizes per-corruption performance trends for degradations from \cite{Michaelis_NeurIPSW_2019} and \cite{Agnolucci_WACV_2024}. Across the evaluated set, robustness-style corruptions generally induce stronger performance drops than IQA-motivated distortions. This reflects their different design goals: robustness corruptions are constructed to disrupt semantic and structural cues more aggressively, whereas IQA distortions model visually plausible quality variations that may preserve object-level structure. Overall, detection performance degrades monotonically with increasing severity for most corruptions, consistent with prior robustness analyses for object detection~\cite{Oksuz_CVPR_2023,Michaelis_NeurIPSW_2019,Liu_IJCV_2024}.

\renewcommand{\subw}{0.45\textwidth}
\begin{figure*}
	\centering
		\begin{subfigure}{\subw}
			\includegraphics[width=\linewidth]{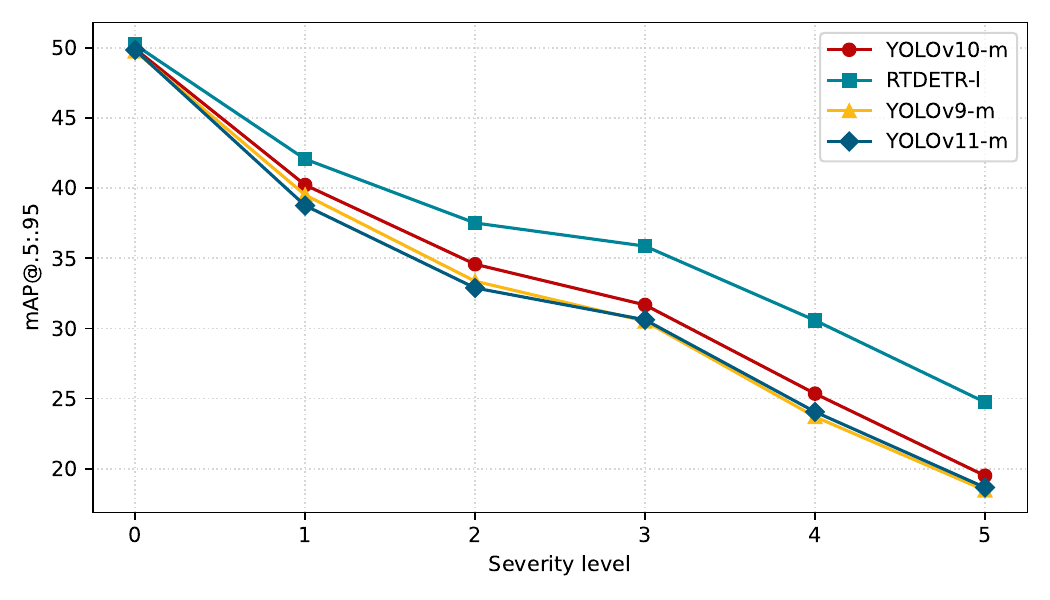}
			\caption*{mAP@.5:.95}
		\end{subfigure}
		\hfill
		\begin{subfigure}{\subw}
			\includegraphics[width=\linewidth]{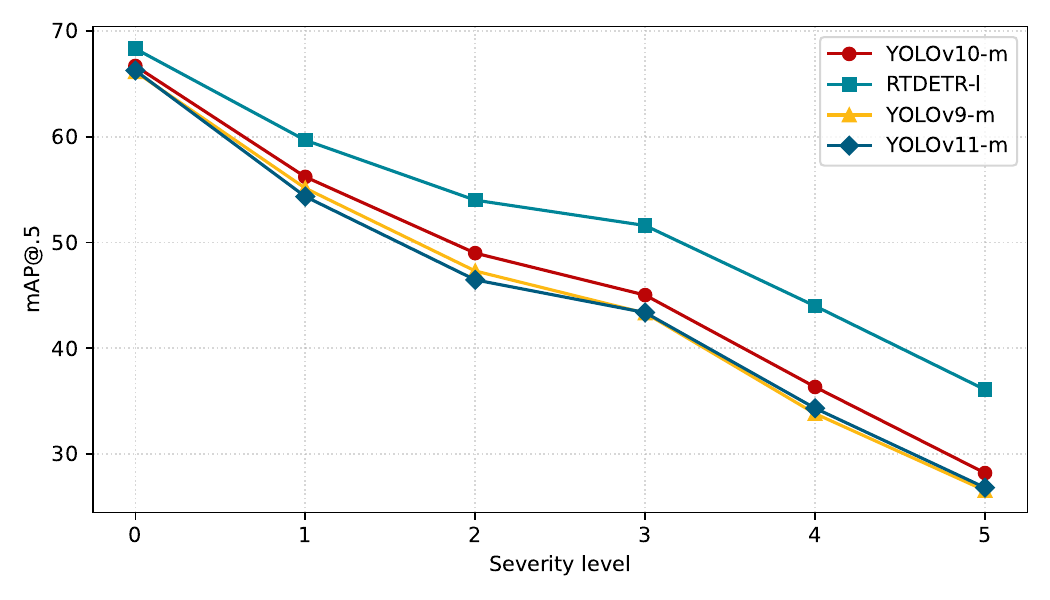}
			\caption*{mAP@.5}
		\end{subfigure}
		
		\caption{Detection performance of different detection models under increasing synthetic corruption severity on COCO \texttt{val2017}. Left: mAP@0.5:0.95. Right: mAP@0.5. Results are shown for YOLOv9-m~\cite{Wang_ECCV_2024}, YOLOv10-m~\cite{Wang_NeurIPS_2024}, YOLOv11-m~\cite{yolo11_ultralytics}, and RT-DETR-l~\cite{Zhao_2024_CVPR}.}			
		\label{fig:map_coco_c} 
		\vspace{-0.5cm} 
\end{figure*}

\renewcommand{\subw}{0.45\textwidth}
\begin{figure*}[h!]
	\captionsetup[subfigure]{justification=centering}
	\centering
	\begin{subfigure}{\subw}
		\centering
		\includegraphics[width=\textwidth]{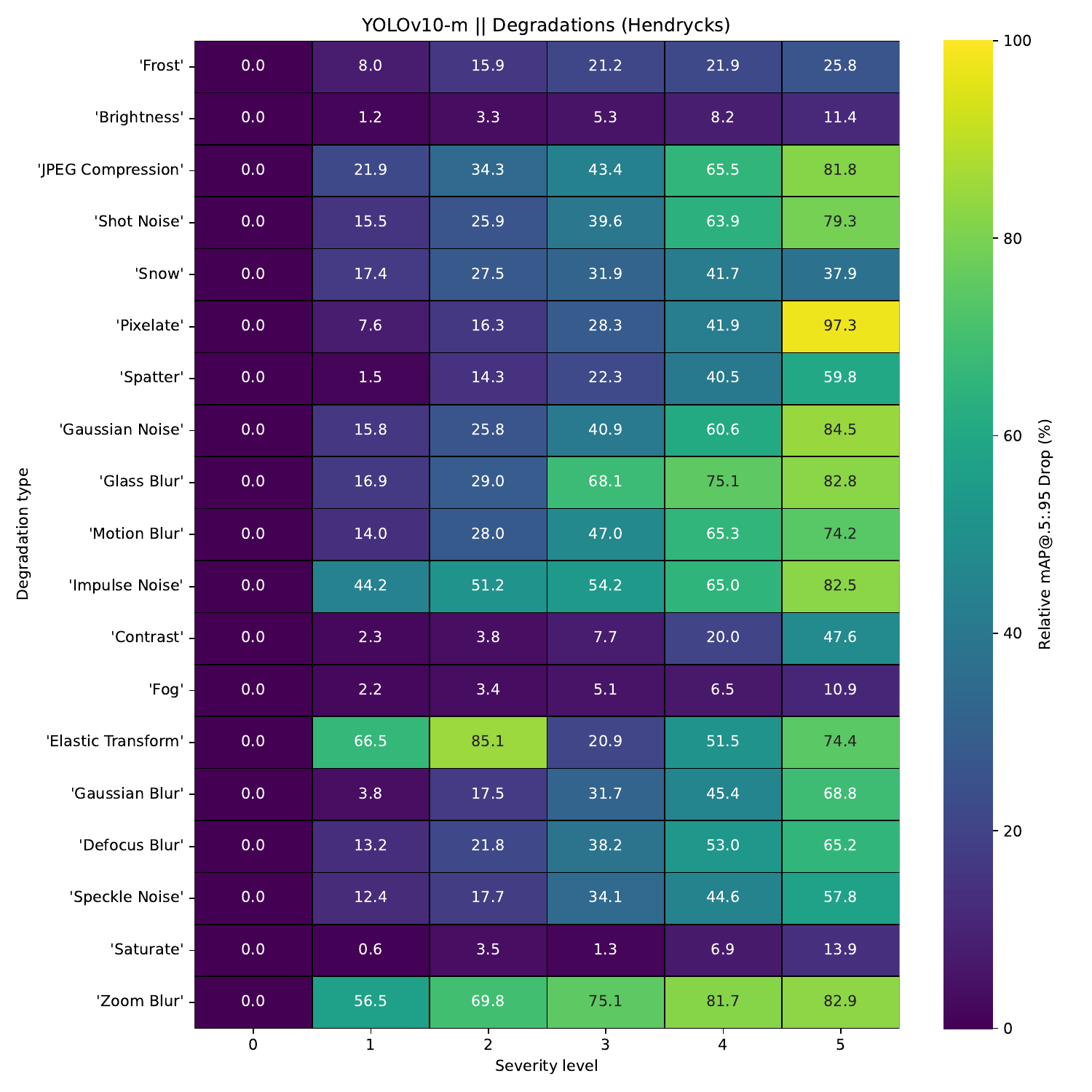}
		\caption*{Degradation functions from~\cite{Hendrycks_ICLR_2019}}
	\end{subfigure}
	\hfill
	\begin{subfigure}{\subw}
		\centering 
		\includegraphics[width=\textwidth]{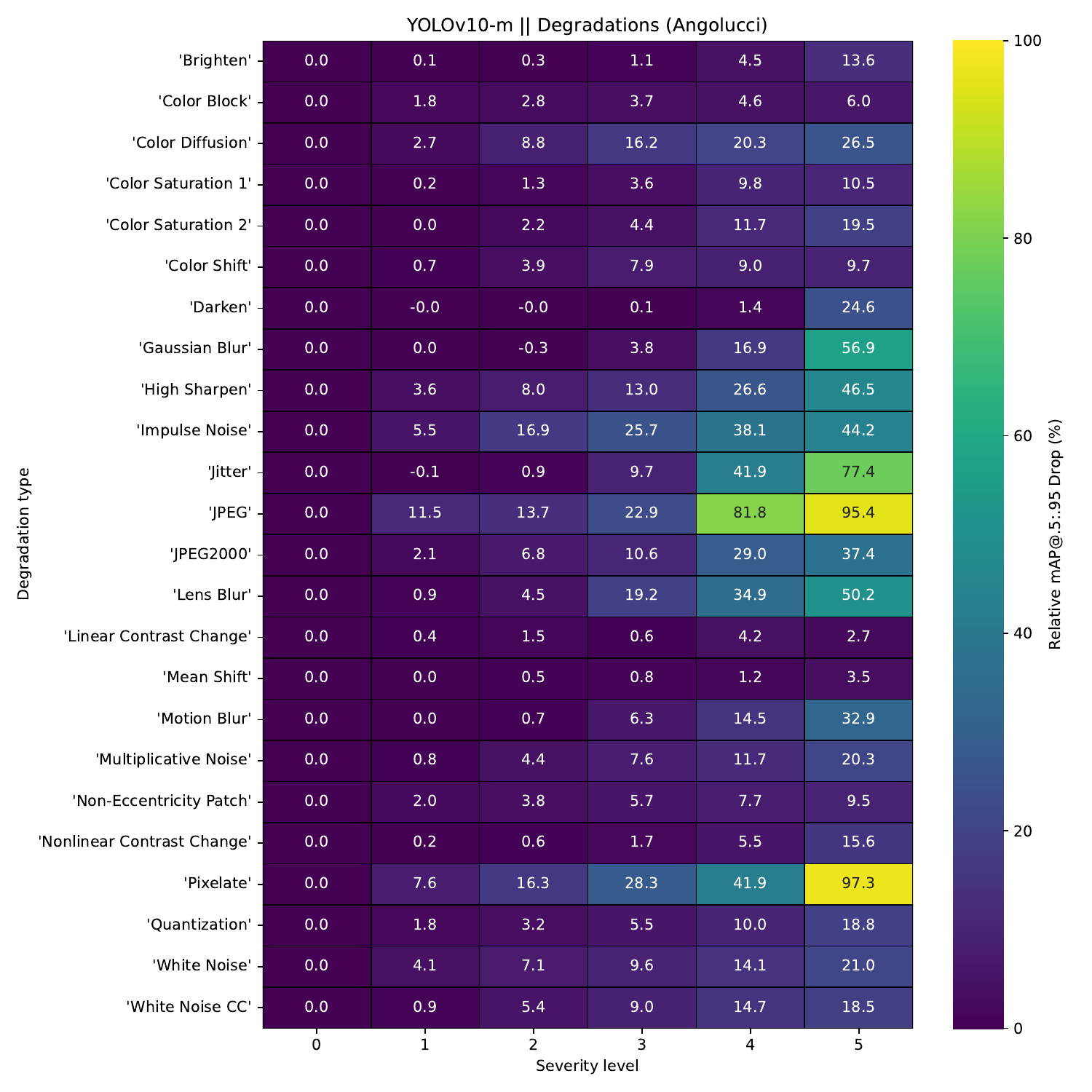}
		\caption*{Degradation functions from~\cite{Agnolucci_WACV_2024}}
	\end{subfigure}
	\caption{Relative detection-performance drop of YOLOv10-m~\cite{Wang_NeurIPS_2024} on COCO \texttt{val2017} under increasing native severity levels (mAP@0.5:0.95). Each cell shows the percentage drop relative to the clean baseline. Brighter cells indicate larger drops. Left: robustness-style degradations from~\cite{Michaelis_NeurIPSW_2019,Hendrycks_ICLR_2019}. Right: IQA-style degradations from~\cite{Agnolucci_WACV_2024}.}	
	\label{fig:map_vs_severity_all}
\end{figure*}

\subsection{Degradation Manifold Analysis}
\begin{figure*}[h!]
	\centering
		\begin{subfigure}{0.7\linewidth}
			\includegraphics[width=\linewidth]{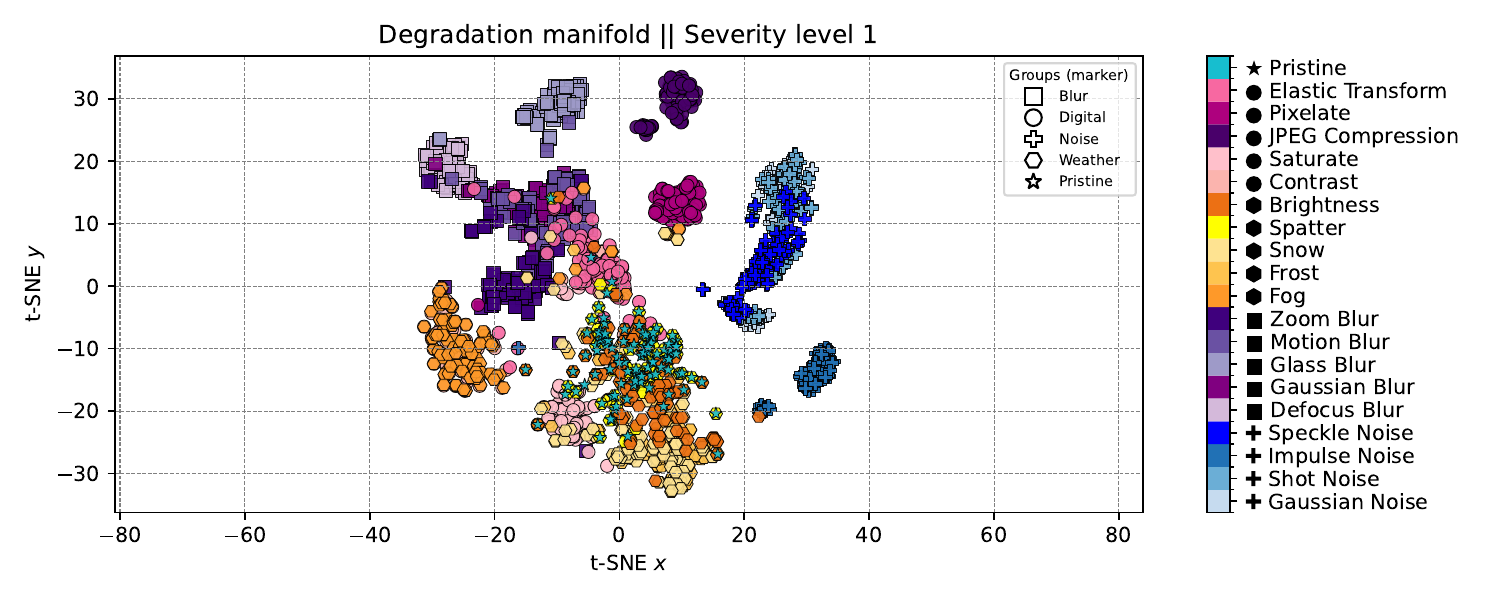}
			\caption{Severity level 1 (\cite{Hendrycks_ICLR_2019})}
		\end{subfigure}
		
		\begin{subfigure}{0.7\linewidth}
			\includegraphics[width=\linewidth]{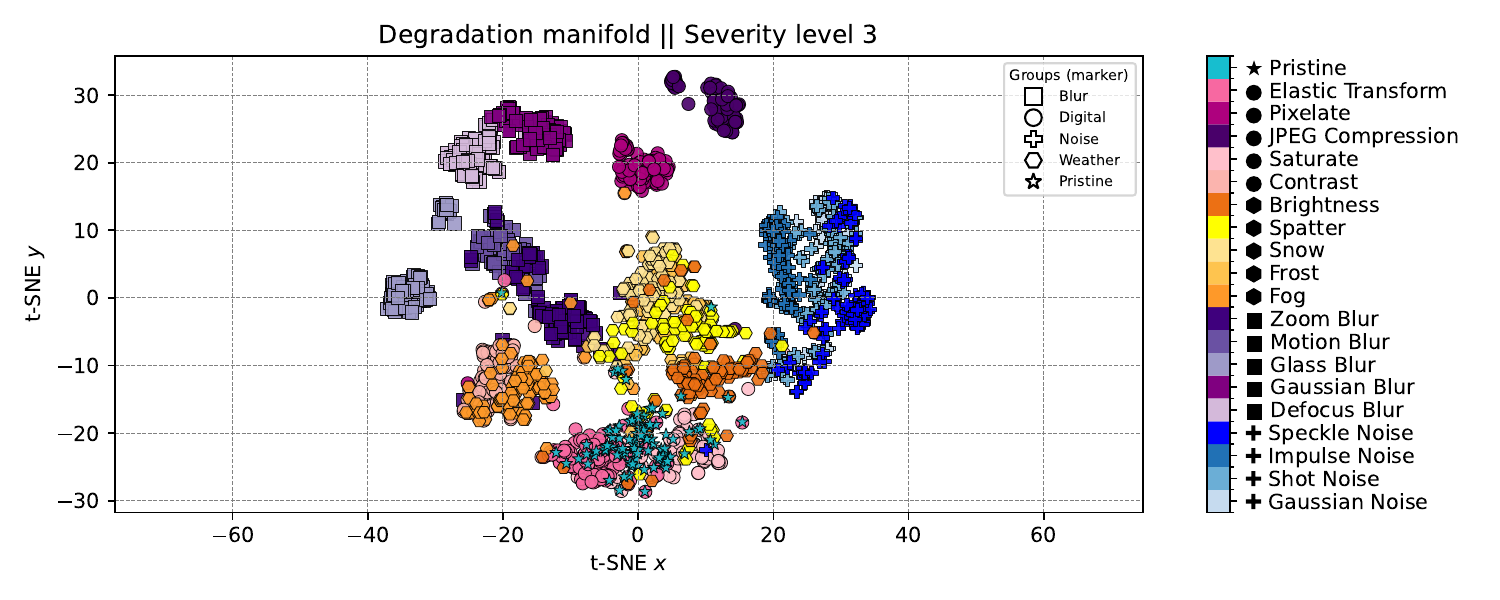}
			\caption{Severity level 3 (\cite{Hendrycks_ICLR_2019})}
		\end{subfigure}
		
		\begin{subfigure}{0.7\linewidth}
			\includegraphics[width=\linewidth]{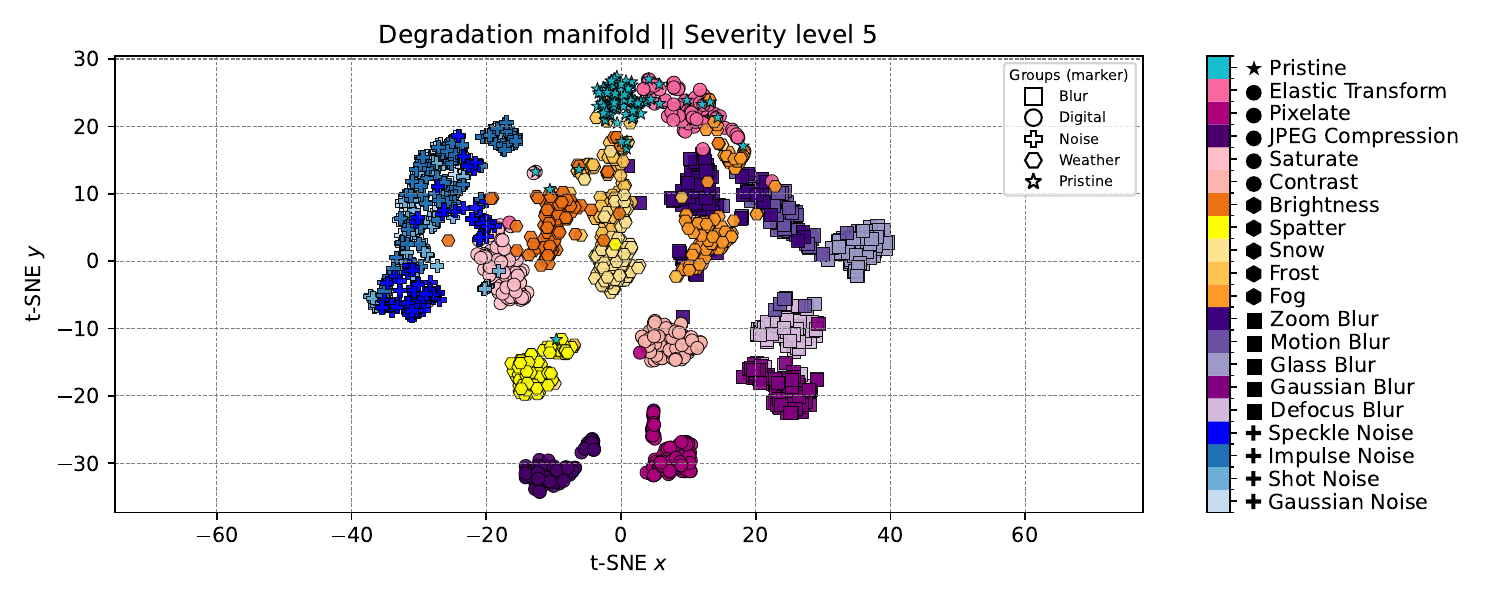}
			\caption{Severity level 5 \cite{Hendrycks_ICLR_2019}}
		\end{subfigure}		
		\caption{t-SNE visualization~\cite{Hinton_Roweis_2003} of the learned degradation manifold under the corruption taxonomy of~\cite{Michaelis_NeurIPSW_2019}, based on~\cite{Hendrycks_ICLR_2019}, shown for severity levels 1, 3, and 5. Each color denotes a degradation type, and marker styles denote degradation groups. For visualization, 100 COCO~\cite{Lin_ECCV_2014} samples are corrupted at the respective severity level. Training uses degradation compositions inspired by~\cite{Agnolucci_WACV_2024}, demonstrating zero-shot generalization across corruption families.}			
		\label{fig:embeddding_hendrycks}  
		\vspace{-0.5cm} 
\end{figure*}

\begin{figure*}
	\centering
		\begin{subfigure}{0.7\linewidth}
			\includegraphics[width=\linewidth]{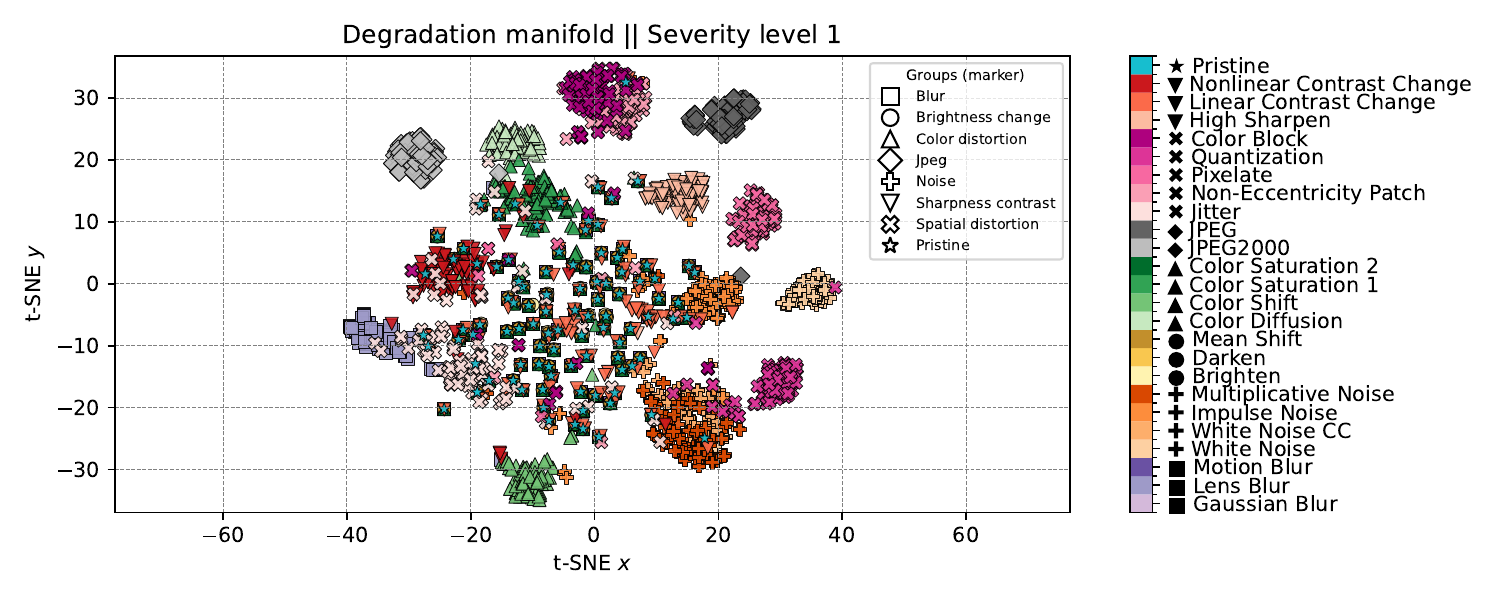}
			\caption{Severity level 1 (\cite{Agnolucci_WACV_2024})}
		\end{subfigure}
		
		\begin{subfigure}{0.7\linewidth}
			\includegraphics[width=\linewidth]{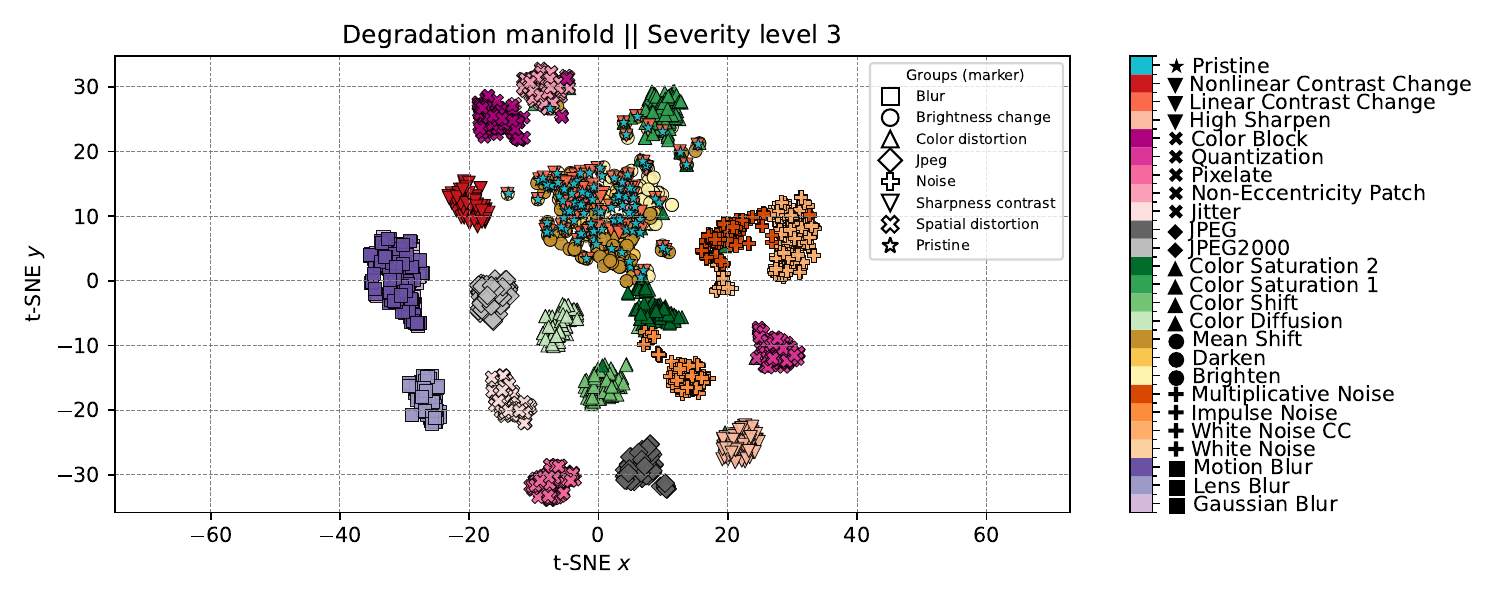}
			\caption{Severity level 3 (\cite{Agnolucci_WACV_2024})}
		\end{subfigure}
		
		\begin{subfigure}{0.7\linewidth}
			\includegraphics[width=\linewidth]{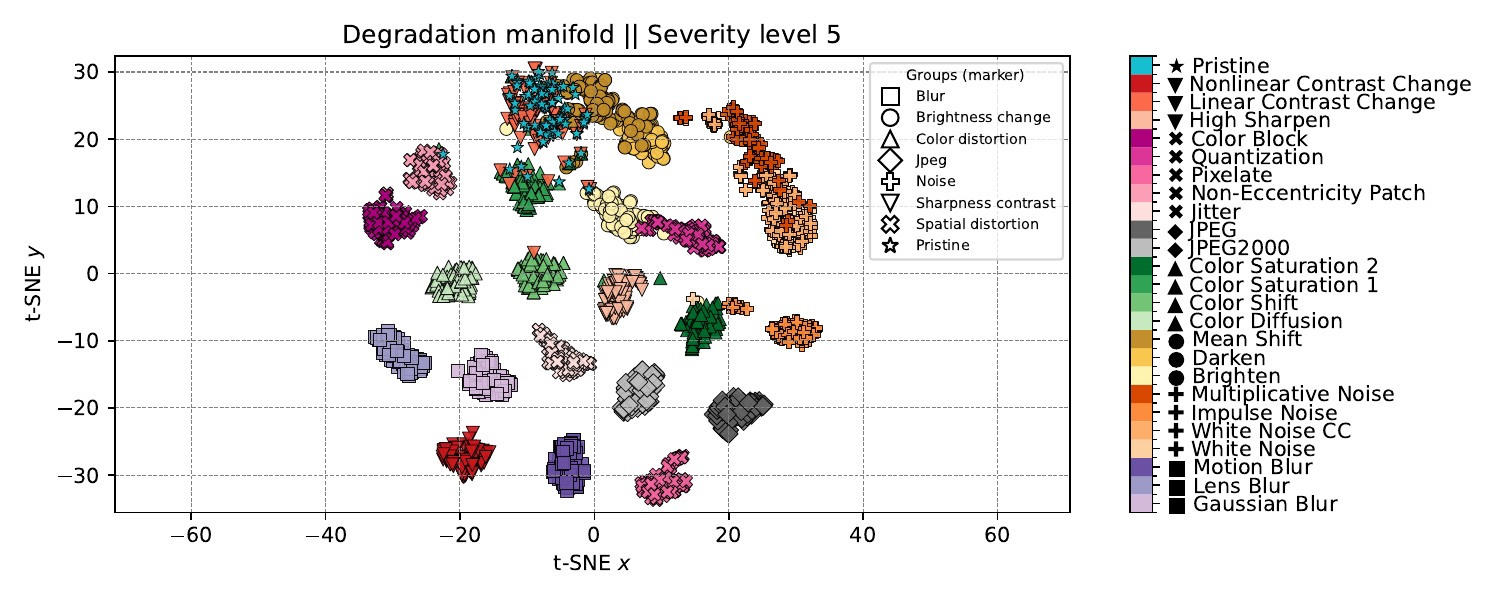}
			\caption{Severity level 5 (\cite{Agnolucci_WACV_2024})}
		\end{subfigure}		
		\caption{t-SNE visualization~\cite{Hinton_Roweis_2003} of the learned degradation manifold under the IQA-style corruption taxonomy of~\cite{Agnolucci_WACV_2024}, shown for severity levels 1, 3, and 5. Each color denotes a degradation type, and marker styles denote degradation groups. For visualization, 100 COCO~\cite{Lin_ECCV_2014} samples are corrupted at the respective severity level.}		
		\label{fig:embeddding_agnolucci}  
\end{figure*}

Figs.~\ref{fig:embeddding_hendrycks} and~\ref{fig:embeddding_agnolucci} visualize the learned embedding space using t-SNE~\cite{Hinton_Roweis_2003} for robustness-style and IQA-style degradations across several severity levels. Across both corruption taxonomies, degraded samples form clearly separated clusters by degradation type despite the absence of degradation labels during training, indicating that the representation is organized primarily by degradation characteristics rather than semantic content.

Pristine embeddings exhibit minor dispersion due to natural acquisition variability in COCO. Despite this, degraded samples form well-separated structures, confirming that the learned geometry captures systematic degradation beyond nominal data variation.


\begin{figure*}
	\centering
		\begin{subfigure}{0.9\linewidth}
			\includegraphics[width=\linewidth]{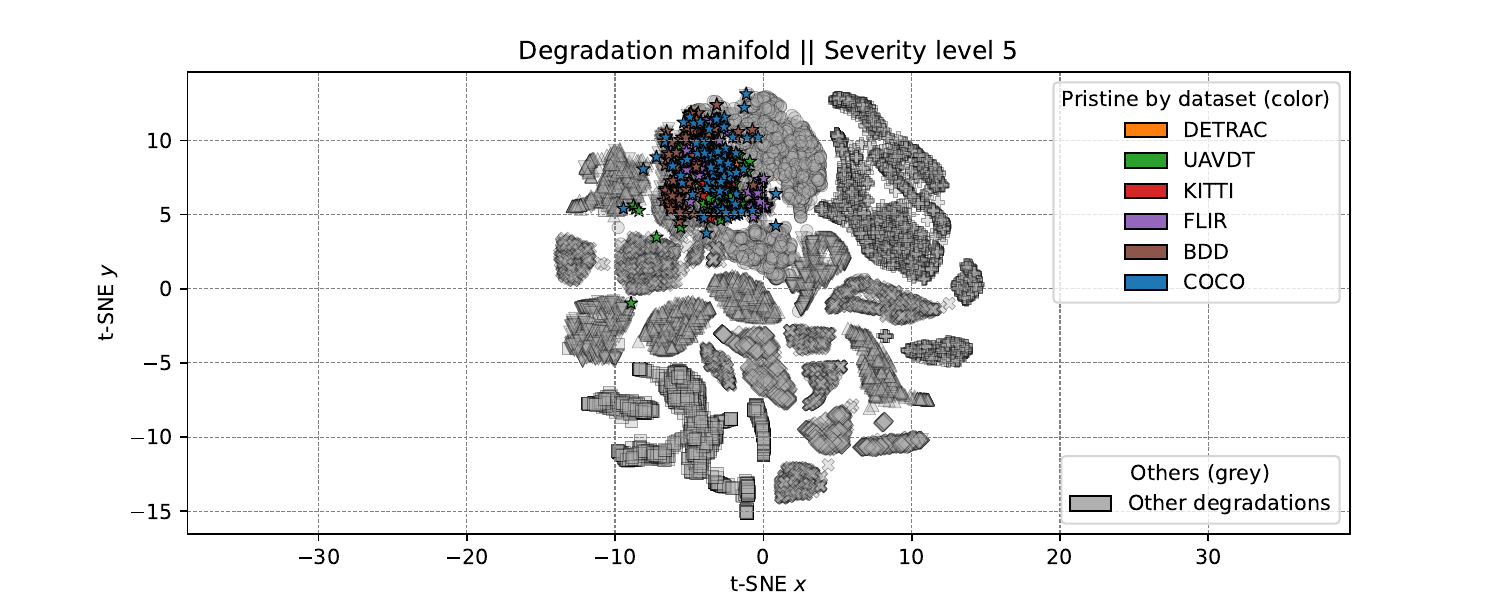}
			\caption*{}
		\end{subfigure}
		\caption{t-SNE visualization~\cite{Hinton_Roweis_2003} of the learned degradation manifold across multiple datasets. Degraded samples remain clearly separated by degradation type, while pristine samples from different datasets form a compact cluster. This indicates that the learned representation is primarily organized by degradation characteristics rather than semantic content or dataset identity.}			
		\label{fig:embeddding_datasets} 
\end{figure*}

Fig.~\ref{fig:embeddding_datasets} further illustrates cross-dataset stability of the pristine representation: pristine images from multiple datasets cluster together despite substantial differences in scene content and acquisition conditions. Notably, BDD100K~\cite{Yu_CVPR_2020} and KITTI~\cite{Geiger_CVPR_2012} were not part of the training pool, indicating that the learned embedding generalizes across datasets and remains largely aligned with degradation rather than dataset identity.

\subsection{Self-Awareness under Degradation}

\paragraph{Probabilistic Detectors.}
The full results for probabilistic detectors~\cite{Oksuz_CVPR_2023,Harakeh_NeurIPS_2021} are reported in Table~\ref{tab:top3_uncertainties_detectors}. We evaluate nine probabilistic detector variants across three backbones and three scoring rules (ES, NLL, DMM). Across models, degradation shift is reflected in detector outputs, but the quality and stability of the resulting image-level signals depend strongly on the training objective.

Energy Score (ES) variants yield more balanced predictive distributions under shift, whereas NLL-trained models often produce diffuse uncertainties and DMM variants tend to be overconfident. Among detector-derived monitoring signals, raw confidence consistently achieves the strongest AUROC, outperforming entropy- and covariance-based metrics. This is consistent with the fact that detector confidence is often better calibrated for high-quality detections than higher-order uncertainty statistics are for image-level aggregation.

Despite this sensitivity to shift, detector-derived monitoring has clear limitations. All output-based signals depend on the presence and stability of object hypotheses and implicitly assume meaningful detections in the scene. Under strong degradations, detections may disappear or become unstable, reducing the reliability of image-level aggregation. Even the best-performing confidence-based variants remain clearly below the proposed degradation-manifold approach.

For evaluation, we use the models and probabilistic extensions of~\cite{Harakeh_NeurIPS_2021} (\href{https://github.com/asharakeh/probdet}{repository}, accessed February 2026). Models are either trained on COCO~\cite{Lin_ECCV_2014} using default settings or evaluated using the provided COCO-pretrained weights. An input size of $640$ is used throughout.

\begin{table}[!h]
	\aurocTableSetup
	\resizebox{.7\linewidth}{!}{
		\begin{tabular}{c llccccc}
			\toprule
			& & & \multicolumn{5}{c}{\underline{Severity levels AUROC $\uparrow$}} \\
			& Detector & Model & 1 & 2 & 3 & 4 & 5 \\
			\midrule
			\multirow{40}{*}{\rotatebox{90}{\scalebox{0.75}{\textbf{Top-$3$ Detection Uncertainties}}}}
			& \multirow{5}{*}{FasterRCNN DMM}
			& Trace                  & 51.22 & 53.49 & 54.84 & 57.75 & 62.91 \\
			&   & Determinant            & 48.63 & 47.15 & 47.18 & 47.42 & 51.31 \\
			&   & Entropy (Regression)   & 50.71 & 52.55 & 53.85 & 57.77 & 62.40 \\
			&   & Confidence Score       & 53.51 & 59.04 & 60.50 & 67.91 & 71.23 \\
			&   & Entropy (Classification)& 55.82 & 59.30 & 63.76 & 70.17 & 74.93 \\
			\cmidrule(lr){2-8}
			& \multirow{5}{*}{FasterRCNN ES}
			& Trace                  & 50.68 & 50.86 & 52.32 & 53.79 & 60.63 \\
			&   & Determinant            & 48.52 & 48.13 & 46.47 & 51.14 & 53.65 \\
			&   & Entropy (Regression)   & 52.87 & 53.36 & 56.70 & 57.41 & 61.34 \\
			&   & Confidence Score       & 55.18 & 58.82 & 63.10 & 64.94 & 73.20 \\
			&   & Entropy (Classification)& 54.70 & 59.97 & 62.84 & 68.47 & 77.60 \\
			\cmidrule(lr){2-8}
			& \multirow{5}{*}{FasterRCNN NLL}
			& Trace                  & 52.84 & 51.43 & 53.40 & 54.91 & 58.77 \\
			&   & Determinant            & 47.62 & 46.72 & 49.00 & 43.65 & 51.91 \\
			&   & Entropy (Regression)   & 51.65 & 51.66 & 55.72 & 56.90 & 64.87 \\
			&   & Confidence Score       & 54.14 & 56.22 & 59.59 & 66.34 & 74.65 \\
			&   & Entropy (Classification)& 53.31 & 59.65 & 64.35 & 70.43 & 77.33 \\
			\cmidrule(lr){2-8}
			& \multirow{5}{*}{DETR DMM}
			& Trace                  & 50.03 & 50.50 & 50.47 & 51.21 & 51.63 \\
			&   & Determinant            & 52.01 & 53.71 & 54.50 & 57.29 & 56.99 \\
			&   & Entropy (Regression)   & 49.78 & 50.87 & 48.74 & 49.67 & 50.62 \\
			&   & Confidence Score       & 53.79 & 56.10 & 59.04 & 66.47 & 71.17 \\
			&   & Entropy (Classification)& 55.40 & 57.82 & 62.83 & 70.09 & 75.82 \\
			\cmidrule(lr){2-8}
			& \multirow{5}{*}{DETR ES}
			& Trace                  & 48.73 & 51.12 & 53.47 & 55.49 & 56.07 \\
			&   & Determinant            & 45.82 & 39.34 & 34.04 & 32.80 & 31.50 \\
			&   & Entropy (Regression)   & 49.94 & 55.67 & 58.25 & 61.90 & 63.13 \\
			&   & Confidence Score       & 51.31 & 56.26 & 58.92 & 64.96 & 70.05 \\
			&   & Entropy (Classification)& 57.50 & 57.78 & 62.59 & 68.43 & 74.73 \\
			\cmidrule(lr){2-8}
			& \multirow{5}{*}{DETR NLL}
			& Trace                  & 51.14 & 54.83 & 58.96 & 59.68 & 63.61 \\
			&   & Determinant            & 45.81 & 45.06 & 47.58 & 51.58 & 51.93 \\
			&   & Entropy (Regression)   & 54.36 & 56.09 & 55.81 & 62.89 & 63.06 \\
			&   & Confidence Score       & 54.59 & 58.09 & 60.53 & 66.66 & 69.54 \\
			&   & Entropy (Classification)& 55.11 & 59.33 & 63.08 & 69.14 & 74.09 \\
			\cmidrule(lr){2-8}
			& \multirow{5}{*}{RetinaNet DMM}
			& Trace                  & 53.04 & 53.18 & 55.46 & 59.73 & 64.39 \\
			&   & Determinant            & 49.32 & 51.49 & 47.22 & 51.91 & 49.41 \\
			&   & Entropy (Regression)   & 51.44 & 53.18 & 56.31 & 59.69 & 64.67 \\
			&   & Confidence Score       & 53.55 & 58.10 & 61.77 & 67.49 & 77.19 \\
			&   & Entropy (Classification)& 51.34 & 49.72 & 51.92 & 53.83 & 51.49 \\
			\cmidrule(lr){2-8}
			& \multirow{5}{*}{RetinaNet ES}
			& Trace                  & 52.61 & 55.31 & 57.14 & 60.29 & 65.44 \\
			&   & Determinant            & 49.65 & 50.57 & 51.29 & 52.66 & 56.39 \\
			&   & Entropy (Regression)   & 52.73 & 55.15 & 57.67 & 60.82 & 66.51 \\
			&   & Confidence Score       & 52.55 & 56.77 & 60.72 & 68.10 & 75.28 \\
			&   & Entropy (Classification)& 52.20 & 51.91 & 52.97 & 53.28 & 51.85 \\
			\cmidrule(lr){2-8}
			& \multirow{5}{*}{RetinaNet NLL}
			& Trace                  & 53.11 & 54.25 & 55.59 & 58.46 & 65.18 \\
			&   & Determinant            & 49.48 & 51.09 & 50.94 & 54.13 & 58.23 \\
			&   & Entropy (Regression)   & 53.37 & 54.12 & 57.55 & 62.49 & 66.10 \\
			&   & Confidence Score       & 52.91 & 56.93 & 61.51 & 69.69 & 77.65 \\
			&   & Entropy (Classification)& 50.83 & 50.06 & 51.23 & 51.09 & 51.15 \\
			\bottomrule 
		\end{tabular}
	}
	 \caption{AUROC scores for top-$3$ detection-uncertainty signals~\cite{Oksuz_CVPR_2023} using the probabilistic detector framework of~\cite{Harakeh_NeurIPS_2021} on synthetically corrupted COCO variants.}
	\label{tab:top3_uncertainties_detectors}
\end{table}

\subsection{Feature-space OoD Baselines}
\label{app:fsood}

We additionally compare against feature-space OoD monitors adapted to object detection in order to test whether generic distribution-shift scoring on detector activations is sufficient for degradation monitoring. For all methods, we use frozen multi-scale YOLOv10-m features from \(P3\), \(P4\), and \(P5\). Descriptor-based methods pool and concatenate the selected feature maps into an image-level representation before applying the respective OoD scoring rule. All methods are fitted using in-distribution training data only, and larger scores indicate stronger deviation from the nominal feature distribution.

\paragraph{GRAM.}
GRAM is applied to statistics derived from the selected detector feature maps and scores test samples by their deviation from statistics estimated on in-distribution training data.

\paragraph{Mahalanobis.}
Mahalanobis scoring is applied to the pooled multi-scale image descriptor using statistics estimated from in-distribution training features.

\paragraph{KNN.}
KNN scoring compares the pooled multi-scale image descriptor with stored in-distribution training descriptors in feature space.

\paragraph{ViM.}
ViM is applied to the same pooled multi-scale image representation using parameters estimated from the in-distribution training descriptors.

\subsection{Mixed-Dataset Evaluation}
To further assess robustness to content and dataset variation, we evaluate the degradation manifold under mixed-dataset conditions. In this setting, pristine images are sampled from a combination of datasets, while degraded images originate from the degraded counterpart of the same combined pool. To avoid dataset-imbalance effects, the number of COCO samples is subsampled to match the size of the paired secondary dataset.

This protocol introduces simultaneous variation in scene content, acquisition characteristics, and degradation severity. Importantly, the monitor is not trained to distinguish datasets, but to capture degradation-induced deviation from nominal imaging conditions.

As shown in Table~\ref{tab:auroc_per_level_mixed_datasets}, the degradation manifold maintains strong pristine--degraded separability across severity levels under mixed-dataset conditions. The consistently high AUROC indicates that the learned representation remains aligned with degradation structure rather than dataset identity despite concurrent variation in scene content and acquisition characteristics.

\begin{table*}[h!]
	\centering
	\aurocTableSetup
	\begin{tabular}{l ccccc} 
		\toprule			
		\multicolumn{6}{l}{\textcolor{Fraunhoferred}{\textbf{Degradation Manifold (Ours)}}} \\
		\midrule
		& \multicolumn{5}{c}{\underline{Severity levels AUROC $\uparrow$}} \\		Datasets & 1 & 2 & 3 & 4 & 5\\	
		\midrule
		COCO~\cite{Lin_ECCV_2014} \& VISDRONE~\cite{zhu2021detection} & 83.54& 89.91 & 90.03 & 95.06 & 95.95 \\		
		COCO~\cite{Lin_ECCV_2014} \& KITTI~\cite{Geiger_CVPR_2012}       & 89.27 & 92.44 & 93.86 & 96.40 & 97.21\\		
		COCO~\cite{Lin_ECCV_2014} \& DETRAC~\cite{Wen_CVIU_2020}                          & 92.26 & 93.29 & 94.92  & 98.62&  98.97     \\	
		COCO~\cite{Lin_ECCV_2014} \& FLIR (VIS)~\cite{flirDataset} & 82.45 &88.92  & 89.32 & 94.32 & 95.97 \\	
		COCO~\cite{Lin_ECCV_2014} \& UAVDT~\cite{Du_ECCV_2018}& 89.77 &92.12  & 92.39  & 97.51 & 98.21 \\	
		\bottomrule 
	\end{tabular}
	\caption{
		AUROC for pristine–degraded separability under mixed-dataset evaluation.
		Pristine samples are drawn from combined datasets, while degraded samples originate from a degraded version of the datasets.
		COCO is subsampled for class-balanced comparison.
		Results demonstrate stable degradation-aware separability under simultaneous content and degradation variation.
	}
	\label{tab:auroc_per_level_mixed_datasets}
\end{table*}

\subsection{Controlled Baselines for Degradation Supervision}
\label{app:controlled_degradation_baselines}

To determine whether the performance of the degradation manifold can be
explained by simple image-level fidelity cues, information already present in
pretrained detector features, or direct access to degradation supervision, we
evaluate four complementary controlled baseline families on the same
pristine--degraded COCO protocol as DM.

\paragraph{Training-free low-level fidelity baselines.}
We first evaluate simple image-level cues that require neither detector features
nor degradation-specific learning. Entropy uses the pristine-normalized
deviation of grayscale Shannon entropy, BRISQUE uses the standard pretrained
quality score, and FFT combined uses the pristine-normalized deviations of
high-frequency power and spectral-centroid statistics. Entropy and FFT are
calibrated using pristine COCO training images only; all scores are computed
directly from image pixels before detector preprocessing.

\paragraph{Raw detector-feature prototype.}
We next use frozen pretrained YOLOv10-m features without degradation-specific
training or a learned manifold head. Selected feature maps are globally average
pooled and $\ell_2$-normalized. The single-layer variant uses the final selected
backbone layer (768-D), while the multi-layer variant concatenates normalized
features from six detector layers (2256-D). A pristine prototype is computed as
the normalized mean representation of clean COCO training images, and
degradation is scored by cosine distance to this prototype.

\paragraph{Matched binary degradation classifier.}
We replace the degradation-manifold objective with direct binary
pristine/degraded classification while retaining the same best-performing
multi-layer detector-feature configuration, training images, and cascaded
synthetic degradation compositions used by DM. Thus, this comparison changes
the learning objective rather than the available detector features or
degradation data. The classifier receives explicit pristine/degraded supervision
and is evaluated without adaptation on the same unseen robustness-style
corruptions as DM.

\paragraph{Degradation-strength classifier.}
We additionally evaluate explicit supervision of degradation strength using the
best-performing frozen YOLOv10-m multi-layer feature configuration of DM.
The manifold objective is replaced by a six-class classifier predicting
\emph{pristine} or one of five degradation-strength levels. To obtain an
unambiguous target, each training image yields one pristine view and five views
generated with the same randomly sampled degradation type at increasing
ARNIQA-style severity levels. This necessarily differs from the cascaded DM
training protocol, where independently parameterized operators may be combined
and therefore do not define a unique scalar severity label. At inference, the
predicted severity itself is not used; instead, we define
\begin{equation}
	S_{\mathrm{cls}}(\mathbf{x})
	= 1 - P(\mathrm{pristine}\mid\mathbf{x}).
\end{equation}
The classifier reuses the strongest DM feature configuration without additional
classifier-specific architecture search.

\begin{table}[h!]
	\centering
	\aurocTableSetup
	\begin{tabular}{l ccccc}
		\toprule
		& \multicolumn{5}{c}{\underline{Severity levels AUROC $\uparrow$}} \\
		Method & 1 & 2 & 3 & 4 & 5 \\
		\midrule
		Entropy
		& 54.55 & 56.57 & 57.17 & 58.94 & 59.69 \\
		BRISQUE
		& 74.69 & 78.55 & 79.11 & 81.41 & 82.51 \\
		FFT combined
		& 56.62 & 61.01 & 63.40 & 65.79 & 67.89 \\
		\midrule
		Raw YOLOv10-m (last layer)
		& 53.75 & 55.02 & 57.38 & 62.73 & 64.78 \\
		Raw YOLOv10-m (multi-layer)
		& 63.58 & 68.19 & 74.88 & 77.80 & 80.41 \\
		Binary pristine/degraded classifier
		& 85.39 & 85.93 & 86.37 & 88.59 & 90.19 \\
		Degradation-strength classifier
		& 84.87 & 84.98 & 85.47 & 87.89 & 89.49 \\
		\midrule
		\textcolor{Fraunhoferred}{\textbf{Degradation Manifold} DM (Ours)}
		& \textbf{88.64} & \textbf{89.70} & \textbf{89.75}
		& \textbf{95.28} & \textbf{97.14} \\
		\bottomrule
	\end{tabular}
	\caption{Controlled baselines on the same pristine--degraded COCO
		evaluation as DM. The first group contains training-free image-level
		fidelity baselines. Raw detector-feature variants use no
		degradation-specific training. The binary classifier uses the same
		multi-layer detector features and cascaded degradation data as DM but
		replaces manifold learning with direct pristine/degraded supervision.
		The degradation-strength classifier uses the same feature configuration
		with explicit single-degradation strength supervision.}
	\label{tab:controlled_degradation_baselines}
\end{table}

The training-free and raw detector-feature baselines remain substantially below
DM, showing that neither simple pixel-level fidelity cues nor degradation
information already present in pretrained detector representations is sufficient
to explain the full result. Direct supervision provides a substantially stronger
signal: both the binary and degradation-strength classifiers achieve high
pristine--degraded separability, but remain below DM at every severity level.
For the strength classifier, the gap increases from $3.77$ AUROC points at
severity~1 to $7.65$ points at severity~5.

These results are consistent with the component ablation in
Table~\ref{tab:ablation_deg_mani}: without hard-negative mining, DM lies closer
to the directly supervised classifiers, whereas the full configuration achieves
the strongest separability. Together, the controls indicate that the performance
of DM cannot be attributed solely to simple image statistics, native detector
features, or access to degradation supervision, but instead results from their
degradation-aware organization through the learned representation, multi-layer
readout, and hard-negative construction.

\subsection{Ablation Study}
Table~\ref{tab:ablation_deg_mani} analyzes the impact of multi-layer readout, attention-based pooling, and hard negative mining using a YOLOv10-m backbone. The full configuration consistently achieves the strongest separability. Removing multi-layer aggregation or hard negatives leads to systematic drops in AUROC, highlighting the importance of capturing degradation across multiple feature hierarchies and enforcing fine-grained separation between related fidelity regimes.

\begin{table*}[h!]
	\aurocTableSetup
	\begin{tabular}{l ccccc}
		\toprule			
		\multicolumn{6}{l}{\textcolor{Fraunhoferred}{\textbf{Degradation Manifold (Ours)}}} \\
		\midrule
		& \multicolumn{5}{c}{\underline{Severity levels AUROC $\uparrow$}} \\
		Variant & 1 & 2 & 3 & 4 & 5 \\
		\midrule
		Full model (multi-layer + attention + hard neg.) 
		& 88.64 & 89.70 & 89.75 & 95.28 & 97.14 \\
		Last backbone layer only 
		& 81.87 & 85.13 & 87.56 & 89.13 & 92.75 \\
		Global average pooling (no attention) 
		& 85.51 & 88.44 & 89.49 & 95.01 & 96.30 \\
		Without hard negative mining 
		& 86.00 & 87.17 & 87.58 & 90.41 & 91.54 \\
		\bottomrule
	\end{tabular}
	\caption{Ablation study of the proposed degradation manifold using a YOLOv10-m backbone. 
		We analyze the impact of multi-layer readout, attention-based pooling, and hard negative mining on AUROC across severity levels. 
		The full configuration consistently achieves the strongest separability.}
	\label{tab:ablation_deg_mani}
\end{table*}

\FloatBarrier
\subsection{Single-Degradation Analysis}
To further characterize the learned representation, we analyze separability for individual corruption types. Tables~\ref{tab:degradation_strengths_all_metrics_Agnolucci} and~\ref{tab:degradation_strengths_all_metrics_hendrycks} report AUROC for pristine--degraded separability per degradation type and severity level for (i) IQA-style degradations from~\cite{Agnolucci_WACV_2024} and (ii) robustness-style corruptions from~\cite{Hendrycks_ICLR_2019}.

Although the degradation manifold is trained using degradation compositions inspired by~\cite{Agnolucci_WACV_2024}, it transfers well to evaluation corruptions with different operators and parameterizations. For most degradation types, separability increases with corruption strength, indicating that the learned embedding captures structured degradation geometry beyond the exact transformations observed during training. The detailed results also reveal where this ordering is weaker, motivating the geometric analysis in Section~\ref{app:geometry}.

\begin{table*}[h!]
	\centering
	\aurocTableSetup
	\begin{tabular}{l|c: c: c: c : c |}
		\hline
		& \multicolumn{5}{c|}{} \\
		& \multicolumn{5}{c|}{\underline{Severity Levels AUROC $\uparrow$}} \\
		Degradation Function \cite{Agnolucci_WACV_2024}
		& 1 & 2 & 3 & 4 & 5 \\
		\hline
		\rowcolor{Set3-08!20}
		\degfunc{Gaussian Blur} 
		& 48.15 & 61.56 & 85.20 & 99.84 & 99.89 \\
		\rowcolor{gray!20}
		\degfunc{Lens Blur}
		& 98.02 & 98.45 & 98.90 & 99.20 & 99.50 \\
		\rowcolor{Set3-08!20}
		\degfunc{Motion Blur}  
		& 48.93 & 93.43 & 99.62 & 99.92 & 99.95 \\
		\rowcolor{gray!20}
		\degfunc{White Noise}
		& 98.38 & 99.28 & 99.60 & 99.75 & 99.85 \\
		\rowcolor{Set3-08!20}
		\degfunc{White Noise in Color Component}
		& 85.23 & 98.72 & 99.36 & 99.70 & 99.85 \\
		\rowcolor{gray!20}
		\degfunc{Impulse Noise}
		& 80.02 & 98.93 & 99.70 & 99.80 & 99.88 \\
		\rowcolor{Set3-08!20}
		\degfunc{Multiplicative Noise}
		& 92.25 & 98.02 & 99.16 & 99.40 & 99.60 \\
		\rowcolor{gray!20}
		\degfunc{Brighten}
		& 48.68 & 47.73 & 48.88 & 81.28 & 97.16 \\
		\rowcolor{Set3-08!20}
		\degfunc{Darken}
		& 48.88 & 49.42 & 49.98 & 50.57 & 98.13 \\
		\rowcolor{gray!20}
		\degfunc{Mean Shift}
		& 48.16 & 50.37 & 50.70 & 62.20 & 64.77 \\			
		\rowcolor{Set3-08!20}
		\degfunc{Color Diffusion}
		& 96.51 & 98.08 & 98.04 & 98.56 & 98.15 \\
		\rowcolor{gray!20}
		\degfunc{Color Shift}
		& 95.60 & 98.75 & 98.85 & 99.04 & 99.20 \\
		\rowcolor{Set3-08!20}
		\degfunc{Color Saturation 1}
		& 54.53 & 80.02 & 76.53 & 72.30 & 79.84 \\
		\rowcolor{gray!20}
		\degfunc{Color Saturation 2}
		& 48.81 & 75.74 & 93.54 & 98.17 & 98.50 \\
		\rowcolor{Set3-08!20}
		\degfunc{JPEG}
		& 99.59 & 99.70 & 99.80 & 99.85 & 99.90 \\
		\rowcolor{gray!20}
		\degfunc{JPEG2000}
		& 98.83 & 99.75 & 99.85 & 99.92 & 100.00 \\
		\rowcolor{Set3-08!20}
		\degfunc{Jitter} 
		& 60.29 & 97.35 & 99.81 & 99.89 & 99.79\\
		\rowcolor{gray!20}
		\degfunc{Non-Eccentricity Patch}
		& 85.95 & 94.67 & 96.89 & 97.66 & 98.05 \\
		\rowcolor{Set3-08!20}
		\degfunc{Pixelate} 
		& 97.17 & 99.92 & 99.95 & 99.96 & 100.00 \\			
		\rowcolor{gray!20}
		\degfunc{Quantization}
		& 98.09 & 99.20 & 99.69 & 99.83 & 99.90 \\
		\rowcolor{Set3-08!20}
		\degfunc{Color Block} 
		& 86.04 & 94.54 & 97.12 & 98.04 & 99.22 \\			
		\rowcolor{gray!20}			
		\degfunc{High Sharpen} 
		& 98.25 & 99.62 & 99.67 & 99.80 & 99.88 \\
		\rowcolor{Set3-08!20}
		\degfunc{Linear Contrast Change}
		& 48.23 & 59.96 & 51.24 & 84.43 & 56.86\\			
		\rowcolor{gray!20}
		\degfunc{Nonlinear Contrast Change}
		& 60.39 & 97.10 & 99.17 & 99.71 & 99.78 \\
		\bottomrule
	\end{tabular}
	\caption{AUROC for pristine--degraded separability under individual IQA-style synthetic degradations on COCO, following~\cite{Agnolucci_WACV_2024}.}	
	\label{tab:degradation_strengths_all_metrics_Agnolucci}
\end{table*}

\begin{table*}[h!]
	\centering
	\aurocTableSetup
	\begin{tabular}{l|c: c: c: c : c |}	
		\toprule
		& \multicolumn{5}{c}{\underline{Severity Levels AUROC $\uparrow$}} \\
		Degradation Function \cite{Hendrycks_ICLR_2019}
		& 1 & 2 & 3 & 4 & 5 \\
		\midrule
		\rowcolor{Set3-08!20}
		\degfunc{Gaussian Noise}      & 99.91 & 99.94 & 99.93 & 99.94 & 99.92 \\
		\rowcolor{gray!20}
		\degfunc{Shot Noise}          & 99.63 & 99.74 & 99.83 & 99.81 & 99.85 \\
		\rowcolor{Set3-08!20}
		\degfunc{Impulse Noise}       & 99.96 & 99.94 & 99.92 & 99.91 & 99.90 \\
		\rowcolor{gray!20}
		\degfunc{Speckle Noise}       & 99.70 & 99.72 & 99.73 & 99.72 & 99.71 \\
		\rowcolor{Set3-08!20}
		\degfunc{Defocus Blur}        & 99.86 & 99.87 & 99.88 & 99.87 & 99.86 \\
		\rowcolor{gray!20}
		\degfunc{Gaussian Blur}       & 99.76 & 99.78 & 99.79 & 99.78 & 99.77 \\
		\rowcolor{Set3-08!20}
		\degfunc{Glass Blur}          & 99.98 & 99.97 & 99.96 & 99.95 & 99.94 \\
		\rowcolor{gray!20}
		\degfunc{Motion Blur}         & 99.61 & 99.63 & 99.65 & 99.64 & 99.62 \\
		\rowcolor{Set3-08!20}
		\degfunc{Zoom Blur}           & 99.05 & 99.10 & 99.15 & 99.18 & 99.20 \\
		\rowcolor{gray!20}
		\degfunc{Fog}                 & 98.67 & 98.89 & 99.04 & 98.96 & 99.18 \\
		\rowcolor{Set3-08!20}
		\degfunc{Frost}               & 87.51 & 93.31 & 94.89 & 94.76 & 94.80 \\
		\rowcolor{gray!20}
		\degfunc{Snow}                & 82.90 & 88.40 & 92.15 & 96.72 & 99.31 \\
		\rowcolor{Set3-08!20}
		\degfunc{Spatter}             & 54.95 & 60.55 & 86.97 & 99.77 & 99.92 \\
		\rowcolor{gray!20}
		\degfunc{Brightness}          & 54.74 & 76.28 & 89.53 & 96.29 & 98.55 \\
		\rowcolor{Set3-08!20}
		\degfunc{Contrast}            & 98.78 & 99.16 & 99.32 & 99.41 & 99.47 \\
		\rowcolor{gray!20}
		\degfunc{Saturate}            & 71.40 & 79.31 & 81.53 & 99.44 & 99.55 \\
		\rowcolor{Set3-08!20}
		\degfunc{JPEG}                & 99.99 & 99.98 & 99.97 & 99.96 & 99.95 \\
		\rowcolor{gray!20}
		\degfunc{Pixelate}            & 99.96 & 99.95 & 99.94 & 99.93 & 99.92 \\
		\rowcolor{Set3-08!20}
		\degfunc{Elastic Transform}   & 92.72 & 92.83 & 90.84 & 84.89 & 88.96 \\
		\bottomrule
	\end{tabular}
	\caption{AUROC for pristine--degraded separability under individual robustness-style synthetic degradations on COCO, following~\cite{Hendrycks_ICLR_2019,Michaelis_NeurIPSW_2019}.}
	\label{tab:degradation_strengths_all_metrics_hendrycks}
\end{table*}

\subsection{Additional Light-Weather Evaluation}
In addition to the heavy-weather conditions used in the main paper, we also evaluate DSERT-RoLL~\cite{Cho_CVPR_2026} on \texttt{light snow} and \texttt{light rain}. These conditions define a more challenging and more ambiguous natural-shift setting, since their visual impact is often weaker and less scene-dominant than the corresponding heavy-weather subsets. We therefore treat them as auxiliary results rather than part of the main benchmark. The substantially lower AUROC without weather augmentation, especially for \texttt{light snow}, is consistent with this weaker ID/OoD separation. Synthetic weather augmentation still improves performance, but the gain remains smaller than for the heavy-weather subsets, reflecting the softer shift boundary.

\begin{table}[h!]
	\centering
	\aurocTableSetup
	\begin{tabular}{l l c c}
		\toprule
		\multicolumn{4}{l}{\textcolor{Fraunhoferred}{\textbf{Degradation Manifold (Ours)}}} \\
		\midrule
		Dataset & Subset (OoD) & Weather aug. (train) & AUROC $\uparrow$ \\
		\midrule
		DSERT-RoLL~\cite{Cho_CVPR_2026} & \texttt{light snow} & \xmark & 51.32 \\
		DSERT-RoLL~\cite{Cho_CVPR_2026} & \texttt{light snow} & \cmark & 81.02 \\
		DSERT-RoLL~\cite{Cho_CVPR_2026} & \texttt{light rain} & \xmark & 60.32 \\
		DSERT-RoLL~\cite{Cho_CVPR_2026} & \texttt{light rain} & \cmark & 77.62 \\
		\bottomrule
	\end{tabular}
	\caption{Auxiliary DSERT-RoLL results for light-weather conditions. Compared to the heavy-weather subsets used in the main paper, light weather induces a more ambiguous natural-shift boundary and therefore yields weaker pristine--degraded separability.}
	\label{tab:auroc_naturalshifts_dataset_weather_light}
\end{table}

\subsection{Downstream Task Relevance of Degradation Separability}
\label{app:auroc_map_correlation}

To assess whether degradation separability is related to downstream task degradation, we correlate the AUROC of the degradation monitor with the relative detection-performance drop of YOLOv10-m under the IQA-style degradations from~\cite{Agnolucci_WACV_2024}. For each degradation $\degfunc{d}$ and native severity level $l \in \{1,\ldots,5\}$, we define the relative drop in mAP@0.5:0.95 as
\begin{equation}
	\Delta \mathrm{mAP}_{d,l} =
	100 \cdot
	\frac{
		\mathrm{mAP}_{\mathrm{clean}} - \mathrm{mAP}_{d,l}
	}{
		\mathrm{mAP}_{\mathrm{clean}}
	}.
\end{equation}
We then correlate $\Delta \mathrm{mAP}_{d,l}$ with the corresponding pristine--degraded AUROC value across all 24 degradations and five severity levels, resulting in 120 paired observations.

\begin{table}[h]
	\centering
	\caption{Correlation between degradation separability, measured by AUROC, and the relative mAP@0.5:0.95 drop of YOLOv10-m across 24 IQA-style degradations and five native severity levels.}
	\label{tab:auroc_map_correlation}
	\begin{tabular}{lcc}
		\toprule
		Correlation measure & Coefficient & $p$-value \\
		\midrule
		Pearson $r$     & 0.405 & $4.46 \times 10^{-6}$ \\
		Spearman $\rho$ & 0.846 & $5.46 \times 10^{-34}$ \\
		Kendall $\tau$  & 0.653 & $5.04 \times 10^{-26}$ \\
		\bottomrule
	\end{tabular}
\end{table}

The results indicate that degradation separability is strongly associated with detector-performance degradation at the rank level. In particular, the high Spearman correlation shows that degradations that are more separable from pristine inputs generally induce larger drops in mAP. The weaker Pearson correlation indicates that this relationship is not linear.

This is expected: AUROC tends to saturate near perfect separability for stronger degradations, whereas the corresponding mAP drop can still vary substantially across degradation families. AUROC should therefore not be interpreted as a calibrated predictor of task failure. Instead, it provides a task-relevant input-fidelity signal: large separability indicates substantial deviation from the nominal imaging regime, which is often, but not linearly, associated with reduced detection performance.

\subsection{Geometric Behavior of the Degradation Manifold}
\label{app:geometry}

To complement the separability results, we analyze the geometry of the learned degradation manifold through the cosine distance to the pristine prototype. For each degradation type, we compute the mean cosine distance over all images in the synthetically corrupted COCO validation variants for severity levels~1--5. We then quantify severity ordering using Spearman rank correlation $\rho$ between native severity level and mean distance.

The learned manifold is clearly severity-aware, but strict monotonicity is not universal. For the IQA-style degradations of~\cite{Agnolucci_WACV_2024}, most structural degradations exhibit strong monotonic or near-monotonic  distance growth with severity. This is particularly evident for motion blur, white noise, multiplicative noise, compression, pixelation, sharpening, and nonlinear contrast changes. Table~\ref{tab:geometry_monotonicity_agnolucci} reports the corresponding per-degradation rank correlations.

For the robustness-style corruptions of~\cite{Hendrycks_ICLR_2019}, the geometric behavior is more heterogeneous. Several noise- and blur-based degradations exhibit saturation or non-monotonic trajectories at higher severity levels. In particular, \degfunc{Gaussian Noise}, \degfunc{Impulse Noise}, \degfunc{Defocus Blur}, and \degfunc{Zoom Blur} show negative rank correlation. Across this benchmark, only 5 out of 19
degradation types are strictly monotonic, highlighting the heterogeneity of the broader robustness-style corruption families.

This behavior should not be interpreted as a failure of the manifold alone. Native severity levels do not always define a perfectly linear progression in detector-relevant information loss, and different degradation families affect signal statistics, perceptual quality, and task performance in different ways. In particular, color-related shifts and selected blur variants may induce only weak geometric progression, either because the detector backbone is comparatively insensitive to them or because the corruption family itself does not define a clean one-dimensional severity trajectory.

Overall, these results support the interpretation of the proposed monitor as a geometry-based indicator of deviation from nominal imaging conditions rather than a universally linear surrogate for perceptual corruption strength or task failure. The manifold preserves a meaningful severity ordering for most degradations aligned with the training regime, while naturally reflecting the heterogeneity of broader robustness-style corruption families.

\begin{table}[h!]
	\centering
	\aurocTableSetup
	\begin{tabular}{lcc}
		\toprule
		Degradation family & Spearman $\rho$ & Behavior \\
		\midrule
		\degfunc{Gaussian Blur} & 0.90 & near-monotonic \\
		\degfunc{Lens Blur} & 0.30 & weak / saturated \\
		\degfunc{Motion Blur} & 1.00 & monotonic \\
		\degfunc{White Noise} & 1.00 & monotonic \\
		\degfunc{White Noise in Color Component} & 1.00 & monotonic \\
		\degfunc{Impulse Noise} & 1.00 & monotonic \\
		\degfunc{Multiplicative Noise} & 1.00 & monotonic \\
		\degfunc{Brighten} & 1.00 & monotonic \\
		\degfunc{Darken} & 1.00 & monotonic \\
		\degfunc{Mean Shift} & 0.90 & near-monotonic \\
		\degfunc{Color Diffusion} & 1.00 & monotonic \\
		\degfunc{Color Shift} & 0.10 & weak \\
		\degfunc{Color Saturation 1} & 0.20 & weak \\
		\degfunc{Color Saturation 2} & 1.00 & monotonic \\
		\degfunc{JPEG} & 1.00 & monotonic \\
		\degfunc{JPEG2000} & 1.00 & monotonic \\
		\degfunc{Jitter} & 0.70 & partially monotonic \\
		\degfunc{Non-Eccentricity Patch} & 1.00 & monotonic \\
		\degfunc{Pixelate} & 1.00 & monotonic \\
		\degfunc{Quantization} & 1.00 & monotonic \\
		\degfunc{Color Block} & 1.00 & monotonic \\
		\degfunc{High Sharpen} & 1.00 & monotonic \\
		\degfunc{Linear Contrast Change} & 0.30 & weak / irregular \\
		\degfunc{Nonlinear Contrast Change} & 1.00 & monotonic \\
		\bottomrule
	\end{tabular}
	\caption{
		Spearman rank correlation between native severity level and mean cosine distance to the pristine prototype for the degradations of~\cite{Agnolucci_WACV_2024}, computed over all images in the synthetically corrupted COCO validation variants.
	}
	\label{tab:geometry_monotonicity_agnolucci}
\end{table}

\begin{table}[h!]
	\centering
	\aurocTableSetup
	\begin{tabular}{lcc}
		\toprule
		Degradation family & Spearman $\rho$ & Behavior \\
		\midrule
		\degfunc{Gaussian Noise} & -0.90 & non-monotonic / saturated \\
		\degfunc{Shot Noise} & -0.70 & non-monotonic / saturated \\
		\degfunc{Impulse Noise} & -0.90 & non-monotonic / saturated \\
		\degfunc{Speckle Noise} & 0.30 & weak / irregular \\
		\degfunc{Defocus Blur} & -0.90 & non-monotonic / saturated \\
		\degfunc{Gaussian Blur} & -0.40 & non-monotonic / saturated \\
		\degfunc{Glass Blur} & 0.70 & weak / partial \\
		\degfunc{Motion Blur} & 0.60 & weak / partial \\
		\degfunc{Zoom Blur} & -0.80 & non-monotonic / saturated \\
		\degfunc{Fog} & 0.60 & weak / partial \\
		\degfunc{Frost} & 0.70 & weak / partial \\
		\degfunc{Snow} & 0.90 & near-monotonic \\
		\degfunc{Spatter} & 1.00 & monotonic \\
		\degfunc{Brightness} & 1.00 & monotonic \\
		\degfunc{Contrast} & 1.00 & monotonic \\
		\degfunc{Saturate} & 0.70 & weak / irregular \\
		\degfunc{JPEG Compression} & 1.00 & monotonic \\
		\degfunc{Pixelate} & 1.00 & monotonic \\
		\degfunc{Elastic Transform} & -0.70 & non-monotonic / irregular \\
		\bottomrule
	\end{tabular}
	\caption{
		Spearman rank correlation between native severity level and mean cosine distance to the pristine prototype for robustness-style corruptions from~\cite{Hendrycks_ICLR_2019}, computed over all images in the synthetically corrupted COCO validation variants.
	}
	\label{tab:geometry_monotonicity_hendrycks}
\end{table}

\subsection{Score Distribution}
Fig.~\ref{fig:dist_score} visualizes the distribution of $S_{\mathrm{deg}}(\mathbf{x})$ for ID and OoD samples at different corruption severity levels using the proposed degradation-manifold monitor with a YOLOv10-m backbone. As severity increases, the score distributions separate more clearly, which is consistent with the AUROC trends reported in the main paper.

\renewcommand{\subw}{0.45\textwidth}
\begin{figure*}[h!]
	\centering
	\begin{subfigure}{\subw}
		\includegraphics[width=\linewidth]{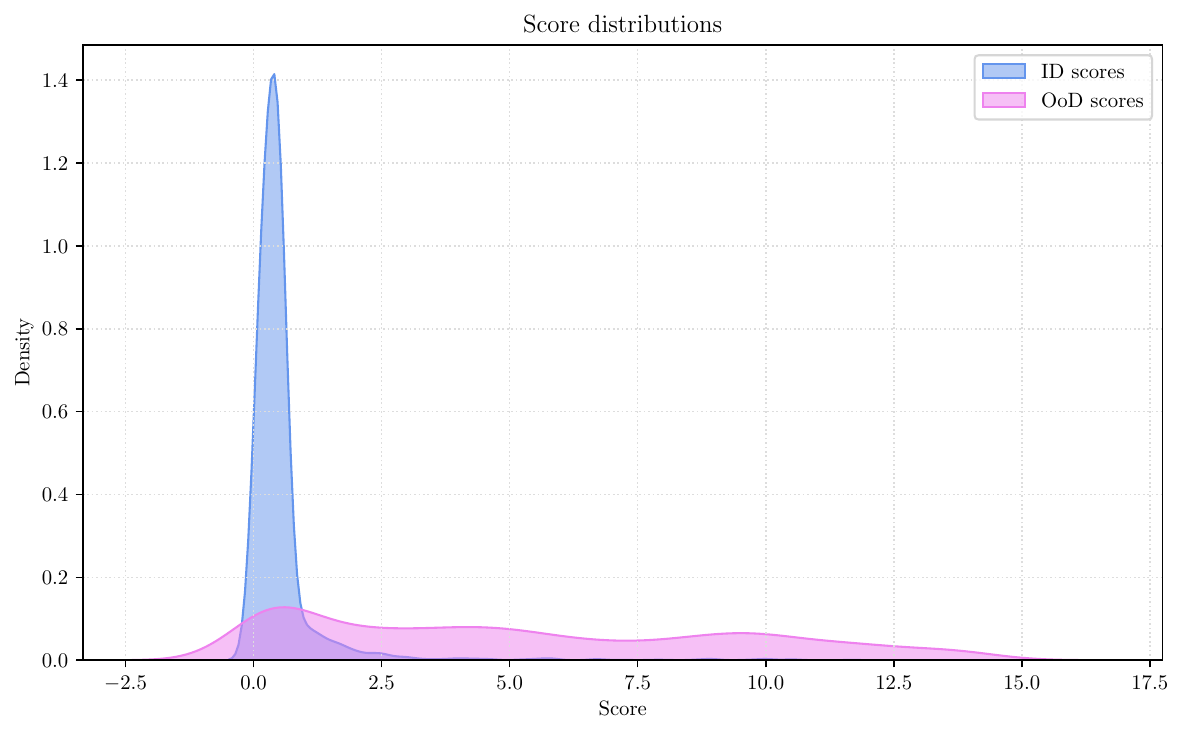}
		\caption{COCO-C1}
	\end{subfigure}
	\hfill
	\begin{subfigure}{\subw}
		\includegraphics[width=\linewidth]{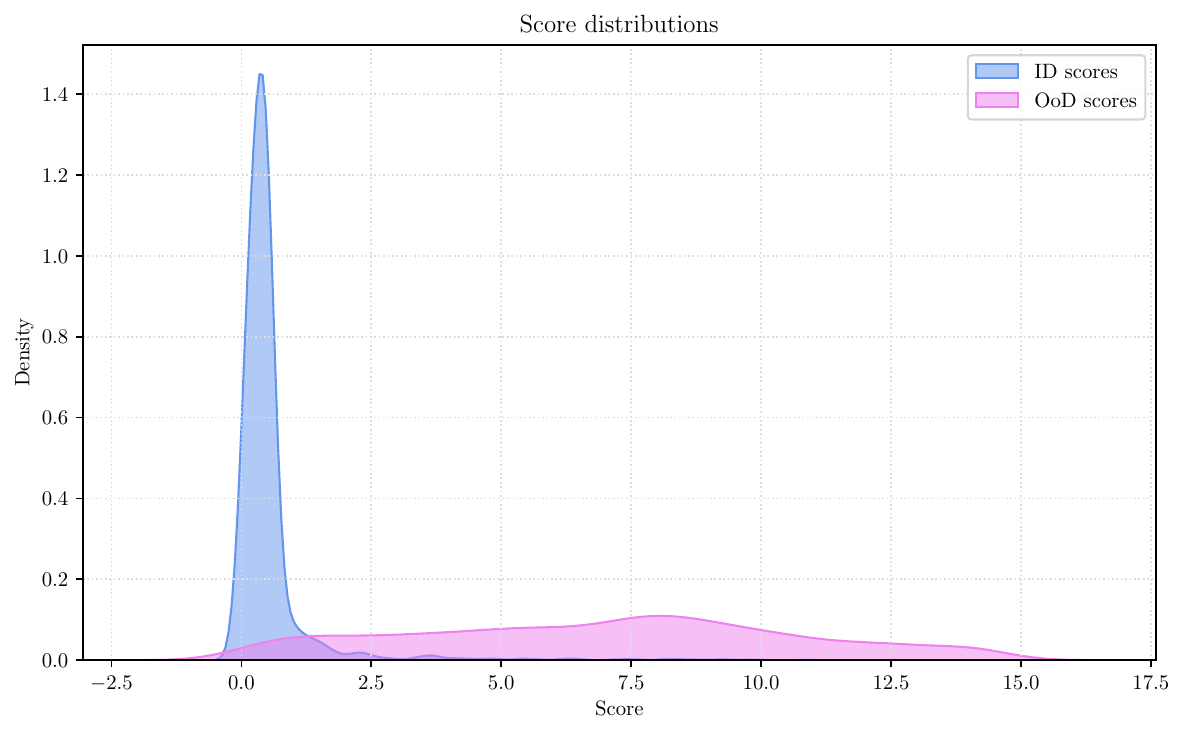}
		\caption{COCO-C5}
		\label{fig:short-c}
	\end{subfigure}
	\\
	\caption{Distribution of $S_{\mathrm{deg}}(\mathbf{x})$ for ID and OoD samples at different corruption severity levels using the proposed degradation-manifold monitor with a YOLOv10-m backbone.}
	\label{fig:dist_score} 
	\vspace{-0.5cm} 
\end{figure*}

\subsection{Operating-Point Metrics}
In addition to AUROC, we report two operating-point metrics commonly used in OoD and covariate-shift evaluation~\cite{Hendrycks_ICLR_2019,Liang_ICLR_2018}. These metrics characterize monitor behavior at practically relevant decision thresholds.

\textbf{False Positive Rate at 95\% True Positive Rate (FPR@95TPR).}
This metric measures the false alarm rate on pristine samples when the threshold is chosen such that $95\%$ of degraded samples are correctly identified:
\[
\mathrm{FPR}(\tau)=\frac{\mathrm{FP}}{\mathrm{FP}+\mathrm{TN}},
\qquad \text{with } \mathrm{TPR}(\tau)=0.95.
\]
Lower values indicate fewer false alarms at high degradation-detection recall.

\textbf{False Negative Rate at 95\% True Negative Rate (FNR@95TNR).}
This metric measures the proportion of degraded samples missed when the threshold is chosen such that $95\%$ of pristine samples are correctly accepted:
\[
\mathrm{FNR}(\tau)=\frac{\mathrm{FN}}{\mathrm{FN}+\mathrm{TP}},
\qquad \text{with } \mathrm{TNR}(\tau)=0.95.
\]
Lower values indicate better degradation detection under a constrained false-alarm regime.

These metrics complement AUROC: while AUROC evaluates ranking quality across all possible thresholds, FPR@95TPR and FNR@95TNR reflect performance at deployment-relevant operating points.

\begin{table*}[h!]
	\centering
	\aurocTableSetup
	\resizebox{0.9\linewidth}{!}{%
		\begin{tabular}{l|c c c c c|c c c c c|}
			\toprule			
			& \multicolumn{5}{c|}{FNR@95 TNR [\%] $\downarrow$} 	
			& \multicolumn{5}{c|}{FPR@95 TPR [\%] $\downarrow$} \\		
			& \multicolumn{5}{c|}{\underline{Severity levels}} 
			& \multicolumn{5}{c|}{\underline{Severity levels}}\\	
			Model
			& 1 & 2 & 3 & 4 & 5	
			& 1 & 2 & 3 & 4 & 5\\	
			\midrule
			
			\multicolumn{11}{l}{\textbf{Image Quality Assessment (IQA)}}\\		
			\midrule
			
			MANIQA (score-based)
			& 97.72 & 98.26 & 97.62 & 97.69 & 98.00
			& 94.16 & 92.86 & 92.88 & 94.12 & 93.22 \\		
			
			MANIQA (embedding-based)
			& 71.56 & 62.62 & 55.36 & 47.94 & 44.74
			& 92.62 & 89.62 & 89.58 & 86.14 & 79.00\\	
			
			QualiCLIP (score-based)
			& 96.46 & 95.16 & 88.17 & 77.36 & 74.58
			& 95.10 & 94.28 & 90.42 & 87.64 & 86.04\\	
			
			QualiCLIP (embedding-based)
			& 97.04 & 97.12 & 97.55 & 93.66 & 88.94 
			& 89.44 & 84.12 & 84.04 & 79.02 & 73.06\\	
			
			CLIPIQA (score-based)             
			& 97.80 & 97.24 & 96.66 & 89.40 & 81.96 
			& 92.30 & 91.12 & 91.58 & 87.58 & 83.00\\	
			
			CLIPIQA (embedding-based)        
			& 97.24 & 98.06 & 98.50 & 99.24 & 99.10 
			& 95.66 & 95.38 & 95.30 & 93.96 & 91.30\\	
			
			CLIPIQA+ (score-based)             
			& 97.12 & 97.92 & 96.88 & 93.84 & 93.24 
			& 92.22 & 92.06 & 92.26 & 89.12 & 87.88\\	
			
			CLIPIQA+ (embedding-based)        
			& 96.90 & 97.70 & 98.08 & 99.18 & 98.80 
			& 95.66 & 95.58 & 95.00 & 94.08 & 91.46\\	
			
			ARNIQA (score-based)           
			& 92.98 & 91.84 & 88.61 & 88.28 & 86.55 
			& 96.34 & 95.22 & 94.54 & 94.38 & 94.14\\	
			
			ARNIQA (embedding-based)     
			& 55.98 & 48.16 & 44.16 & 38.86 & 37.32 
			& 51.92 & 49.84 & 56.46 & 52.62 & 49.62\\		
			
			\midrule
			
			{\textbf{Top-$3$ Detection Uncertainties (\cite{Oksuz_CVPR_2023,Harakeh_NeurIPS_2021})}}\\
			\midrule
			
			Trace (RetinaNet ES)        
			& 92.12 & 89.50 & 87.73 & 84.90 & 84.91
			& 94.50 & 95.08 & 93.48 & 93.74 & 91.64\\	
			
			Trace (DETR NLL)        
			& 89.66 & 86.70 & 82.23 & 78.38 & 74.26
			& 94.94 & 94.42 & 93.98 & 94.06 & 91.92\\	
			
			Determinant (RetinaNet NLL)           
			& 93.60 & 89.84 & 89.10 & 87.83 & 83.85
			& 95.76 & 96.14 & 97.20 & 96.58 & 96.38\\	
			
			Determinant (RetinaNet ES)            
			& 92.88 & 91.90 & 89.56 & 86.09 & 87.21
			& 95.78 & 96.62 & 97.28 & 97.34 & 97.92\\	
			
			Entropy (Regression) (RetinaNet ES)  
			& 90.99 & 90.37 & 88.93 & 85.24 & 83.44
			& 95.08 & 94.60 & 94.10 & 93.60 & 91.10\\	
			
			Entropy (Regression) (FasterRCNN NLL)  
			& 94.67 & 92.30 & 90.15 & 87.25 & 83.69
			& 94.70 & 95.34 & 94.60 & 93.34 & 91.16\\	
			
			Confidence Score (RetinaNet NLL)      
			& 90.02 & 85.76 & 78.81 & 66.53 & 53.27
			& 94.10 & 93.00 & 92.48 & 91.80 & 86.42\\	
			
			Confidence Score (RetinaNet ES)        
			& 91.08 & 86.87 & 80.07 & 67.24 & 54.00
			& 94.66 & 93.58 & 92.58 & 90.84 & 86.66\\	
			
			Entropy (Classification) (FasterRCNN ES)  
			& 87.74 & 80.99 & 75.23 & 66.32 & 57.45
			& 94.06 & 92.88 & 93.37 & 90.62 & 87.58\\	
			
			Entropy (Classification) (FasterRCNN NLL) 
			& 87.66 & 82.44 & 76.92 & 68.55 & 62.29
			& 94.50 & 93.83 & 92.56 & 91.42 & 86.63\\
			
			\midrule
				{\textbf{Feature-space OoD methods}} (\cite{Zimmermann_CVPRW_2026,Becker_ICPR_2026})\\		
			\midrule    	
			
			\midrule  
			
			GRAM \cite{Sastry_ICML_2020} (YOLOv10-m feature maps) 
			& 84.40 & 75.24 & 64.08 & 51.72 & 45.86
			& 88.44 & 82.04 & 78.54 & 70.94 & 62.62 \\		
			
			Mahalanobis \cite{Lee_NEURIPS_2018} (YOLOv10-m feature maps)
			& 79.92 & 73.72 & 69.66 & 62.84 & 57.30
			& 89.08 & 87.70 & 87.26 & 83.90 & 82.78 \\
			
			KNN \cite{Sun_ICML_2022} (YOLOv10-m feature maps)
			& 83.05 & 76.76 & 73.38 & 69.31 & 66.44
			& 88.66 & 86.68 & 84.98 & 78.78 & 74.32 \\
			
			VIM \cite{Wang_2022_CVPR} (YOLOv10-m feature maps)
			& 81.16 & 75.64 & 72.28 & 68.72 & 65.28
			& 88.56 & 86.66 & 85.92 & 82.26 & 80.54 \\
			
			Activation histograms \cite{Becker_ICPR_2026} (YOLOv10-m feature maps)
			& 89.00 & 83.32 & 74.66 & 59.68 & 40.22
			& 85.24 & 77.88 & 74.56 & 64.76 & 60.42 \\				
			
			\midrule		
			\textbf{Normalizing Flows} (adapted from \cite{lotfi2025medoodflow})\\
			\midrule                
			
			NF pooled feature maps (C4, C5)            
			& 96.70 & 95.95 & 94.70 & 93.02 & 87.36
			& 95.64 & 94.80 & 94.20 & 94.60 & 93.72\\
			
			NF pooled feature maps (C1, C2, C3)        
			& 90.10 & 85.84 & 77.60 & 68.47 & 59.80
			& 93.20 & 91.84 & 91.40 & 88.68 & 87.40\\
			
			NF pooled feature maps (all; C1--C5)       
			& 90.14 & 87.60 & 85.60 & 73.54 & 62.80 
			& 95.01 & 94.96 & 94.72 & 93.18 & 91.14\\            
			
			M-NF pooled feature maps (C4, C5)          
			& 97.32 & 95.70 & 95.30 & 90.30 & 84.20
			& 95.30 & 95.90 & 95.00 & 93.00 & 92.60\\
			
			M-NF on backbone features (C1, C2, C3)      
			& 87.05	& 78.16 & 67.90 & 68.26 & 59.38 
			& 95.04 & 94.41 & 92.96 & 92.78 & 90.30\\
			
			M-NF on backbone features (all; C1--C5)     
			& 93.38 & 88.90 & 80.30 & 71.30 & 61.20
			& 95.40 & 92.90 & 94.40 & 92.30 & 88.00\\
			
			\midrule
			
			\multicolumn{11}{l}{\textcolor{Fraunhoferred}{\textbf{Degradation Manifold (Ours)}}}\\
			\midrule
			
			DM (YOLOv10-m) 
			& \colorbox{colorFst}{26.28} & \colorbox{colorFst}{19.72} & \colorbox{colorTrd}{21.03} & \colorbox{colorFst}{12.05} & \colorbox{colorFst}{8.75}
			& \colorbox{colorTrd}{57.08} & \colorbox{colorTrd}{35.15} & \colorbox{colorSnd}{39.00} & \colorbox{colorFst}{12.48} & \colorbox{colorSnd}{8.80}\\		
			
			DM (RT-DETR-l)        
			& \colorbox{colorSnd}{29.53} & \colorbox{colorSnd}{21.95} & \colorbox{colorSnd}{20.65} & \colorbox{colorTrd}{13.03} & \colorbox{colorSnd}{8.93}
			& 48.25 & \colorbox{colorSnd}{36.98} & \colorbox{colorFst}{33.75} & \colorbox{colorTrd}{14.50} & \colorbox{colorFst}{8.75}\\		
			
			DM (YOLOv9-m)                          
			& \colorbox{colorTrd}{25.90} & \colorbox{colorTrd}{20.14} & \colorbox{colorFst}{19.62} & \colorbox{colorSnd}{11.66} & \colorbox{colorTrd}{8.30}
			& \colorbox{colorSnd}{49.94} & \colorbox{colorFst}{27.32} & \colorbox{colorTrd}{34.92} & \colorbox{colorSnd}{12.08} & \colorbox{colorFst}{8.06}\\			
			
			DM (YOLOv11-m)
			& 33.26 & 22.30 & 20.68 & 12.60 & 9.26
			& \colorbox{colorFst}{51.06} & 27.58 & 32.76 & \colorbox{colorFst}{11.98} & \colorbox{colorTrd}{8.48}\\				
			\bottomrule 
		\end{tabular}
	}
	\caption{FNR@95TNR and FPR@95TPR for degradation-induced covariate-shift detection on synthetically corrupted COCO variants. Lower values are better for both metrics. The table complements the main AUROC comparison by reporting performance at fixed operating points.}
	\label{tab:fnr95_fpr95_degradation_manifold_vs_all}
\end{table*}
\clearpage
\section{Joint Detection--Degradation Training Study}
\label{app:detdeg}
In the main paper, we adopt an auxiliary two-path configuration in which degradation-manifold learning is separated from the detector’s primary objective. This design preserves detection performance while enabling reliable degradation monitoring. In this section, we analyze the trade-off between detection robustness and degradation sensitivity under joint optimization.

\subsection{Experimental Setup}
To keep computational cost manageable for this side study, we conduct experiments on COCO-mini~\cite{Samet_ECCV_2020}, a stratified subset of COCO with preserved class distribution. We additionally evaluate a person-only subset (person instances larger than 40 pixels), where detection complexity is reduced and the effects of representation modification become more clearly observable.

Using YOLOv10-m as backbone, we compare the following configurations:
\begin{itemize}
	\item \textbf{Default detector training:} standard detection objective only.
	\item \textbf{Contrastive-only training:} backbone fine-tuned solely with the degradation contrastive loss.
	\item \textbf{Naive multi-task training:} joint optimization of detection loss and degradation contrastive loss using
	\[
	\mathcal{L}_{\text{total}} = \mathcal{L}_{\text{det}} + \lambda \mathcal{L}_{\text{deg}},
	\]
	with fixed weighting $\lambda$.
\end{itemize}

\subsection{Results}
Fig.~\ref{fig:map_coco_mini_person} compares mAP under different training objectives on the COCO-mini person subset~\cite{Samet_ECCV_2020}. Detection-only training preserves baseline accuracy. In contrast, contrastive-only training substantially degrades task performance, which is expected because the degradation objective promotes sensitivity to fidelity variation, whereas detection training seeks invariance to nuisance perturbations.

Naive multi-task training mitigates this effect considerably. Although a measurable performance gap remains relative to detection-only training, the jointly optimized model retains meaningful task performance while learning structured degradation representations. This indicates that joint optimization is feasible, but introduces a nontrivial trade-off between detection invariance and degradation sensitivity.

\subsection{Discussion}
These results highlight an inherent objective conflict. Detection training encourages invariance to nuisance variation in order to preserve task performance, whereas degradation-manifold learning intentionally amplifies sensitivity to fidelity changes in order to support monitoring. The auxiliary two-path design used in the main paper avoids this conflict and preserves full detection performance. The joint-training results reported here suggest that more advanced balancing strategies (such as dynamic loss weighting, gradient projection, partial backbone freezing, or parameter-efficient adaptation) may further reduce the trade-off and are promising directions for future work.

\renewcommand{\subw}{0.45\textwidth}
\begin{figure*}[h!]
	\centering
	\begin{subfigure}{\subw}
		\includegraphics[width=\linewidth]{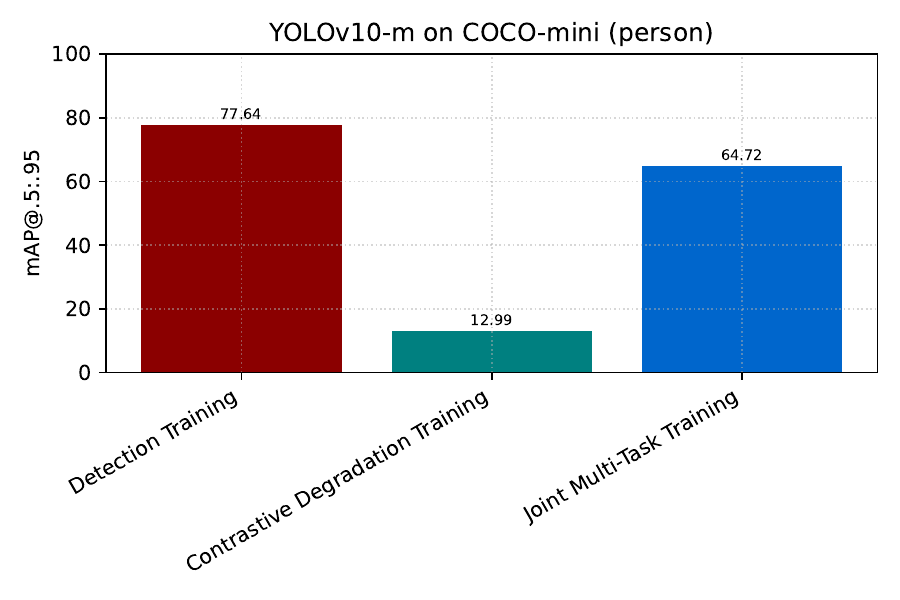}
		\caption*{mAP@0.5:0.95}
	\end{subfigure}
	\hfill
	\begin{subfigure}{\subw}
		\includegraphics[width=\linewidth]{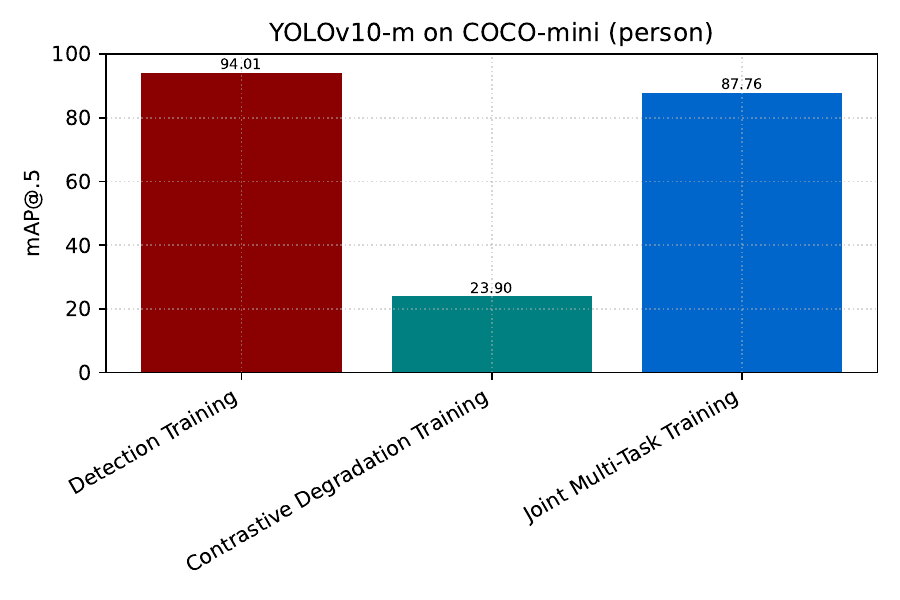}
		\caption*{mAP@0.5}
	\end{subfigure}
	
	\caption{
		Detection performance of YOLOv10-m~\cite{Wang_NeurIPS_2024} under different training objectives on the COCO-mini person subset~\cite{Samet_ECCV_2020}. 
		Detection-only training preserves baseline accuracy. 
		Degradation-only training significantly disrupts semantic detection features. 
		Joint multi-task optimization mitigates this effect while introducing a moderate performance trade-off.
	}
	\label{fig:map_coco_mini_person}
\end{figure*}

\clearpage
\section{Training Details}
\label{app:training}
This section summarizes the training procedure of the degradation manifold, including degradation composition, view generation, hard-negative construction, the contrastive objective, and implementation details.

\subsection{Degradation Sampling and Composition}
For each pristine training image $\mathbf{x} \sim \mathcal{D}$, we generate distorted views by sampling a degradation composition
\[
\mathcal{C} = \{ d_1, \dots, d_{n_{\text{deg}}} \},
\]
that is, an ordered sequence of degradation operators applied sequentially:
\[
\mathbf{x}^{\text{deg}} = d_{n_{\text{deg}}}(\dots d_2(d_1(\mathbf{x}))).
\]

The number of operators $n_{\text{deg}} \in \{1,\dots,N_{\text{deg}}\}$ is sampled uniformly. Each operator is drawn from a predefined pool of degradation groups (blur, noise, compression, brightness, color distortion, spatial distortion, sharpness/contrast), following the grouping strategy of~\cite{Agnolucci_WACV_2024} or using weather effects as described in~\cite{Hendrycks_ICLR_2019}. At most one operator per group is selected within a composition~\cite{Zhang_CVPR_2021,Zhang_CVPR_2022,Agnolucci_WACV_2024}. The order of selected operators is randomly shuffled, and each operator is parameterized by a randomly sampled severity level.

\begin{figure*}[h!]
	\centering
	\resizebox{\textwidth}{!}{%
		\begin{tikzpicture}[
			img/.style={draw, inner sep=0pt},
			arr/.style={->, very thick},
			brace/.style={decorate, decoration={brace, amplitude=6pt}},
			font=\small
			]
			
			\definecolor{brightness}{RGB}{0,180,170}
			\definecolor{spatial}{RGB}{165,90,45}
			\definecolor{colordist}{RGB}{220,190,0}
			\definecolor{sharp}{RGB}{90,180,0}
			\definecolor{blur}{RGB}{0,80,220}
			\definecolor{noise}{RGB}{220,0,0}
			\definecolor{compression}{RGB}{200,0,200}
			
			\def\imgw{1.2cm}
			\def\colsep{2.5cm}
			\def\rowsep{1.6cm}
			
			\node (X0) at (0,0) {};
			\node (X1) at (\colsep*1,0) {};
			\node (X2) at (\colsep*2,0) {};
			\node (X3) at (\colsep*3,0) {};
			\node (X4) at (\colsep*4,0) {};
			\node (X5) at (\colsep*5,0) {};
			
			\coordinate (Y1) at (0, \rowsep*1.5);
			\coordinate (Y2) at (0, \rowsep*0.5);
			\coordinate (Y3) at (0,-\rowsep*0.5);
			\coordinate (Y4) at (0,-\rowsep*1.5);
			
			\node[img] (ref) at ($(X0)+(0,0)$)
			{\includegraphics[width=\imgw]{figure/cascaded_deg/ref.png}};

			\node[img] (c1a) at ($(X1)+(Y1)$) {\includegraphics[width=\imgw]{figure/cascaded_deg/c1_1.png}};
			\node[img] (c2a) at ($(X1)+(Y2)$) {\includegraphics[width=\imgw]{figure/cascaded_deg/c2_1.png}};
			\node[img] (c3a) at ($(X1)+(Y3)$) {\includegraphics[width=\imgw]{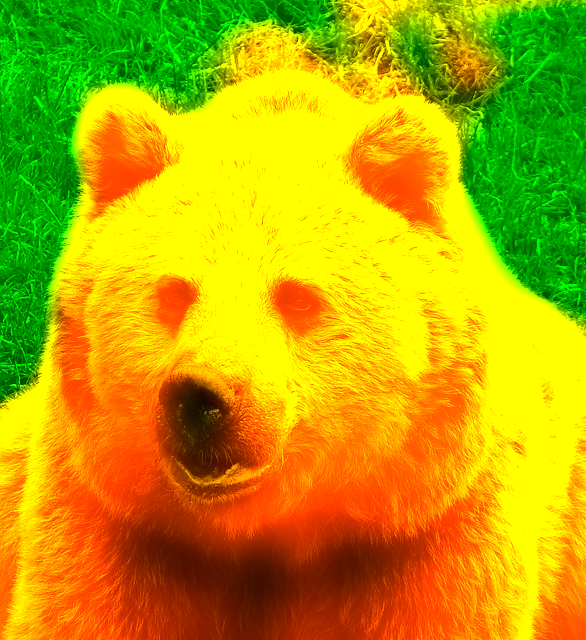}};
			\node[img] (c4a) at ($(X1)+(Y4)$) {\includegraphics[width=\imgw]{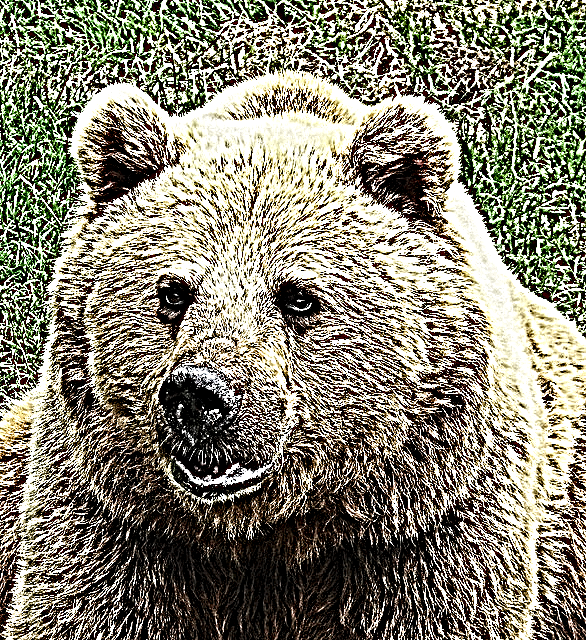}};
			
			\node[left=4pt of c1a] {$C_1$};
			\node[left=4pt of c2a] {$C_2$};
			\node[left=4pt of c3a] {$C_3$};
			\node[left=4pt of c4a] {$C_4$};
			
			\draw[arr, blur]        (ref.east) -- (c1a.west);
			\draw[arr, compression] (ref.east) -- (c2a.west);
			\draw[arr, colordist]   (ref.east) -- (c3a.west);
			\draw[arr, sharp]       (ref.east) -- (c4a.west);
			
			\node[img] (c1b) at ($(X2)+(Y1)$) {\includegraphics[width=\imgw]{figure/cascaded_deg/c1_2.png}};
			\draw[arr, spatial] (c1a.east) -- (c1b.west);
			
			\node[img] (c3b) at ($(X2)+(Y3)$) {\includegraphics[width=\imgw]{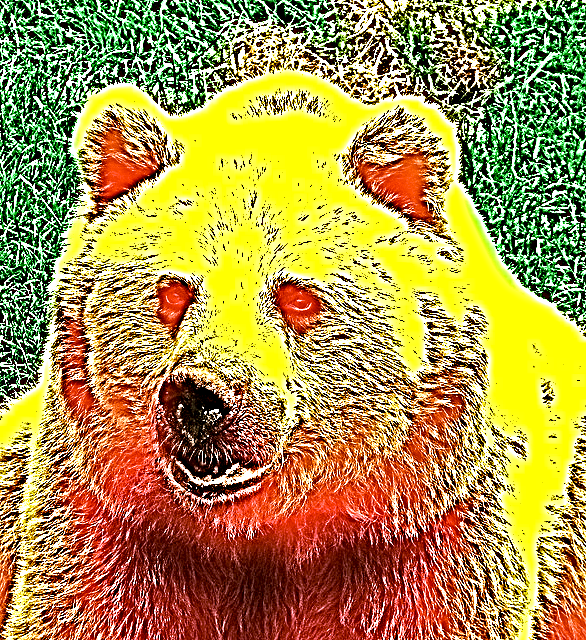}};
			\draw[arr, sharp] (c3a.east) -- (c3b.west);
			
			\node[img] (c4b) at ($(X2)+(Y4)$) {\includegraphics[width=\imgw]{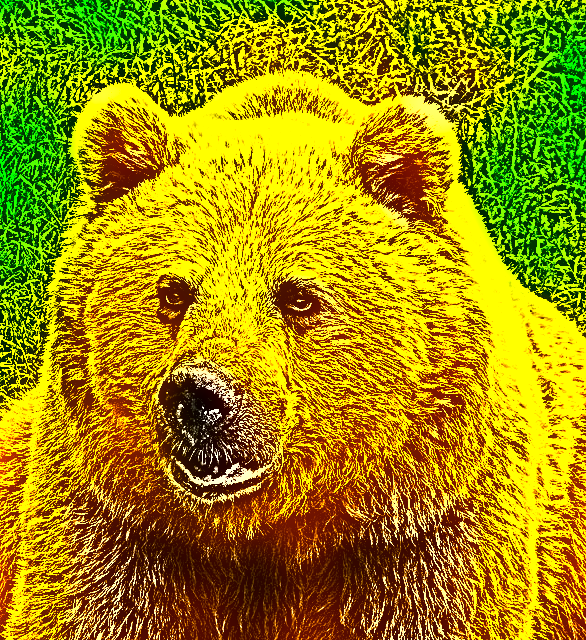}};
			\draw[arr, colordist] (c4a.east) -- (c4b.west);
			
			\node[img] (c3c) at ($(X3)+(Y3)$) {\includegraphics[width=\imgw]{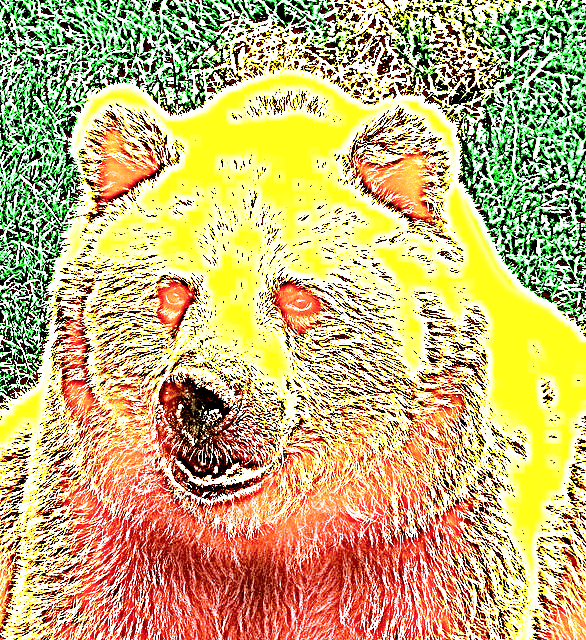}};
			\draw[arr, brightness] (c3b.east) -- (c3c.west);
			
			\node (dots) at ($(X4)+(Y4)$) {$\cdots$};
			\node[img] (final) at ($(X5)+(Y4)$) {\includegraphics[width=\imgw]{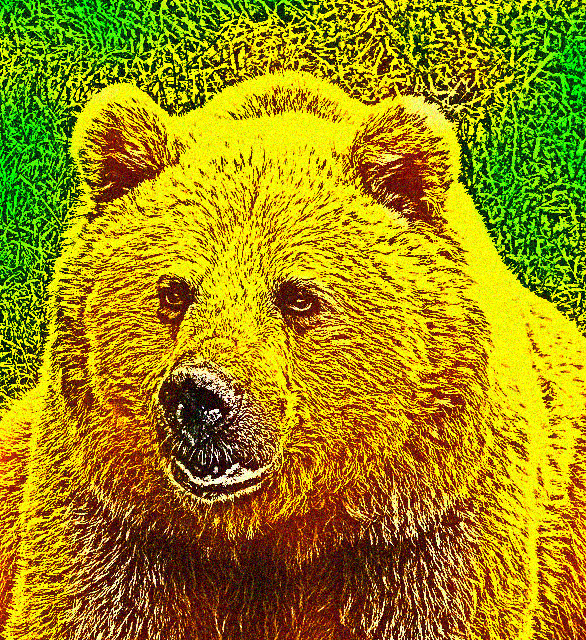}};
			\draw[arr, noise] (dots.east) -- (final.west);
			
			\draw[brace]
			($(final.south east)+(0,-0.35)$)
			--
			($(c4a.south west)+(0,-0.35)$)
			node[midway, below=6pt] {$N_{\text{deg}}$};
			
			\node[draw, dashed, anchor=north east] at ($(final.north east)+(2.0,+4.)$) {
				\begin{tabular}{ll}
					\multicolumn{2}{c}{\textbf{Degradation groups from \cite{Agnolucci_WACV_2024}}} \\[4pt]
					\textcolor{brightness}{\rule{6pt}{6pt}} & Brightness change \\
					\textcolor{spatial}{\rule{6pt}{6pt}} & Spatial distortions \\
					\textcolor{colordist}{\rule{6pt}{6pt}} & Color distortions \\
					\textcolor{sharp}{\rule{6pt}{6pt}} & Sharpness \& contrast \\
					\textcolor{blur}{\rule{6pt}{6pt}} & Blur \\
					\textcolor{noise}{\rule{6pt}{6pt}} & Noise \\
					\textcolor{compression}{\rule{6pt}{6pt}} & Compression
				\end{tabular}
			};
			
		\end{tikzpicture}%
	}
	\caption{Overview of the applied image degradation model. Randomly assembled degradation compositions
		\(C\) are applied sequentially to synthesize diverse degradation patterns. Each degradation composition contains
		up to \(N_{\text{deg}}\) degradations sampled from pre-defined degradation groups (e.g., the degradation groups of \cite{Agnolucci_WACV_2024}).}
	\label{fig:degcomp}		
\end{figure*}

This stochastic chaining yields a large variety of structured degradation regimes while preserving semantic content. Evaluation corruptions follow taxonomies used in robustness benchmarks. They partially differ from those used during training, resulting in partial zero-shot generalization. The degradation-composition process is visualized in Fig.~\ref{fig:degcomp}.

\subsection{View Generation}
For each sampled composition $\mathcal{C}_i$, we generate two independently parameterized degraded views:
\[
(\mathbf{x}_{i,A}^{\text{deg}}, \mathbf{x}_{i,B}^{\text{deg}}).
\]
Both views share the same operator sequence but differ in sampled parameters, such as blur kernel size or noise variance. This enforces invariance within a degradation regime while preserving regime identity.

All images are resized to the detector’s default input resolution (e.g., $640\times640$) before feature extraction.

To strengthen discrimination between subtle fidelity changes, we construct additional hard negatives. For each degraded view, we extract a centered crop covering half the spatial extent and resize it back to the detector’s default input resolution:
\[
\tilde{\mathbf{x}}^{\text{deg}} 
= 
\mathrm{Resize}\!\left(
\mathrm{CenterCrop}_{1/2}(\mathbf{x}^{\text{deg}})
\right).
\]

\begin{figure*}[h!]
	\centering
	\resizebox{0.9\textwidth}{!}{%
		\begin{tikzpicture}[
			img/.style={draw, inner sep=0pt},
			dashedbox/.style={draw, dashed, thick},
			arr/.style={->, thick},
			posarr/.style={->, thick, green!70!black},
			negarr/.style={->, thick, red},
			hardarr/.style={->, thick, dashed},
			font=\small
			]
			
			\def\colsep{5.0cm}
			\def\rowsep{5.8cm}
			\def\imgW{2.8cm}
			\def\cropW{2.0cm}
			\def\embW{1.9cm}
			\def\embH{0.7cm}
			
			\def\orgY{0.0cm}
			\def\hnY{-3.0cm}
			
			\node (C0) at (0,0) {};
			\node (C1) at (\colsep,0) {};
			\node (C2) at (2*\colsep,0) {};
			
			\coordinate (R1) at (0, 0.5*\rowsep);
			\coordinate (R2) at (0,-0.5*\rowsep);
			
			\node at ($(C0)+(0, 0.95*\rowsep)$) {\textbf{Pristine images}};
			\node at ($(C1)+(0, 0.95*\rowsep)$) {\textbf{Degraded images}};
			\node at ($(C2)+(0, 0.95*\rowsep)$) {\textbf{Embeddings}};
			
			\node[img] (p1) at ($(C0)+(R1)+(0,\orgY)$)
			{\includegraphics[width=\imgW]{figure/contrastive_fig/img1.png}};
			
			\node[img, dashedbox] (p1dsc) at ($(C0)+(R1)+(0,\hnY)$)
			{\includegraphics[width=\cropW]{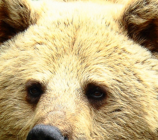}};
			\draw[arr] (p1.south) -- node[right]{CenterCrop$_{1/2}$} (p1dsc.north);
			
			\node[img] (d1) at ($(C1)+(R1)+(0,\orgY)$)
			{\includegraphics[width=\imgW]{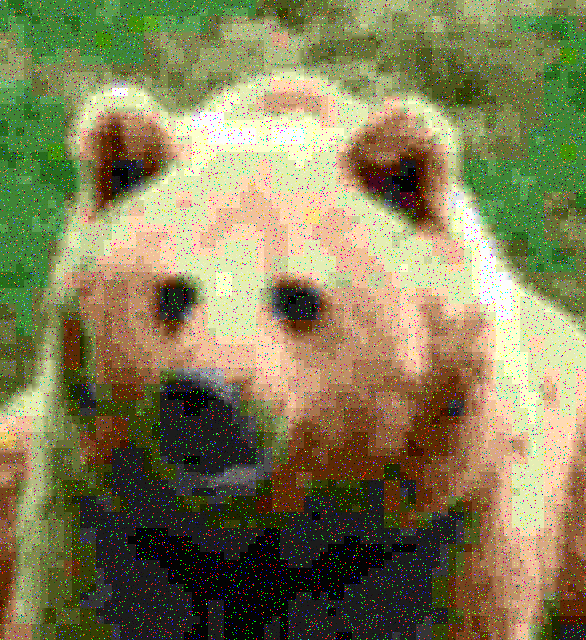}};
			\draw[arr] (p1.east) -- node[above]{Degrade} (d1.west);
			
			\node[img, dashedbox] (d1h) at ($(C1)+(R1)+(0,\hnY)$)
			{\includegraphics[width=\cropW]{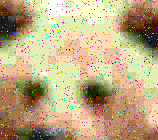}};
			\draw[arr] (p1dsc.east) -- node[above]{Degrade} (d1h.west);
			
			\node[dashedbox, minimum width=\embW, minimum height=\embH] (z1)
			at ($(C2)+(R1)+(0,\orgY)$) {$z^1$};
			\node[dashedbox, minimum width=\embW, minimum height=\embH] (zh1)
			at ($(C2)+(R1)+(0,\hnY)$) {$z^1_{\text{hn}}$};
			
			\draw[arr] (d1.east)  -- node[above]{Encode} (z1.west);
			\draw[arr] (d1h.east) -- node[above]{Resize + Encode} (zh1.west);
			
			\node[img] (p2) at ($(C0)+(R2)+(0,\orgY)$)
			{\includegraphics[width=\imgW]{figure/contrastive_fig/img2.png}};
			
			\node[img, dashedbox] (p2dsc) at ($(C0)+(R2)+(0,\hnY)$)
			{\includegraphics[width=\cropW]{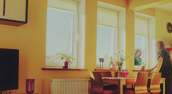}};
			\draw[arr] (p2.south) -- node[right]{CenterCrop$_{1/2}$} (p2dsc.north);
			
			\node[img] (d2) at ($(C1)+(R2)+(0,\orgY)$)
			{\includegraphics[width=\imgW]{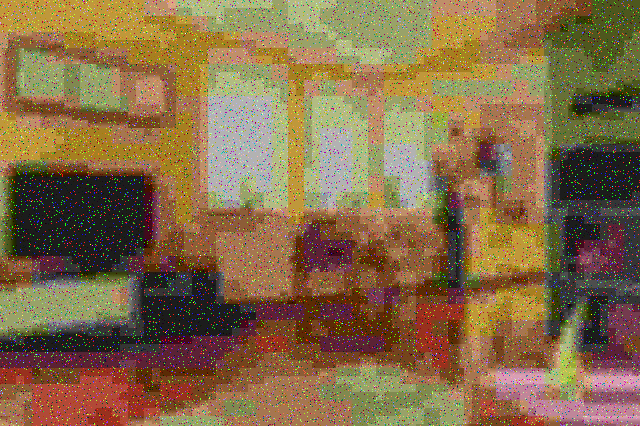}};
			\draw[arr] (p2.east) -- node[above]{Degrade} (d2.west);
			
			\node[img, dashedbox] (d2h) at ($(C1)+(R2)+(0,\hnY)$)
			{\includegraphics[width=\cropW]{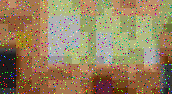}};
			\draw[arr] (p2dsc.east) -- node[above]{Degrade} (d2h.west);
			
			\node[dashedbox, minimum width=\embW, minimum height=\embH] (z2)
			at ($(C2)+(R2)+(0,\orgY)$) {$z^2$};
			\node[dashedbox, minimum width=\embW, minimum height=\embH] (zh2)
			at ($(C2)+(R2)+(0,\hnY)$) {$z^2_{\text{hn}}$};
			
			\draw[arr] (d2.east)  -- node[above]{Encode} (z2.west);
			\draw[arr] (d2h.east) -- node[above]{Resize + Encode} (zh2.west);

			\draw[posarr, bend left=12] (z1)  to (z2);
			\draw[posarr, bend left=12] (zh1) to (zh2);
			
			\draw[negarr, bend right=18] (z1) to (zh1);
			\draw[negarr, bend right=18] (z2) to (zh2);
			
			\draw[hardarr] (z1) -- (zh1);
			\draw[hardarr] (z2) -- (zh2);
			
			\draw[negarr, bend right=35] (z1) to (zh2);
			\draw[negarr, bend right=35] (z2) to (zh1);

				\node[draw, dashed, anchor=north west] at ($(C2)+(2.2cm, 0.0cm)$) {
					\begin{tabular}{ll}
						\textcolor{green!70!black}{$\uparrow$} & Maximize similarity \\
						\textcolor{red}{$\downarrow$} & Minimize similarity \\
						$\dashrightarrow$ & Hard negative example
					\end{tabular}
				};
				
			\end{tikzpicture}%
	}
	\caption{
		Training pair construction for degradation-manifold learning.
		Two pristine images are degraded using the same sampled composition, forming a positive pair in embedding space ($z^1,z^2$).
		To construct hard negatives, we apply an additional half-size crop before resizing to the detector input resolution.
		All degraded views are resized to the detector’s default input size prior to encoding.
		The objective pulls embeddings of equally degraded full-resolution images together while pushing them away from their resolution-perturbed counterparts, encouraging sensitivity to fidelity changes beyond semantic similarity.
	}
		
	\label{fig:degtrainig}		
\end{figure*}

This procedure introduces resolution-induced information loss while largely preserving semantic content. An overview of the training-pair construction is shown in Fig.~\ref{fig:degtrainig}.

\subsection{Contrastive Objective}
Let $z_i^A$ and $z_i^B$ denote embeddings of two degraded full-resolution views for sample $i$, and $\hat{z}_i^A$, $\hat{z}_i^B$ the corresponding hard-negative embeddings. Embeddings are obtained from the detector backbone followed by the projection head.

For a batch of size $B$, we employ the NT-Xent loss~\cite{chen_ICML_2020} with cosine similarity and temperature $\tau$. Define
\[
\gamma(a,b) = \exp\!\left( \frac{\cos(a,b)}{\tau} \right).
\]

Each full-resolution pair $(z_i^A, z_i^B)$ forms a positive pair, while all other embeddings in the batch, including hard negatives, serve as negatives. The per-sample loss term is
\[
\ell_i^{A,B}
=
-
\log
\frac{
	\gamma(z_i^A, z_i^B)
}{
	\sum\limits_{k=1}^{B}
	\big[
	\gamma(z_i^A, z_k^B)
	+
	\gamma(z_i^A, \hat{z}_k^B)
	+
	\gamma(z_i^A, \hat{z}_k^A)
	\big]
	+
	\sum\limits_{k \neq i}^{B}
	\gamma(z_i^A, z_k^A)
}.
\]

Symmetric terms are defined analogously for $(B,A)$ and for hard-negative anchors. The final loss is
\[
\mathcal{L}
=
\frac{1}{4B}
\sum_{i=1}^{B}
\left[
\ell_i^{A,B}
+
\ell_i^{B,A}
+
\hat{\ell}_i^{A,B}
+
\hat{\ell}_i^{B,A}
\right].
\]

This objective pulls embeddings of identically composed degradations together while pushing apart different compositions and resolution-perturbed hard negatives, thereby structuring the embedding space according to degradation regimes.

\subsection{Implementation Details}
The model is implemented in PyTorch~\cite{Paszke_Pytorch_2019}, building on the detector implementation provided by Ultralytics~\cite{ultralytics_2023}. Training uses the AdamW optimizer~\cite{Kingma_ICLR_2015,Loshchilov_ICLR_2019} with an initial learning rate of $1\times10^{-5}$. Experiments are conducted on a single NVIDIA GeForce RTX~5090 GPU with a batch size of $6$. Each batch yields six standard contrastive pairs and six corresponding hard-negative pairs generated via resolution perturbation. Images are resized to the detector’s default input resolution of $640$ during training, and the model is trained for $100$ epochs.

\subsection{Inference Overhead}
\label{app:runtime}
We measure the additional inference cost of the degradation monitor over $5,000$ images under identical preprocessing and runtime settings.

The YOLOv10-m forward pass requires \(4.87\,\mathrm{ms}\), whereas DM including the final degradation score requires \(5.54\,\mathrm{ms}\), corresponding to \(0.67\,\mathrm{ms}\) additional model-side latency.

Mean end-to-end latency increases from \(14.82\,\mathrm{ms}\) to
\(15.47\,\mathrm{ms}\), corresponding to a \(1.044\times\) runtime ratio and
\(4.38\%\) additional latency. Effective throughput changes from \(67.49\)
FPS to \(64.65\) FPS (\(-4.20\%\)). Median end-to-end latency changes from
\(9.23\,\mathrm{ms}\) to \(9.84\,\mathrm{ms}\). Overall, the monitoring branch
adds less than \(1\,\mathrm{ms}\) per image and approximately \(4.4\%\)
end-to-end runtime overhead.

\begin{table}[t]
	\centering	
	\begin{tabular}{lcccc}
		\hline
		Configuration
		& Model (ms)
		& E2E Mean (ms)
		& E2E Median (ms)
		& FPS \\
		\hline
		YOLOv10-m
		& 4.87 & 14.82 & 9.23 & 67.49 \\
		Degradation manifold
		& 5.54 & 15.47 & 9.84 & 64.65 \\
		Change
		& +0.67 ms & +4.38\% & +6.61\% & -4.20\% \\
		\hline
	\end{tabular}
	\caption{Runtime comparison with and without the degradation-monitoring
		branch over 5,000 images. FPS is derived from mean per-image latency.}
	\label{tab:runtime_overhead}
\end{table}

\section{The Use of Large Language Models (LLMs)}
Large Language Models (LLMs) were used solely for writing refinement, coding assistance, literature search, and summarization. All conceptual development, experimental design, implementation, analysis, and initial writing of the paper were carried out by the authors.

\end{document}